\PassOptionsToPackage{dvipsnames,table}{xcolor}
\documentclass[sigconf=true, nonacm=true, review=false, anonymous = false]{acmart}
\usepackage{booktabs} 
\usepackage{xspace}
\usepackage{amsmath}
\usepackage{xspace}
\usepackage{xcolor}
\usepackage{dsfont}
\usepackage{rotating}
\usepackage{makecell}
\usepackage{array}
\usepackage[newcommands]{ragged2e}
\newcommand{\PreserveBackslash}[1]{\let\temp=\\#1\let\\=\temp}
\newcolumntype{L}[1]{>{\PreserveBackslash\raggedright\hyphenpenalty=10000}p{#1}}
\newcolumntype{R}[1]{>{\PreserveBackslash\raggedleft}p{#1}}
\newcolumntype{C}[1]{>{\PreserveBackslash\centering}p{#1}}

\usepackage{newrevisor}
\newrevisor{manuel}{violet!75}
\newrevisor{youngmin}{red!75}
\newrevisor{richard}{blue!75}
\usepackage{amsthm}
\usepackage{subcaption}
\renewcommand{\vec}[1]{\ensuremath{\mathbf{#1}}}
\newcommand{\Lipschitz}{Lipschitz\xspace}
\newcommand{\nth}{\ensuremath{^{\mathrm{th}}}}
\newcommand{\DecisionSpace}{\ensuremath{\mathbb{R}^d}}

\newcommand{\StageOpt}{\textsc{StageOpt}\xspace}

\newcommand{\psosafeopt}{Swarm-\hspace{0pt}based \textsc{SafeOpt}\xspace}

\newcommand{\SafeOpt}{\textsc{SafeOpt}\xspace}
\newcommand{\SafeUCB}{\textsc{Safe-UCB}\xspace}
\AtBeginDocument{%
  \providecommand\BibTeX{{%
    \normalfont B\kern-0.5em{\scshape i\kern-0.25em b}\kern-0.8em\TeX}}}

\copyrightyear{2022} 
\acmYear{2022} 
\setcopyright{acmcopyright}\acmConference[GECCO '22]{Genetic and Evolutionary Computation Conference}{July 9--13, 2022}{Boston, MA, USA}
\acmBooktitle{Genetic and Evolutionary Computation Conference (GECCO '22), July 9--13, 2022, Boston, MA, USA}
\acmPrice{15.00}
\acmDOI{10.1145/3512290.3528818}
\acmISBN{978-1-4503-9237-2/22/07}

\begin{document}

\title{Are Evolutionary Algorithms Safe Optimizers?}
\thanks{Please cite as: Youngmin Kim, Richard Allmendinger, and Manuel López-Ibañez. 2022. Are Evolutionary Algorithms Safe Optimizers?. In
  \emph{Genetic and Evolutionary Computation Conference (GECCO ’22)}, July
  9–13, 2022, Boston, MA, USA. ACM, New York, NY, USA, 9
  pages. \url{https://doi.org/10.1145/3512290.3528818}}

\author{Youngmin Kim}
\email{youngmin.kim@manchester.ac.uk}
\orcid{0000000276996532}
\affiliation{%
  \institution{University of Manchester}
  \streetaddress{Oxford Rd}
  \city{Manchester}
  \country{UK}
  \postcode{M15 6PB}
}

\author{Richard Allmendinger}
\email{richard.allmendinger@manchester.ac.uk}
\orcid{0000000312363143}
\affiliation{%
  \institution{University of Manchester}
  \streetaddress{Oxford Rd}
  \city{Manchester}
  \country{UK}
  \postcode{M15 6PB}
}

\author{Manuel López-Ibáñez}
\email{manuel.lopez-ibanez@manchester.ac.uk}
\orcid{0000000199741295}
\affiliation{%
  \institution{University of Manchester}
  \streetaddress{Oxford Rd}
  \city{Manchester}
  \country{UK}
  \postcode{M15 6PB}}
\additionalaffiliation{
  \institution{ITIS Software, School of Computer Science, University of Málaga}
  \city{Málaga}
  \country{Spain}
  \postcode{29071}
}



\begin{abstract}
We consider a type of constrained optimization problem, where the violation of a constraint leads to an irrevocable loss, such as breakage of a valuable experimental resource/platform or loss of human life. Such problems are referred to as safe optimization problems (SafeOPs). While SafeOPs have received attention in the machine learning community in recent years, there was little interest in the evolutionary computation (EC) community despite some early attempts between 2009 and 2011. Moreover, there is a lack of acceptable guidelines on how to benchmark different algorithms for SafeOPs, an area where the EC community has significant experience in. Driven by the need for more efficient algorithms and benchmark guidelines for SafeOPs, the objective of this paper is to reignite the interest of the EC community in this problem class. To achieve this we (i)~provide a formal definition of SafeOPs and contrast it to other types of optimization problems that the EC community is familiar with, (ii)~investigate the impact of key SafeOP parameters on the performance of selected safe optimization algorithms, (iii)~benchmark EC against state-of-the-art safe optimization algorithms from the machine learning community, and (iv) provide an open-source Python framework to replicate and extend our work. 

\end{abstract}



\maketitle

\section{Introduction}
This work focuses on \emph{safe optimization problems}, a special type of constrained optimization problem that has received rather little attention in the evolutionary computation (EC) community, but more so in the machine learning community. Such problems are subject to constraints that, when violated, result in an irrevocable loss of valuable experimental platform/resource, such as breakage of a machine used for experiments, or even injury to a patient~\cite{KimAllLop2020safe}. Here, these constraints are referred to as safety constraints, and evaluations of input points (candidate solutions) that violate a safety constraint are called unsafe evaluations. Typically, the objective function and any \emph{safety constraint function} that defines one side of a safety constraint (these concepts will be explained in detail in Section~\ref{sec:ps})  are given as black-box functions and their evaluation is expensive. 
Examples of safe optimization problems (SafeOPs) include clinical experiments~\cite{SuiZhuBur2018stageopt,SuiGotBur2015icml}, controller optimization for quadrotor vehicle~\cite{DuiBerCar2017constrained,BerSchKra2016safe,BerKraSch2021bayesian}, engine calibration~\cite{SchHarSka2017safe,KajIkeHaj2009cec}, and simulation-based optimization~\cite{BacHelPic2020gaussian,SacDuvMai2018ego-ls-svm}.

There are two algorithmic branches in safe optimization~\cite{KimAllLop2020safe}: Safe optimization through evolutionary algorithms (safe EAs) vs Gaussian process (GP) regression (safe GPs). Although safe optimization was first considered by the EC community in 2009~\cite{KajIkeHaj2009cec} and 2011~\cite{AllKno2011ecta,Allmendinger2012phd}, we are not aware of any further research on this topic. In comparison, the machine learning community  has  actively worked on SafeOPs  from 2015. 

Research on safe optimization is fragmented with no unified guidelines on how to benchmark algorithms for safe optimization.  
When algorithms for SafeOPs are benchmarked, an aspect of performance relates to the best objective function values achieved~\cite{SchHarSka2017safe,SacDuvMai2018ego-ls-svm,BerSchKra2016safe,BacHelPic2020gaussian,KajIkeHaj2009cec,AllKno2011ecta,Allmendinger2012phd,SuiGotBur2015icml,SuiZhuBur2018stageopt}. Another aspect of performance relates to safety, which can be measured, for example, by the number of unsafe evaluations (or, equivalently, the number of safe evaluations) at each iteration or evaluation step~\cite{BacHelPic2020gaussian,SacDuvMai2018ego-ls-svm,BerSchKra2016safe,KajIkeHaj2009cec}, the proportion of survived solutions, i.e., $u_t/ u_0$ where $u_0$ is the initial parent population size and $u_t$ is the number of survived offspring size at $t\nth$ iteration step (individuals violating the safety constraint are removed)~\cite{AllKno2011ecta,Allmendinger2012phd}, sensitivity and specificity on the evaluations~\cite{SchHarSka2017safe}, and the size of safe set (i.e., the number of input points inferred to be safe)~\cite{SuiGotBur2015icml,SuiZhuBur2018stageopt}. However, most papers that propose an algorithm for SafeOPs only benchmark them against similar type of algorithms. In such studies, the capabilities of the proposed algorithms under slightly different SafeOP scenarios are seldom examined.

The contributions made by our paper are as follows:
\begin{enumerate}
    \item We provide a formal definition of SafeOPs, discuss their real-world application, and contrast them to other types of problem that the EC community is familiar with.
    \item We investigate the impact on  performance of safe optimization algorithms for several key parameters affecting the complexity of SafeOPs.
    \item This is the first study that compares safe EA with safe GP algorithms. Previous studies looked at the two algorithm types in isolation (see, for example,~\cite{SuiGotBur2015icml,BerSchKra2016safe,KajIkeHaj2009cec}).
    \item We propose an initial set of guidelines to carry out benchmark studies of algorithms for SafeOPs. We also make available to the community an open-source Python framework that facilitates the replication and extension of our work.
\end{enumerate}

The remainder of the paper is organized as follows. In Section~\ref{sec:ps}, a formal definition of the particular SafeOPs used for our experiments is presented. Section~\ref{sec:FC} describes the working principles of existing safe optimization algorithms, and Section~\ref{sec:ES} provides the experimental setup for the benchmark study carried out in Section~\ref{sec:Results}. Finally, conclusions and future research are discussed in Section~\ref{sec:Conclusion}.

\section{Problem Statement\label{sec:ps}}
A safe optimization problem (SafeOP) can be formally defined as:
\begin{align}\label{eq:gd}
  \text{maximize}\quad   & y(\vec{x}) = f(\vec{x}) + \epsilon,\enspace\;\; \epsilon \sim \mathcal{N}(0,\sigma^2),\enspace  \vec{x} \in D \subset \DecisionSpace\\
 \text{s.t.}\quad & y(\vec{x}) \geq h  \qquad\qquad \text{(safety constraint)}
\end{align}
%
where $\vec{x}=(x_1,\dotsc,x_d)$ is a $d$-dimensional input point (candidate solution), $D$ is the feasible search space (defined by standard feasibility constraints), $f\colon D \to \mathbb{R}$ is an objective function that is black-box and expensive to evaluate, and observed as a value $y(\vec{x}) = f(\vec{x})+\epsilon$ perturbed by additive Gaussian noise~\cite{SuiGotBur2015icml,BerSchKra2016safe}.\footnote{This is the most widely-used model, however, in practice, the noise may not be Guassian.} A safety constraint is defined by a black-box \emph{safety function} (left-hand side) and a known  \emph{safety threshold}, the constant $h$. For simplicity, in this paper, we consider a single safety constraint and  $y(\vec{x})$ is simultaneously the objective function and the safety function such that  evaluating an input point whose $y(\vec{x})$ value is less than $h$ is unsafe. Evaluating an unsafe solution results in an irrevocable scenario, such as loss of an expensive kit or injury of a patient (more examples below). %
The \textit{safety budget} determines the maximum number of unsafe solutions that may be evaluated, i.e., \emph{unsafe evaluations}, usually ranging from 0 (no unsafe evaluations are allowed) to a small integer number (e.g. in case a limited number of kits are available). The optimization process of a SafeOP is terminated if either the evaluation budget or the safety budget are exhausted. %
To kickstart the optimization of a SafeOP, the optimizer is often provided a set of known safe solutions, also known as the \textit{initial safe seeds}.

A real-world example of a SafeOP is spinal cord therapy where experiments were conducted to rats suffering spinal cord injuries~\cite{SuiGotBur2015icml}. The goal of the study was to search for electrical-stimulating configurations that optimize the resulting activity in lower limb muscles, to improve spinal reflex
and locomotor function. Here, unsafe evaluations have negative impacts on rehabilitation. This is an example of a scenario with zero safety budget (no unsafe evaluations allowed).

The definition of a SafeOP can be made arbitrarily more complex without being less realistic, for example, by having an unknown safety threshold $h$, which itself can be a function of $\vec{x}$, or multiple black-box objective functions and safety constraints. However, for the sake of simplicity, here we limit ourselves to a single safety constraint with a known constant $h$.

Although SafeOPs share similarities with standard constrained optimization problems~\cite{coecarconstraint,Michalewicz1995ASO}, which are subject to feasibility constraints only, there are important differences. 
Feasibility constraints reflect some aspects relevant for an input point $\vec{x}$ to be of practical use, for instance, bounds of instrument settings, physical limitations, or design requirements~\cite{KimAllLop2020safe}. Also, there is no limit on the number of violations of a feasibility constraint, and violation of a feasibility constraint can be predicted when mathematical models are available (meaning violations can be avoided). Independently of whether an infeasible solution can be evaluated or not, nothing severe happens to the optimizer or the environment. The objective function value of infeasible solutions is either undefined or simply penalized to encourage the optimizer to remain in the feasible search space. In comparison, violating a safety constraint beyond the safety budget results in premature termination of an optimization run, and predicting violations is trickier because of their black-box nature. A solution $\vec{x}$ can be feasible but unsafe, and vice versa. 
Furthermore, in safe optimization, an optimization algorithm that achieves a worse performance in terms of $f$ but evaluates fewer unsafe solutions may be preferred over an algorithm that evaluates more unsafe solutions to achieve a better performance in $f$. The real-world cost of violating a safety constraint may be a key factor in balancing this performance-safety trade-off. 

To keep things simple, in this work we assume no feasibility constraints, apart from box constraints on the decision variable values $x_i$. 

\section{Safe Optimization Algorithms: A Brief Review\label{sec:FC}}

\noindent\textbf{EAs for safe optimization. }\quad SafeOPs were initially studied in the EC community between 2009 and 2011. A violation avoidance (VA) method~\cite{KajIkeHaj2009cec} was proposed in 2009, as a universal approach that can be augmented onto any EA when faced with a SafeOP. VA replaces the offspring generation process of an EA by rejecting and then regenerating an offspring solution if its closest solution (or nearest neighbor) in the decision space out of all previously evaluated solutions (i.e. the search history) was unsafe. In~\cite{AllKno2011ecta,Allmendinger2012phd}, a reconfigurable, destructible and unreplaceable experimental platform in closed-loop optimization setting was considered using three different stochastic optimizers: Tournament selection based genetic algorithm (TGA), reproduction of best solutions (RBS), and a population of stochastic hill-climbers (PHC). The three optimizers cannot infer the safety of an input point before it is actually evaluated, however, they guide the population away from unsafe regions in the search space by prohibiting unsafe input points from entering the population.

\noindent\textbf{GPs for safe optimization. }\quad SafeOPs have been actively studied in the machine learning community from 2015. \SafeOpt~\cite{SuiGotBur2015icml} was proposed to deal with a single-objective SafeOPs subject to a single safety constraint inspiring other safe GP algorithms. Later, a modified \SafeOpt~\cite{BerSchKra2016safe} was proposed that makes weaker assumptions about the black-box objective function; \SafeOpt-MC~\cite{BerKraSch2016bayesian,BerKraSch2021bayesian} was designed for dealing with multiple safety constraints in 2016; \psosafeopt~\cite{DuiBerCar2017constrained} was proposed in 2017 that applies a variant of particle swarm optimzation to \SafeOpt; and, in 2018, \StageOpt~\cite{SuiZhuBur2018stageopt} was designed for optimizing the objective function in two independent stages of learning and optimization. The interested reader is referred to~\cite{KimAllLop2020safe} for a detailed review of existing safe GP algorithms. 

Here, we provide a high level description of selected safe GP algorithms: \SafeOpt, modified \SafeOpt, \SafeUCB~\cite{SuiGotBur2015icml} and modified \SafeUCB~\cite{BerSchKra2016safe}, which are the ones that will also be considered in the experimental study of this paper (alongside VA). 
These safe GP algorithms (also VA) are initialized with at least one known safe solution (i.e. an initial safe seed). However, while VA uses the information obtained from the initial safe seed to infer the safety of an offspring using the nearest neighbor method, \SafeOpt and its variant \SafeUCB use the initial safe seed to establish a safe set (set of solutions inferred as being safe) based on the concept of $L$-\Lipschitz continuity. This concept deems an input point $\vec{x} \in D \subset \DecisionSpace$ as safe, if
\begin{equation}\label{eq:lip}
l(\vec{x}^{s})-L\cdot d(\vec{x}^{s},\vec{x}) \geq h 
\end{equation}
for at least one input $\vec{x}^{s} \in S_{t-1}$, where $S_{t-1}$ denotes a safe set estimated at the previous iteration step $t-1$ of the algorithm, $l(\vec{x}^{s})$ is the lower bound of the predicted confidence interval for $\vec{x}^{s}$, $d(\cdot,\cdot)$ denotes distance between two input points (we use the Euclidean distance, which is often used in practice~\cite{BiyMarAli2019acc}), and $L$ is the \Lipschitz constant (see the supplementary material for technical details on the computation of the bounds). Here, $S_{0}$ consists of the initial safe seed(s). On the other hand, in modified \SafeOpt and modified \SafeUCB, the initial safe seed is used for GP regression to find any input point $\vec{x} \in D \subset \DecisionSpace$ that meets
\begin{equation}
    \label{noL}
    l(\vec{x}) \geq h \enspace.
\end{equation}
When using safe GP algorithms, the search space is typically discretized uniformly (into e.g., $50 \times 50$ for two-dimensional input space~\cite{SuiGotBur2015icml}) to provide a finite set of input points for constructing safe set $S_t$ at each iteration step. %

Finally, an input point is selected in two different ways in the literature. One approach uses the safe set to construct further sets called \emph{maximizers set} (set of input points likely to result in high objective values when evaluated) and \emph{expanders set} (set of input points that may increase the size of the safe set when evaluated), and selects for evaluation the input point from the union of the two sets with the largest  width of predicted confidence interval (\SafeOpt and modified \SafeOpt). An alternative and simpler approach selects the point directly from the safe set whose upper confidence bound (UCB) is the maximum (\SafeUCB and modified \SafeUCB).
 Here, selecting an input point according to the width of predicted confidence interval or to the UCB criterion corresponds to adopting two predominant approaches in stochastic optimization, that is, \textit{Bayesian experimental design} (exploring the objective function globally as efficiently as possible) or \textit{multi-armed bandit paradigm} (maximizing cumulative reward), respectively~\cite{srinivas10gaussian}.

In addition to differences in the working principles between safe EAs and safe GP algorithms (at least the ones described above), the two algorithm types vary in further key aspects: Safe GP algorithms are designed to cope with noisy observations and require a discretized search space, while safe EA (VA) does not have a built-in noise-handling strategy nor requires discretization. Also, \SafeOpt and \SafeUCB assume that the \Lipschitz constant $L$ is known, which is a strong assumptions to make in practice. Other safe GP algorithms vary in aspects, such as the method used to infer the safety level of a solution, generate solutions to evaluate next, and reliance on an initial safe seed and the \Lipschitz continuity assumption~\cite{KimAllLop2020safe}. 



\section{Experimental Setup}\label{sec:ES}

This section explains the experimental setup used for benchmarking safe optimization algorithms. In particular, we motivate the choice of algorithms and their parameter settings, and explain how we simulate SafeOPs for benchmarking in the subsequent experimental study. 

\subsection{Algorithm Selection and Settings}\label{sec:discussionforexperiments}
\noindent\textbf{Algorithm selection.} The benchmark study will consider a safe EA representative, VA~\cite{KajIkeHaj2009cec} augmented onto a generational EA with a ($\mu + \lambda$)-ES environmental selection strategy~\cite{baeck2000evolutionary}, and four safe GP algorithms: \SafeOpt~\cite{SuiGotBur2015icml}, \SafeUCB~\cite{SuiGotBur2015icml}, modified \SafeOpt~\cite{BerSchKra2016safe}, modified \SafeUCB~\cite{BerSchKra2016safe}. As a baseline, we will also consider a generational EA with a ($\mu + \lambda$)-ES environmental selection strategy without the VA (we will refer to this algorithm as UnsafeEA), meaning this algorithm is blind to safety constraints. The set of algorithms selected allows us to compare the performance of two different working principles (EAs vs GPs). The selected algorithms are also well-known and widely used in their own communities (especially UnsafeEA and \SafeOpt). Finally, with respect to the safe GPs, we have a good mix of algorithms varying in terms of their assumptions and input point selection criteria. 
\begin{table}[t]
\caption{Algorithm parameter settings used in the experimental study.\label{tab:tab2}}
\resizebox{\columnwidth}{!}{%
\centering\begin{tabular}{L{5em}L{8.5em}L{10.5em}@{}}
\toprule
\bf Algorithm & \bf Parameter                                & \bf Setting                                          \\\midrule
Safe GP      & $\beta_{t}$                                  & 2                                                    \\
 algorithms    & \Lipschitz constant $L$             & Maximum gradient on discretized domain               \\
Safe EA and   & Population size $\mu$                        & Num. of initial safe seeds                         \\
    UnsafeEA  & Offspring  pop.~size $\lambda$ & Num. of initial safe seeds                         \\
              & Parental selection                           & Binary tournament                                    \\
              & Crossover operator                           & Uniform crossover                                    \\
              & Crossover prob.~$p_c$         & 0.8                                                  \\
              & Mutation operator                            & Gaussian mutation,\newline $\mu_m=0$, $\sigma_m=0.1$ \\
              & Mutation prob.~$p_m$          & 1/$d$                                                \\
              & Environment.~selection             & ($\mu$+$\lambda$)-ES                                 \\
        All   & \mbox{Function evaluations}                  & 100                                                  \\  \bottomrule
    \end{tabular}}
\end{table}

\smallskip{}
\noindent\textbf{Algorithm parameter settings. } Table~\ref{tab:tab2} summarizes the algorithm parameters and their values  used in the experimental study. Most of these settings are standard settings as suggested in the original papers of the respective algorithms. 
For VA and UnsafeEA, we set $\mu=\lambda=$ number of initial safe seed(s) provided. Both algorithms generate two offspring at a time by first using binary tournament selection to select two solutions from the current population, which then undergo uniform crossover ($p_c=0.8$) and Gaussian mutation ($p_m=1/d$, mean $\mu_m= 0$, standard deviation $\sigma_m= 0.1$); this process is repeated until $\lambda$ offspring are generated. The variation operator settings were selected based on preliminary experimentation to achieve robust results. 

With respect to safe GPs, to make \SafeOpt and \SafeUCB more practical, we have modified the method for computing the upper and lower confidence interval, and also updated the way the \Lipschitz constant $L$ is estimated (see the supplementary material for the technical details).


As mentioned previously, while the safe GP algorithms have an in-built method for handling noise, the safe EA  (VA) and unsafe ($\mu$ + $\lambda$)-ES do not. To be able to cope with noise meaningfully, we extend the two algorithms with a simple average scheme: within a single run, the algorithm keeps track of all previously evaluated solutions and their noisy objective function values so that any solution $\vec{x}$ that has been evaluated multiple times during the search will use the average of the multiple noisy evaluations as its fitness value. As a result, solutions duplicated in the population will have the same fitness value.

\subsection{Test Problems}
\begin{figure}[t]
     \centering%
     \begin{subfigure}[b]{0.483\columnwidth}
         \centering
         \includegraphics[width=\textwidth]{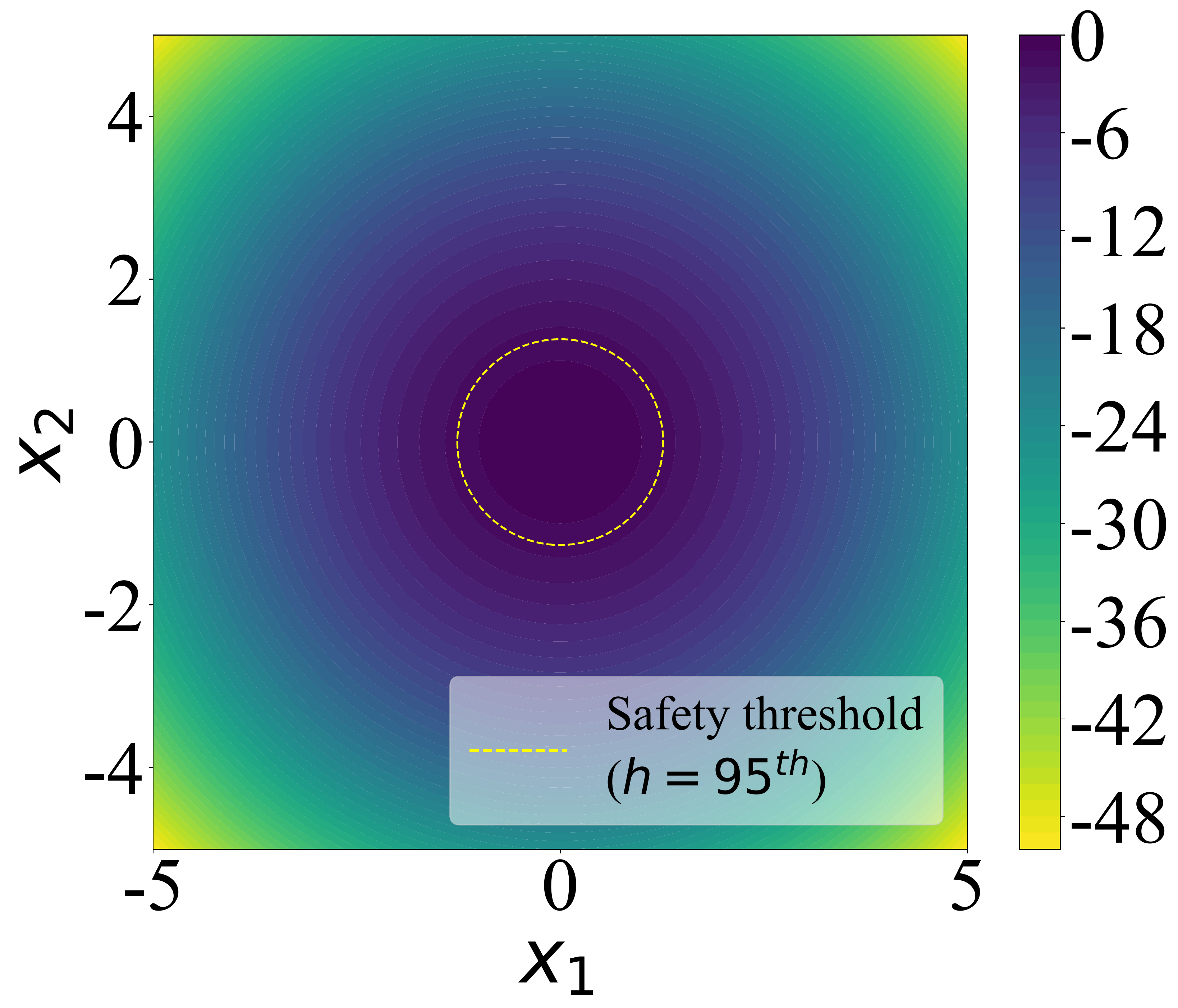}
         \caption{Sphere function}
         \label{fig:Sphere}
     \end{subfigure}
     \hfill%
     \begin{subfigure}[b]{0.497\columnwidth}
         \centering
         \includegraphics[width=\textwidth]{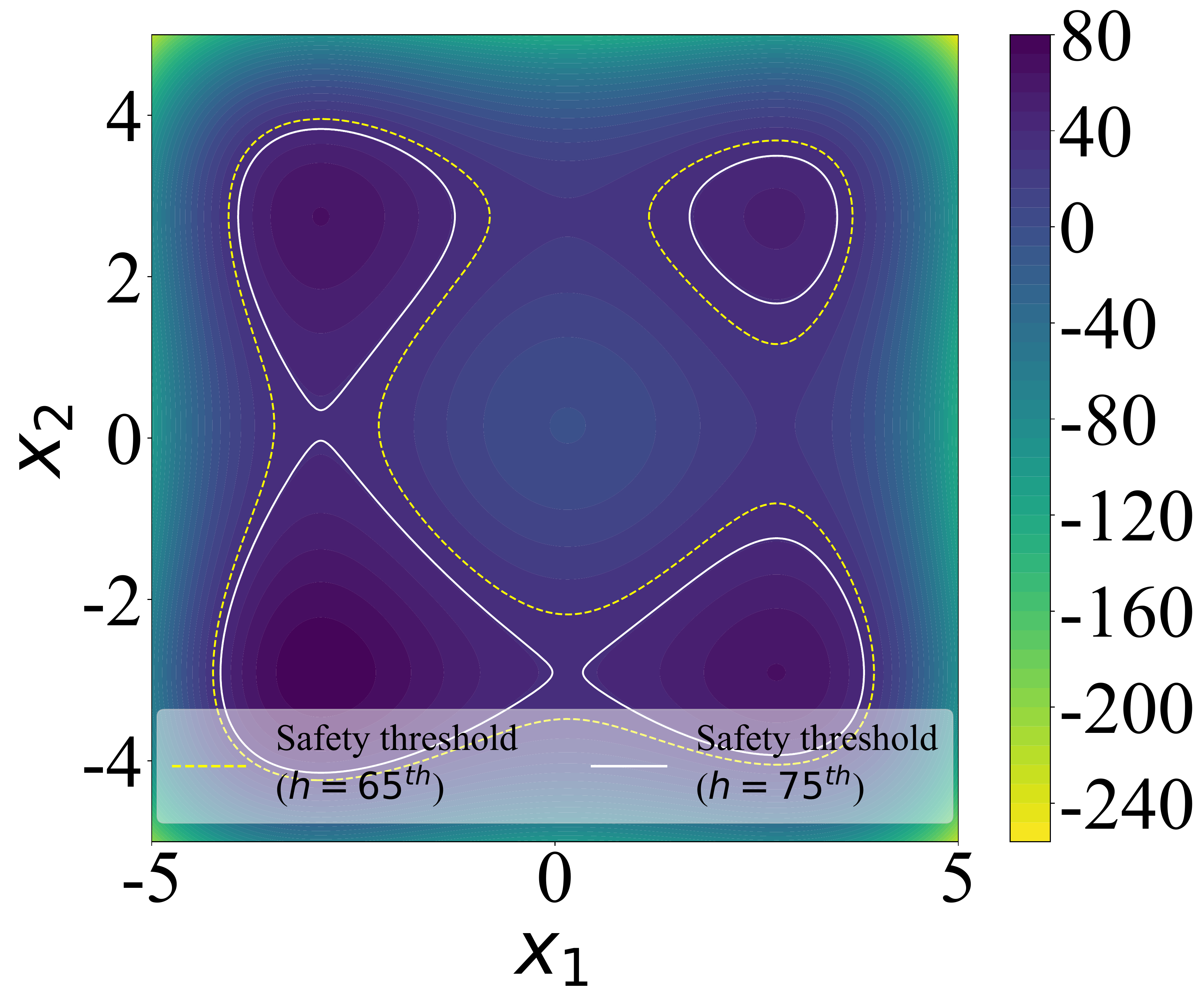}
         \caption{Styblinski-Tang function}
         \label{fig:stybtang}
     \end{subfigure}
        \caption{Contour plots showing the (noiseless) objective function values  in \mbox{$D\subset \mathbb{R}^2$} of (a) the Sphere and  (b) the Styblinski-Tang functions. For both functions we also show the safety thresholds $h$, which bound the safe regions in the search space, considered in the experimental study. The value of $h$ in (b) controls connectedness between safe search space regions centred around local optima. 
        }\label{fig:ObjFuncs}
\end{figure}

As the focus of this paper is on building up a fundamental understanding of how EAs and other algorithms behave on SafeOPs, we do a deep dive analysis of two test problems varying in modality, on which we augment the notion of safety: the unimodal Sphere function and the multimodal Styblinski-Tang function. These two functions will allow us to understand the impact of different problem settings on the search behaviour as well as the global optimization behavior of the algorithms (in the case of the Styblinski-Tang function). The two test problems are shown visually in Figure~\ref{fig:ObjFuncs} (for $d=2$) and defined formally in the following. 


The \textit{Sphere function}~\cite{hansen2009real} is defined as
\begin{equation}\label{eq:Sphere}
    f(\vec{x}) = f(\vec{x}^*) - ||\vec{x}_e||^2\;,
\end{equation}
where $\vec{x}^*$ is the location of the global maximum, $\vec{x}_e = \vec{x}-\vec{x}^*$, and $||\cdot ||$ denotes the Euclidean norm. 
We set $\vec{x}^*=(0,0)$ and $f(\vec{x}^*)=0$.

The \textit{Styblinski-Tang function}~\cite{ZubHidHaj2014} is defined as
\begin{equation}\label{eq:styblinski}
    f(\vec{x}) = -\left( \frac{1}{2} \sum_{k=1}^{d} {x_k^4 -16x_k^2 +5x_k}\right)\;,
\end{equation}
where $d$ is the number of decision variables, the location of the global maximum is $\vec{x^*}=(-2.903534, \cdots,$ $-2.903534)$, and $f(\vec{x^*})= 39.16599\times d$. This multimodal problem has three local maxima (one in bottom right, one in the top left, and one in the top right corner), and one global maximum (bottom left corner). 

Neither function has feasibility constraints. For ease of visualisation of the results, we set $d=2$ for both problems. The additive noise in Eq.~(\ref{eq:gd}) is randomly generated whenever an evaluation is made.

\subsection{Converting a Problem Into a SafeOP\label{sec:EStp}}
To convert an optimization problem without safety constraints into a SafeOP, one needs to decide on mechanisms to (i)~discretize the search space (as needed for safe GPs), (ii)~set the safety threshold $h$, and (iii)~generate the initial safe seed. We explain the mechanisms adopted here in turn. 

 

\smallskip{}
\noindent\textbf{Discretizing the search space.}  We set the search space for both test problems to $D=[-5,5]\times [-5,5]$. The search space is uniformly discretized into $500\times 500$ using inspiration from~\cite{SuiGotBur2015icml}.

\smallskip{}
\noindent\textbf{Setting the safety threshold.}
Recall that a solution with an objective function value below the safety threshold $h$ is deemed  unsafe, then the greater the value of $h$, the larger the portion of the search space deemed to be unsafe and thus the more constrained the search becomes. In this work, we set $h$ such that we can investigate different more or less challenging search scenarios. We set $h$ to be the $k\nth$ percentile of the objective values in the discretized search domain. For simplicity of notation, when we say $h=k\nth$ in the remainder of the paper, we mean that the safety threshold $h$ corresponds to this $k\nth$ percentile value, i.e., a setting of $h=95\nth$ means that $h$ corresponds to an objective value equal to 95\% of the objective values in the discretized domain. A lower percentile means that fewer solutions are unsafe. For the sphere function, the safety threshold does not change the modality of the problem (it remains unimodal regardless of $h$) but simply controls the available safe search space. This is different for the multimodal Styblinski-Tang function, which has three local and one global optima. Here we investigate two settings of $h$: a lower threshold value, which causes one local optimum to be isolated from the other local optima, i.e. the optimizer needs to traverse through or jump over an unsafe region in the search space to reach the isolated local optimum or escape from it; and a higher threshold value, resulting in all local optima being isolated from each other. The effect of the different safety thresholds (percentiles) on the two test problems is also shown in Figure~\ref{fig:stybtang}, and the safety thresholds considered in the study are shown in Table~\ref{tab:tab1}. 

\begin{table}[t]
    \caption{Problem settings used in the experimental study.}\label{tab:tab1}
    \centering%
    \begin{tabular}{L{6.45em}L{10.5em}L{5em}}
    \toprule
    \bf Test problem & \bf Problem parameter  & \bf Parameter setting\\\midrule
       Sphere   & Safety threshold $h$ & $95\nth$ \\[0.5em]
       Styblinski-Tang & Safety threshold $h$ & $\{ 65\nth, 75\nth \}$ \\[0.5em]
         Sphere and & Num. of initial safe seeds & $\{2, 10\}$\\
    Styblinsk-Tang    & Search space dimension $d$ & 2\\
        & Noise level $\sigma$ & 0.1 \\\bottomrule
    \end{tabular}
\end{table}

\smallskip{}
\noindent\textbf{Generating the initial safe seed.} In practice, the initial safe seeds would be known safe solutions to the problem at hand. To create safe seeds for our artificial SafeOPs, we sample the initial safe seeds from input points in the discretized domain that meet $f(\vec{x})-\sigma \cdot \beta \geq h$, where $\sigma$ and $\beta$ denote the standard deviation of the noise in Eq.~(\ref{eq:gd}) and the parameter for the confidence interval of the noise, respectively. We set $\beta=1.96$ to guarantee providing initial safe seeds that are safe from the noise (i.e., $y(\vec{x})\geq h$, Eq.~\ref{eq:gd}) to the algorithms for their initialization with 95\% confidence. The standard deviation of the noise in Eq.~(\ref{eq:gd}) should be statistically estimated in practice, however, for simulation we have the freedom to set the level of noise; thus, we assume that $\sigma = 0.1$. Hence, if we know $h$, then we can use the condition that $f(\vec{x})-\sigma \cdot \beta \geq h$ to sample an arbitrary number of initial safe seeds. We simulate the scenario where the optimizer is given 2 and 10 initial safe seeds, simulating a SafeOP for which a decision maker has small and larger a priori knowledge regarding safety, respectively (see Table~\ref{tab:tab1}); consequently, our EA-based algorithms use also a population size of 2 and 10 (this is smaller than we would normally use in an EA but we need to bear in mind that we also allow only 100 objective function evaluations in total). While the initial safe seeds are generated at random for each of the 20 algorithmic runs, the same 20 sets of seeds are used across the benchmarked algorithms. 



\smallskip{}
\noindent\textbf{Location of the initial safe seeds.}
In the case of the Styblinski-Tang function, we consider three different scenarios according to the location of the initial safe seeds (Figure~\ref{fig:stybtang}): in scenario 1, seeds are sampled only from the safe region of the top-right local optimum (within $x_1 > 0 \land x_2 > 0$), which is always isolated; in scenario 2, seeds are sampled from the safe region of only one of the top-left (within $x_1 < 0 \land x_2 > 0$) or bottom-right (within $x_1 > 0 \land x_2 < 0$)  local optima, which are connected to the safe region of the global optimum only when the safety threshold is low; and in scenario 3, half the seeds are sampled from the top-left region and the other half from the bottom-right. Scenario 1 is the most difficult and represents limited knowledge of the search space: algorithms not willing to risk an unsafe evaluation cannot find the global optimum. Scenario 2 is easier when the safety threshold is low: algorithms not willing to risk an unsafe evaluation can find the global optimum if they explore low quality regions. Finally, the multimodality of scenario 3 gives a hint to the algorithms that there may be a higher-quality local optima within the safe region.



\section{Results\label{sec:Results}}

This section investigates the performance and search behaviour of the safe EA (VA), the four safeGPs, and the UnsafeEA for different problem parameters: locations of initial safe seeds, number of initial safe seeds, level of safety threshold, and safety budget. In the plots shown in this section and supplementary material, we will refer to modified \SafeOpt~\cite{BerSchKra2016safe} and modified \SafeUCB~\cite{BerSchKra2016safe} as M\SafeOpt, and M\SafeUCB, respectively. 

For the analysis of the results we keep track of the best-so-far objective function (BSF) value and total number of unsafe solutions evaluated at each evaluation step by an algorithm. Also, while the algorithms only observe the noisy objective function value $y(\vec{x})$, for the visualisations we will be using the corresponding true objective function value $f(\vec{x})$. If not stated otherwise, the results shown have been obtained across 20 independent algorithmic runs. 

An open-source repository containing object-oriented Python code for generating SafeOPs with different features (including the ones investigated here) and methods for the visualization can be downloaded from \url{https://github.com/YoungminKim93/SOAB}.

\begin{figure}[tb]
     \centering
     \begin{subfigure}[b]{0.49\columnwidth}
         \centering
         \includegraphics[width=\textwidth]{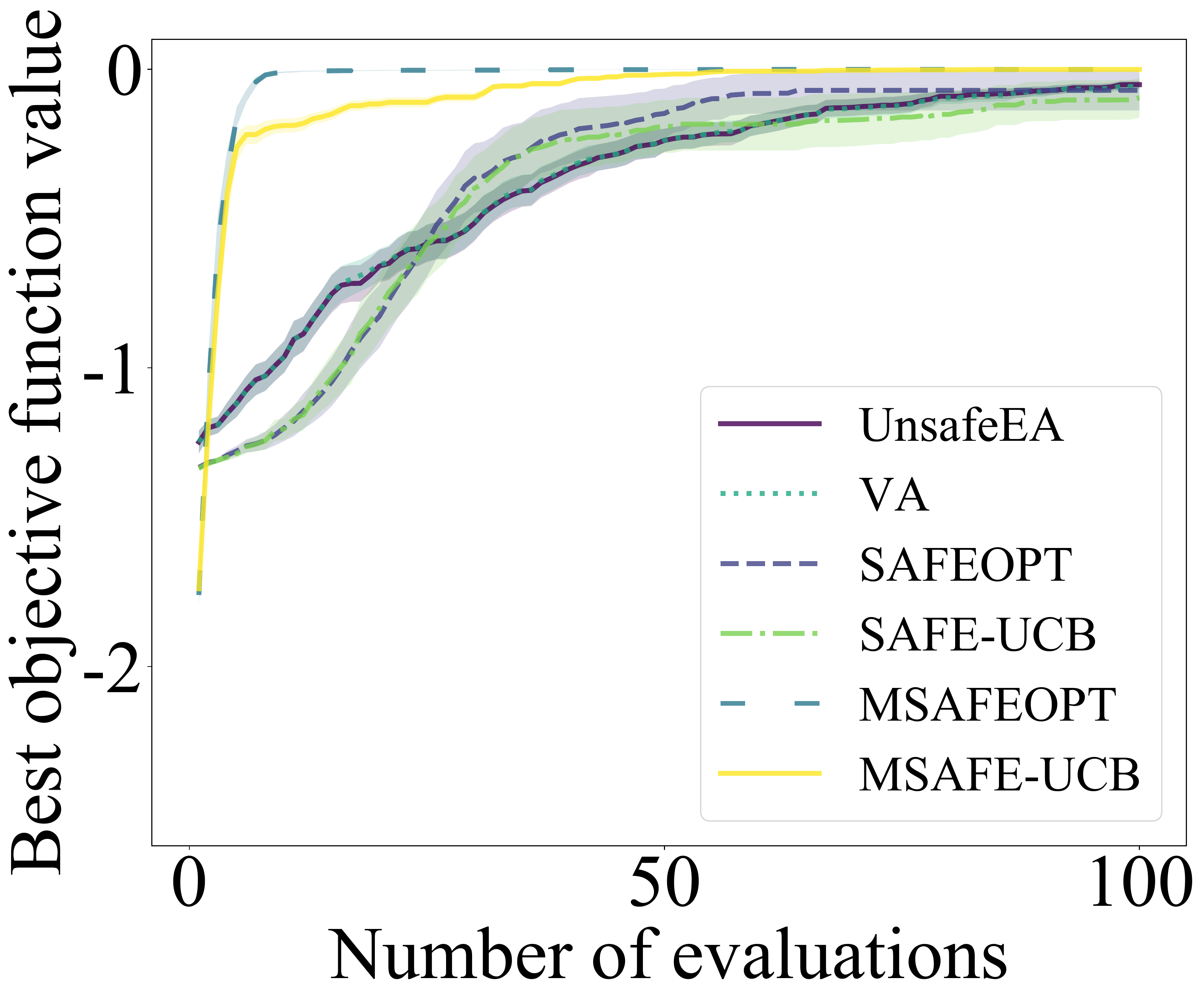}
         \caption{2 initial safe seeds}
         \label{fig:Spherenuminit2BSF}
     \end{subfigure}%
     \hfill%
     \begin{subfigure}[b]{0.49\columnwidth}
         \centering
         \includegraphics[width=\textwidth]{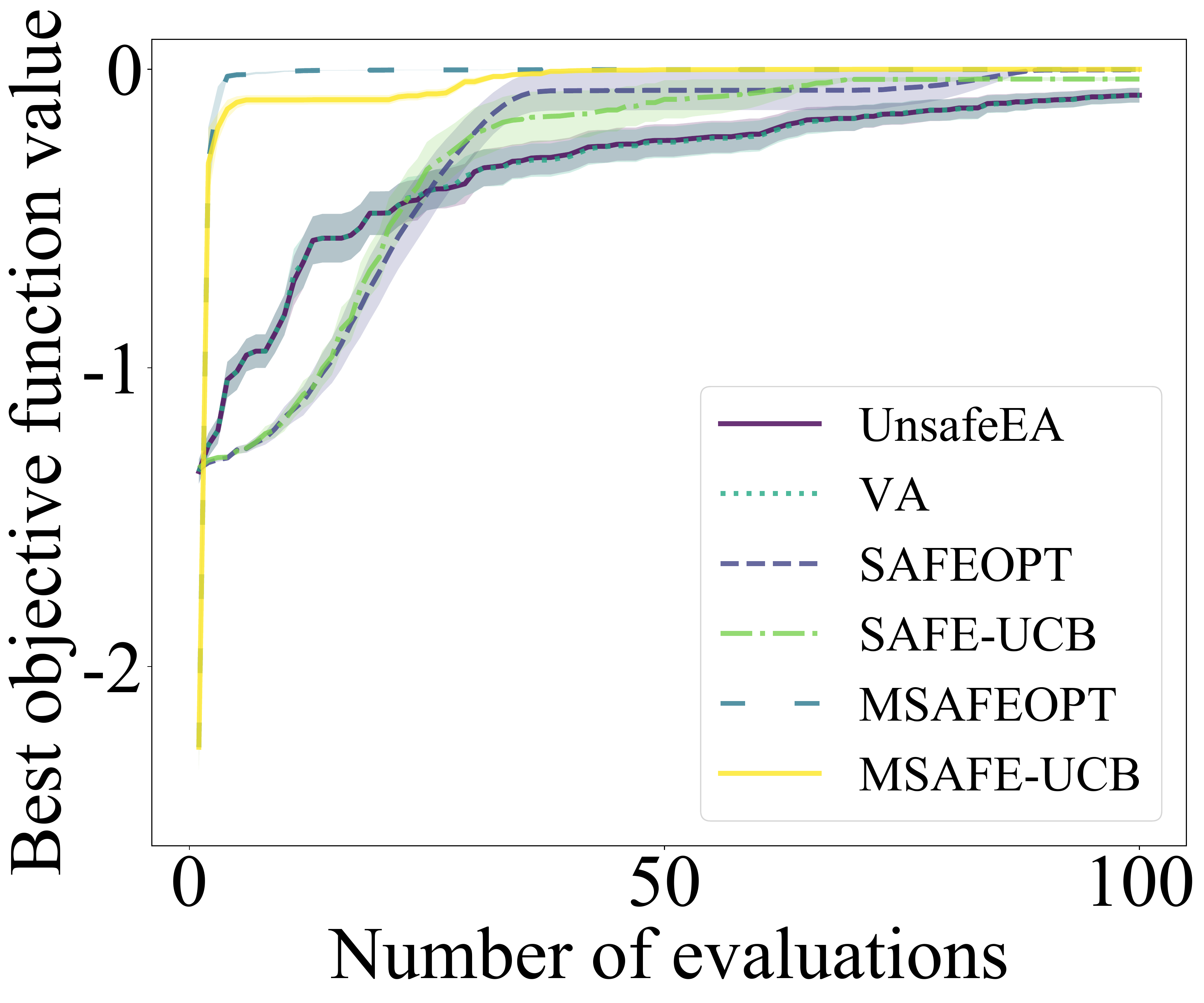}
         \caption{10 initial safe seeds}
         \label{fig:Spherenuminit10BSF}
     \end{subfigure}
     \begin{subfigure}[c]{0.49\columnwidth}
         \centering
         \includegraphics[width=\textwidth]{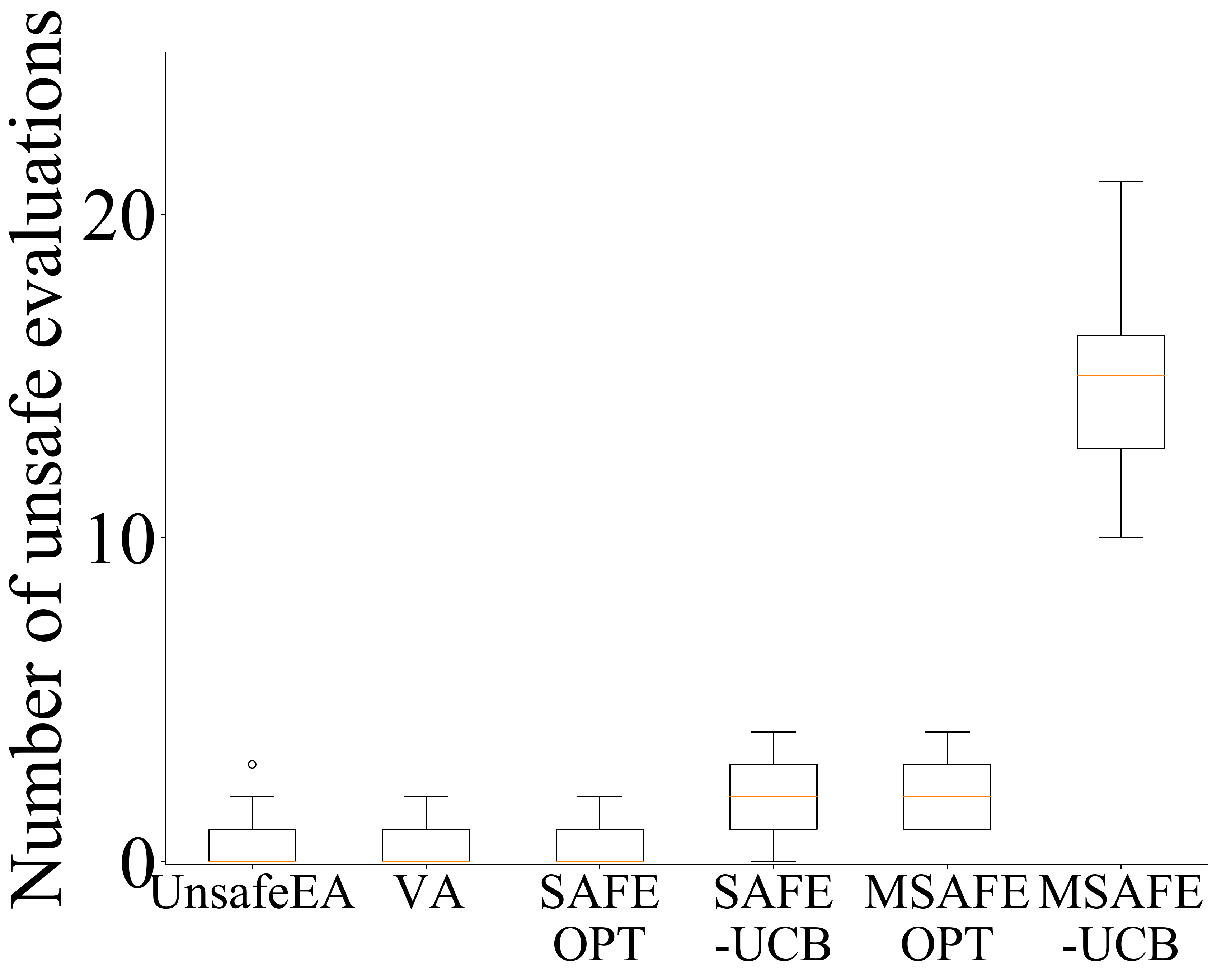}
         \caption{2 initial safe seeds}
         \label{fig:Spherenuminit2Unsafe}
     \end{subfigure}
     \hfill
     \begin{subfigure}[d]{0.49\columnwidth}
         \centering
         \includegraphics[width=\textwidth]{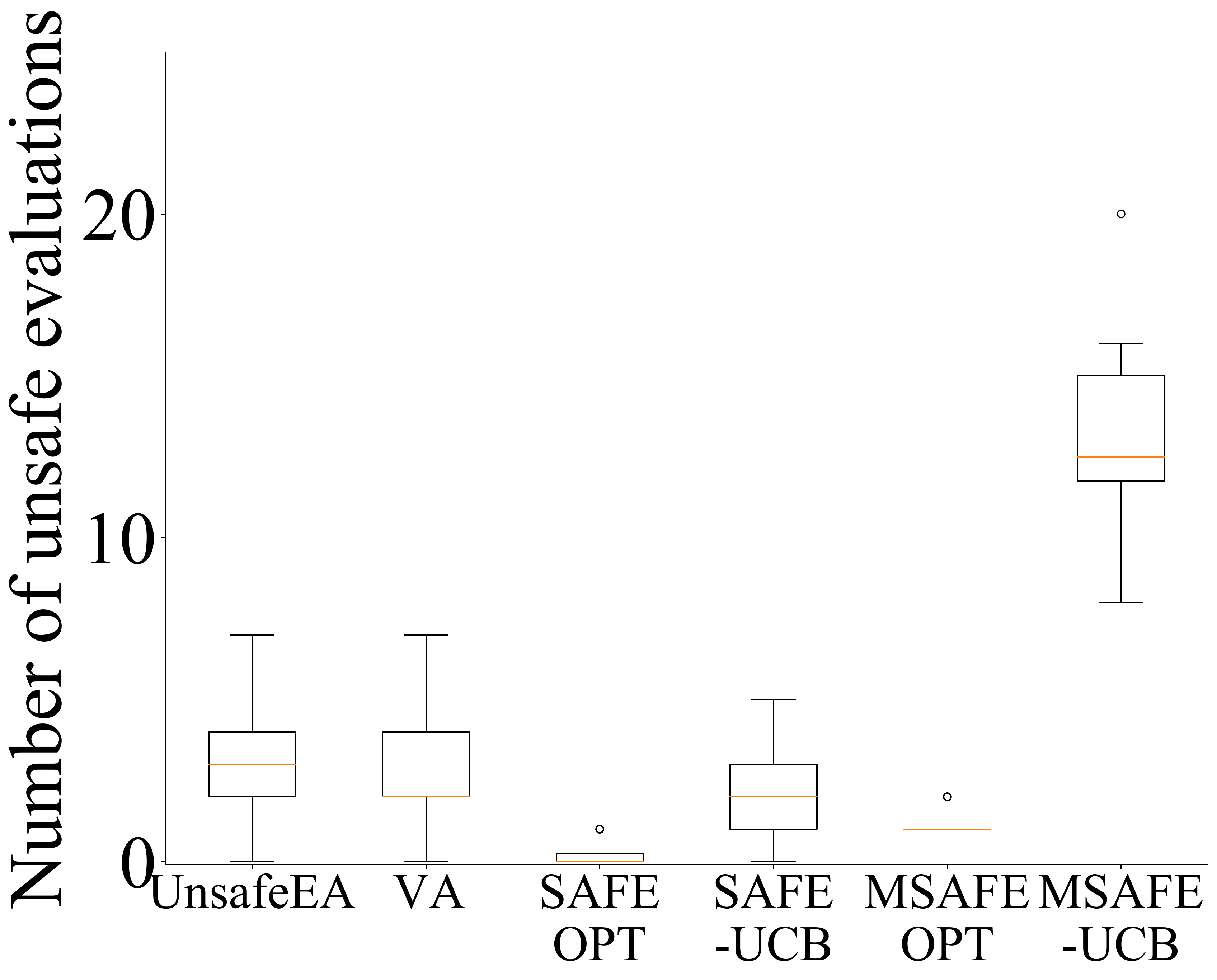}
         \caption{10 initial safe seeds}
         \label{fig:Spherenuminit10Unsafe}
     \end{subfigure} 
        \caption{Plots show (top row) the mean best objective function value (BSF) and standard error as function of the number of function evaluations  and (bottom row) distribution of the number of unsafe solutions evaluated across 20 algorithmic runs,  obtained by the various algorithms for the Sphere function (\mbox{$h=95\nth$}). 
        }\label{fig:InitSafe}
\end{figure}

\subsection{Impact of the Number of Initial Safe Seeds\label{sec:ImpactNumInit}}

Since the Sphere function is a unimodal function, we do not need to account for any search biases induced by local optima or other landscape features. 

Figure~\ref{fig:InitSafe} shows the BSF and number of unsafe solutions evaluated for the different algorithms for two initial safe seed settings (2 vs 10). In general, we observe that increasing the number of initial safe seeds is beneficial in terms of converging to an optimum (top row plots) more quickly and more robustly (as indicated by the slightly lower standard error for 10 initial safe seeds). However, we can see that more initial safe seeds (bottom row of plots) results in more unsafe solutions being evaluated by the two EA-based methods (VA and UnsafeEA), while it has the opposite effect on the safeGPs except for \SafeUCB. The pattern of the EA-based methods is explained by the fact that for an initial safe seed of 2, we also have a population size of 2, and therefore there is a limited exploration power by the two algorithms reducing the chance of evaluating an unsafe solution. On the contrary, having more initial safe seeds allows the safeGPs to build a more accurate model of the search landscape and hence be more certain about the safety status of a solution, leading to fewer unsafe solutions being evaluated. 

The similar performance of VA and UnsafeEA for the two initial safe seed settings may imply that, in an environment where unsafe solutions are generally less likely to be evaluated, it is difficult to estimate the safety level of a solution by looking at the safety status of the nearest neighbor (as done by VA). However, VA shows, in general, better performance in terms of the number of unsafe evaluations (at least equal but in most cases slightly better, see additional figures in the supplementary material).

It is expected that the two algorithms, modified \SafeOpt and modified \SafeUCB, outperform the original counterparts in terms of BSF at the cost of evaluating more unsafe solutions. This is because the modified versions are set up to be more explorative/risky, as they do not rely on $L$ (Eq.~\ref{noL}).

\subsection{Impact of the Level of Safety Threshold\label{sec:ResultsSafeThres}}

\begin{figure}[!tb]
  \centering
      \begin{subfigure}{0.49\columnwidth}
         \centering
         \includegraphics[width=\textwidth]{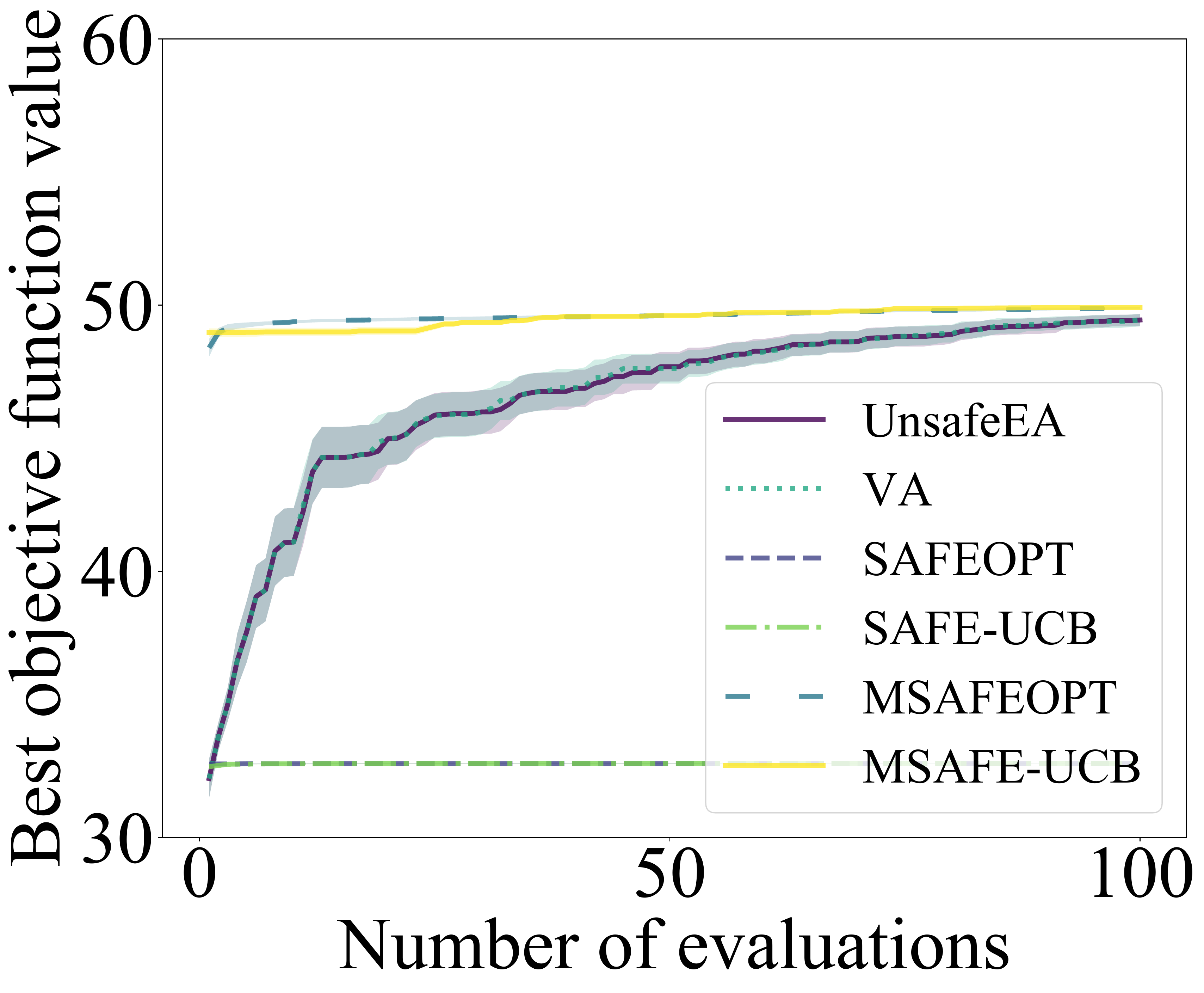}
         \caption{\mbox{$h=65\nth$}}
         \label{fig:isolated65}
     \end{subfigure}\hfill%
     \begin{subfigure}{0.49\columnwidth}
         \centering
         \includegraphics[width=\textwidth]{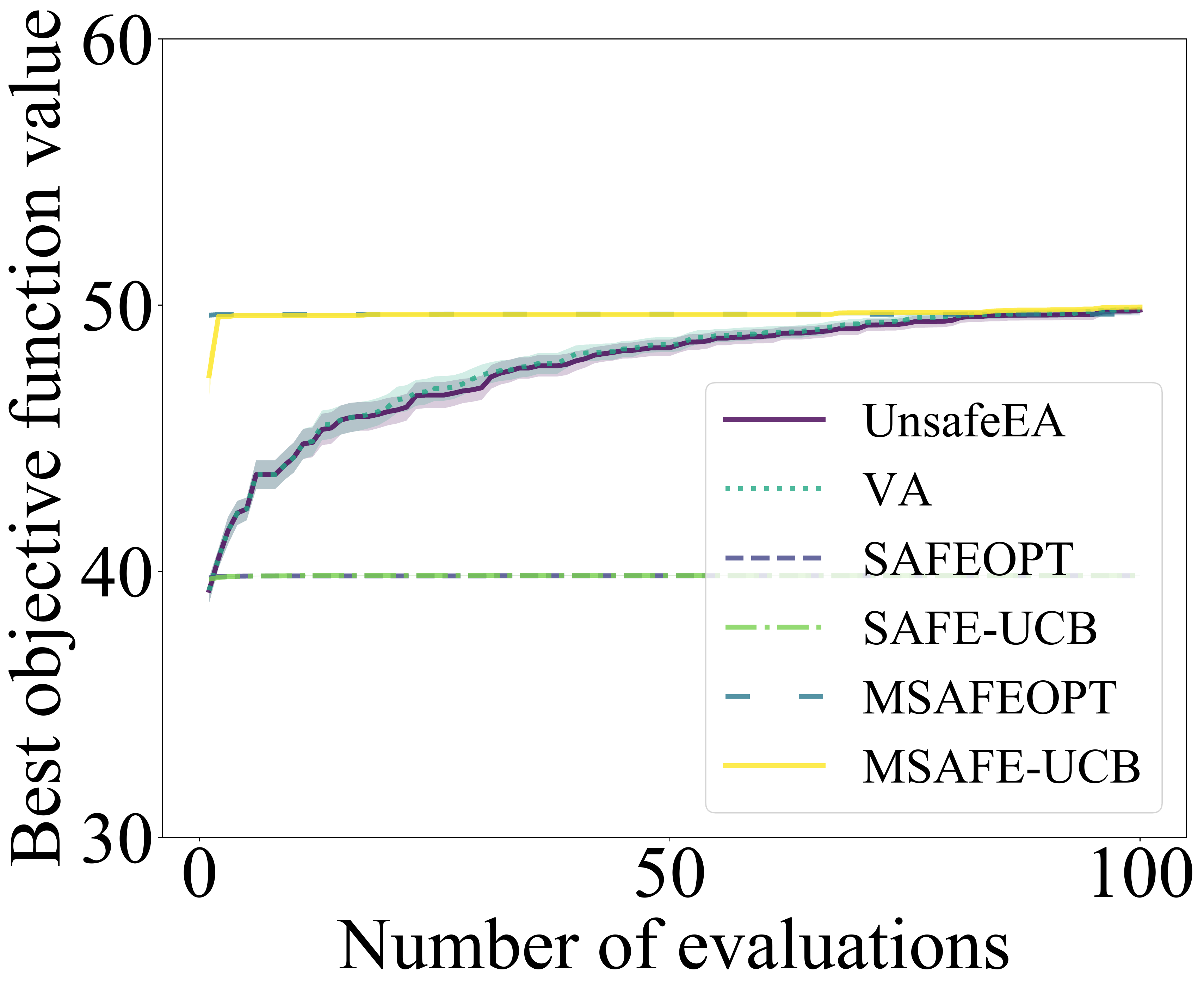}
         \caption{\mbox{$h=75\nth$}}
         \label{fig:isolated75}
     \end{subfigure}
        \caption{Mean best objective function value (BSF) and standard error as function of the number of function evaluations across 20 algorithmic runs obtained for the various algorithms for the Styblinski-Tang function using ten initial safe seeds. The algorithms were initialized with scenario 1. With \mbox{$h=65\nth$}, 3 out of 4 local optima fall in the connected safe search space region, whereas all local optima are in disconnected safe search space regions when \mbox{$h=75\nth$}.}
        \label{fig:Isolatedstyb}
\end{figure}

Let us now have a closer look at the impact of the safety threshold $h$ on performance and search behaviour. 

Considering the Styblinski-Tang function, when the algorithms are initialized with scenario 1, none of the algorithms performed global optimization because they were not able to jump out of that local safe region regardless of the level of safety threshold (see Figure~\ref{fig:Isolatedstyb}). However, an extreme behavior was observed for scenario 2 when \mbox{$h=65\nth$}  (see supplementary material). Modified \SafeUCB~\cite{BerSchKra2016safe} performed global optimization (it discovered solutions of quality better than the local optima around which the initial safe seeds were placed) achieving great BSF value, but produced a lot of unsafe evaluations. However, it did not show any potential for global optimization when the level of safety threshold increased. On the other hand, the other algorithms did not perform global optimization in scenario 2 regardless of the level of safety threshold. A similar behavior was observed for scenario 3 when ten initial safe seeds are provided, that is, extreme behavior of modified \SafeUCB was observed when \mbox{$h=65\nth$}, but it did not perform global optimization when $h=75\nth$. However, in this scenario, we observe the potential of global optimization of VA.


\begin{figure}
     \centering
     \begin{subfigure}{0.49\columnwidth}
         \centering
         \includegraphics[width=\textwidth]{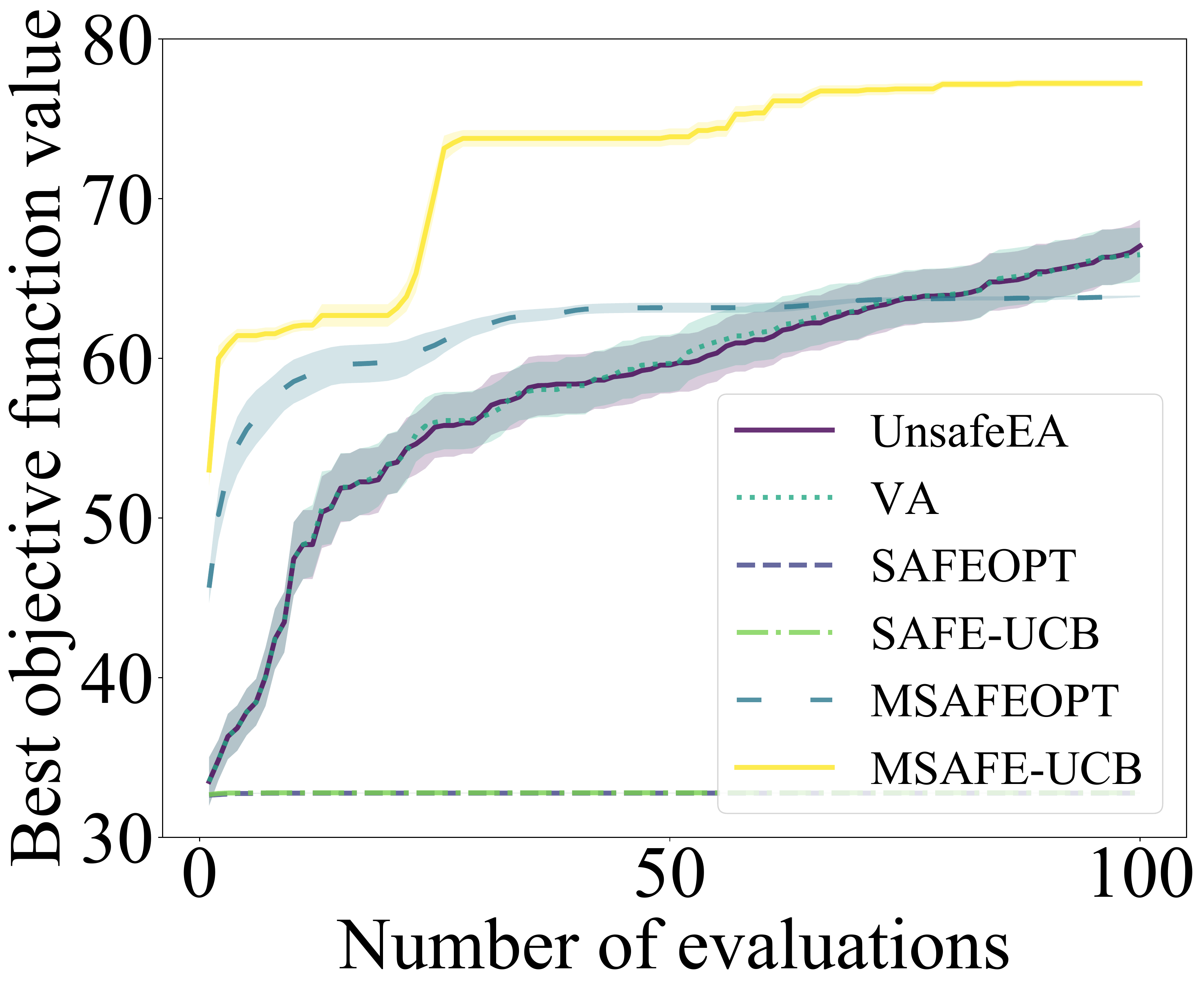}
         \caption{\mbox{$h=65\nth$}}
         \label{fig:StybSafeThresBSF65}
     \end{subfigure}
     \hfill
     \begin{subfigure}{0.49\columnwidth}
         \centering
         \includegraphics[width=\textwidth]{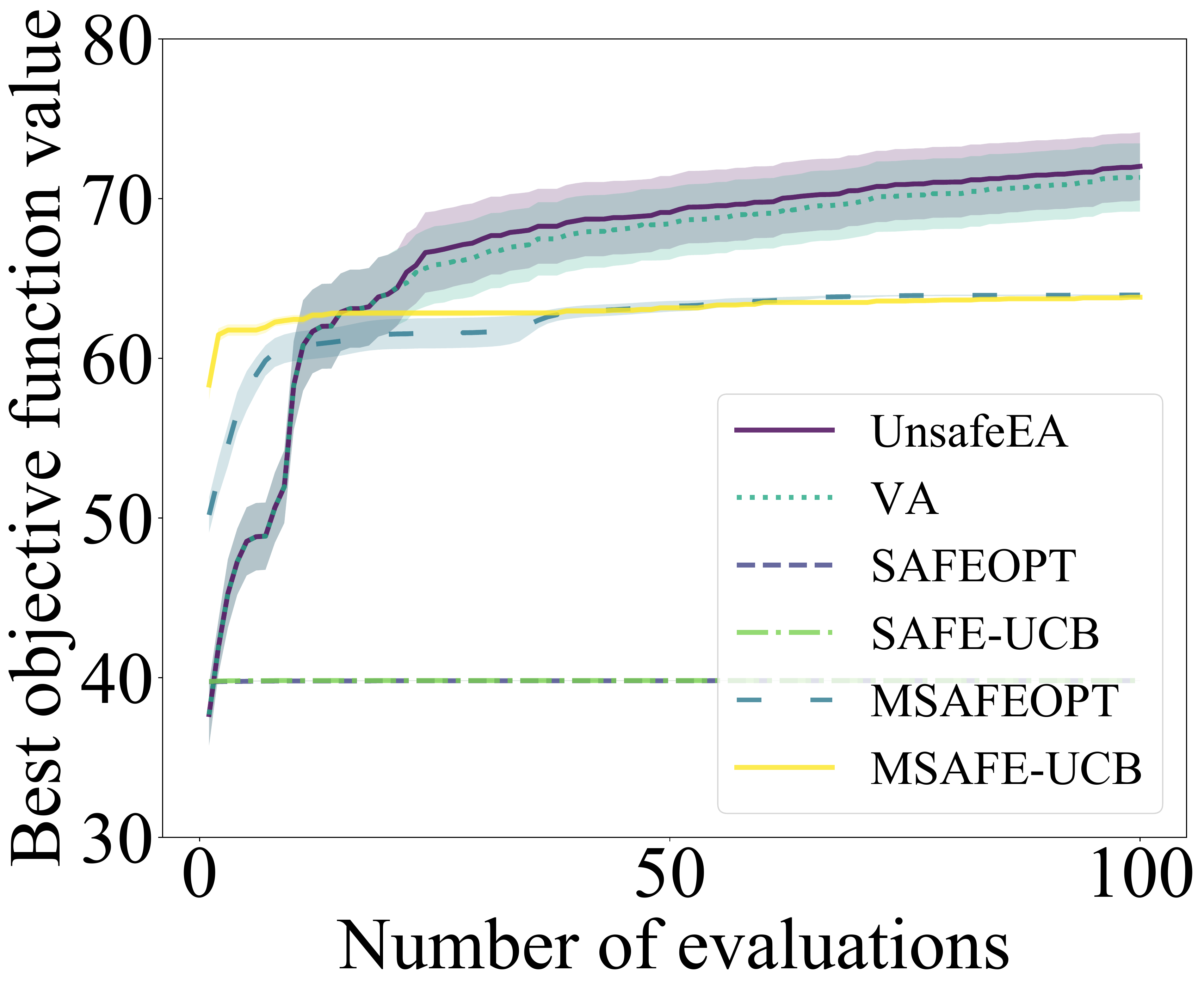}
         \caption{\mbox{$h=75\nth$}}
         \label{fig:StybSafeThresBSF75}
     \end{subfigure}
     \begin{subfigure}{0.49\columnwidth}
         \centering
         \includegraphics[width=\textwidth]{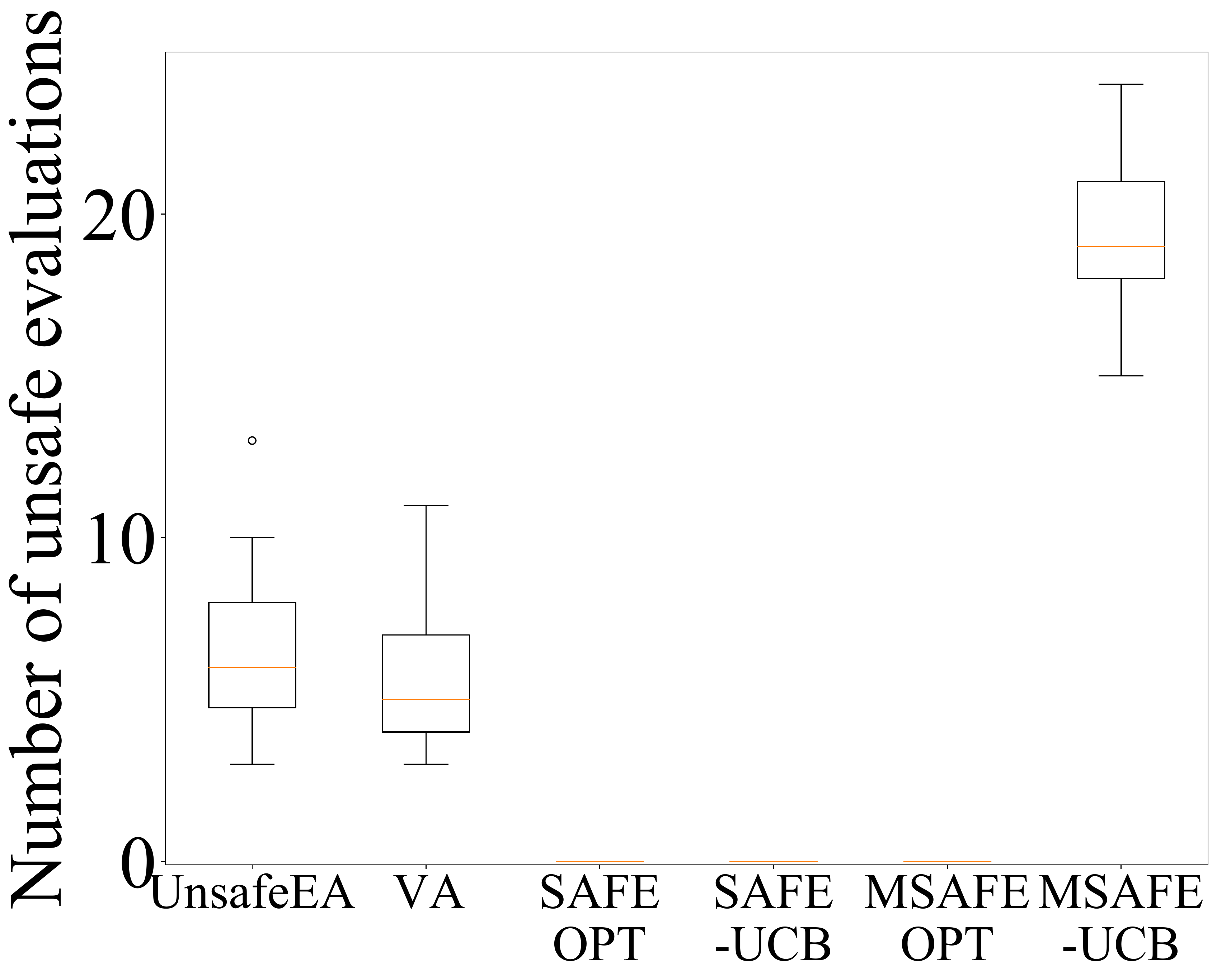}
         \caption{\mbox{$h=65\nth$}}
         \label{fig:StybSafeThresUnsafe65}
     \end{subfigure}
     \hfill
     \begin{subfigure}{0.49\columnwidth}
         \centering
         \includegraphics[width=\textwidth]{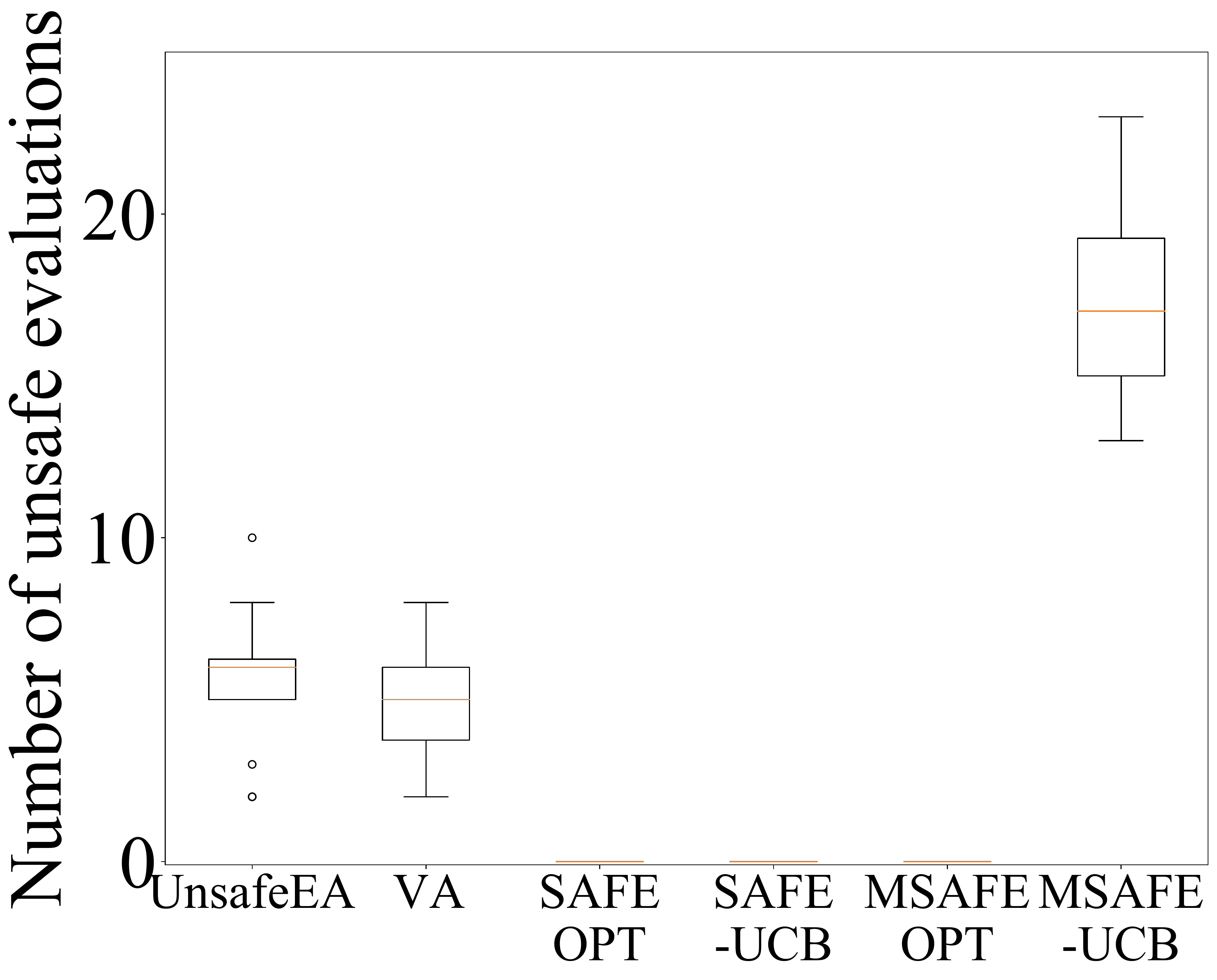}
         \caption{\mbox{$h=75\nth$}}
         \label{fig:StybSafeThresUnsafe}
     \end{subfigure}
        \caption{Plots show (top row) the mean best objective function value (BSF) and standard error as function of the number of function evaluations and (bottom row) distribution of the number of unsafe solutions evaluated across 20 algorithmic runs obtained by the various algorithms on the Styblinski-Tang function using ten initial safe seeds. Here, the initial safe seeds are sampled with scenario 3. With \mbox{$h=65\nth$}, 3 out of 4 local optima fall in the connected safe search space region, whereas all local optima are in disconnected safe search space regions when \mbox{$h=75\nth$}.
        }
        \label{fig:SafeThres}
\end{figure}

Figure~\ref{fig:SafeThres} shows BSF and number of unsafe solutions evaluated for the Styblinski-Tang function using two different safety thresholds where the algorithms are initialized with ten initial safe seeds. We observe that increasing the safety threshold level $h$ (i.e. increasing the presence of unsafe solutions) improves the BSF discovered by VA and UnsafeEA, while reducing the likelihood for evaluating unsafe solutions. This may happen as VA is initialized with initial safe seeds whose output values are significantly greater than the initial safe seeds sampled when $h=65\nth$. Also, this implies that connectivity of the local optima is not a necessary condition for global optimization of VA, and having diverse information about the safe regions around the global optimum is likely to help global optimization. For example, all local optima are isolated with each other when $h=75\nth$, but VA is able to perform global optimization when it is initialized at both local optima which were connected to the global optimum when $h=65\nth$. On the other hand, connectivity may be interpreted as the key factor for global optimization of modified \SafeUCB because it performed global optimization when $h=65\nth$, while producing almost equal number of unsafe evaluations to those produced when $h=75\nth$, which is a test problem where the algorithm was unable to perform global optimization.
An interesting finding is that, while \SafeOpt and \SafeUCB performed well in terms of BSF and unsafe solutions evaluated for the Sphere function, the two algorithms are not able to improve the BSF significantly following initialization on Styblinski-Tang function. It may happen due to the limitation of estimating the \Lipschitz constant (Eq.~\ref{eq:lip}). We estimated the \Lipschitz constant over the whole search space (see supplementary material), and as shown in Figure~\ref{fig:stybtang}, the gradient around the edge of the search space is much greater than the gradient in the safe regions, thus resulting in highly risk-averse attitude of \SafeOpt and \SafeUCB. We also observe that modified \SafeOpt is not evaluating any unsafe solutions when compared to the Sphere function regardless of the value of $h$. 
\begin{figure}
     \centering
     \begin{subfigure}{0.49\columnwidth}
         \centering
         \includegraphics[width=\textwidth]{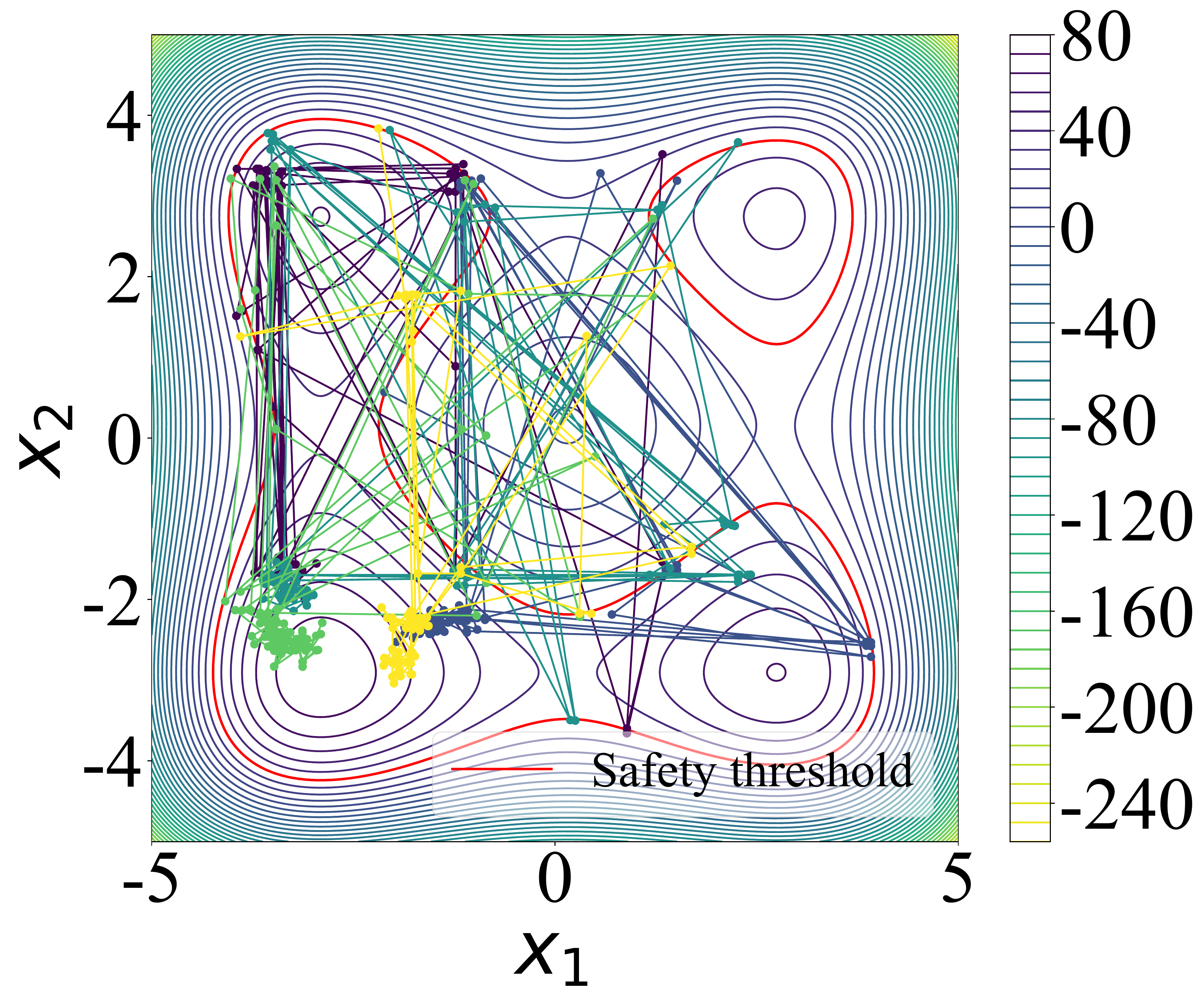}
         \caption{UnsafeEA}
         \label{fig:trajEA}
     \end{subfigure}
     \hfill
     \begin{subfigure}{0.49\columnwidth}
         \centering
         \includegraphics[width=\textwidth]{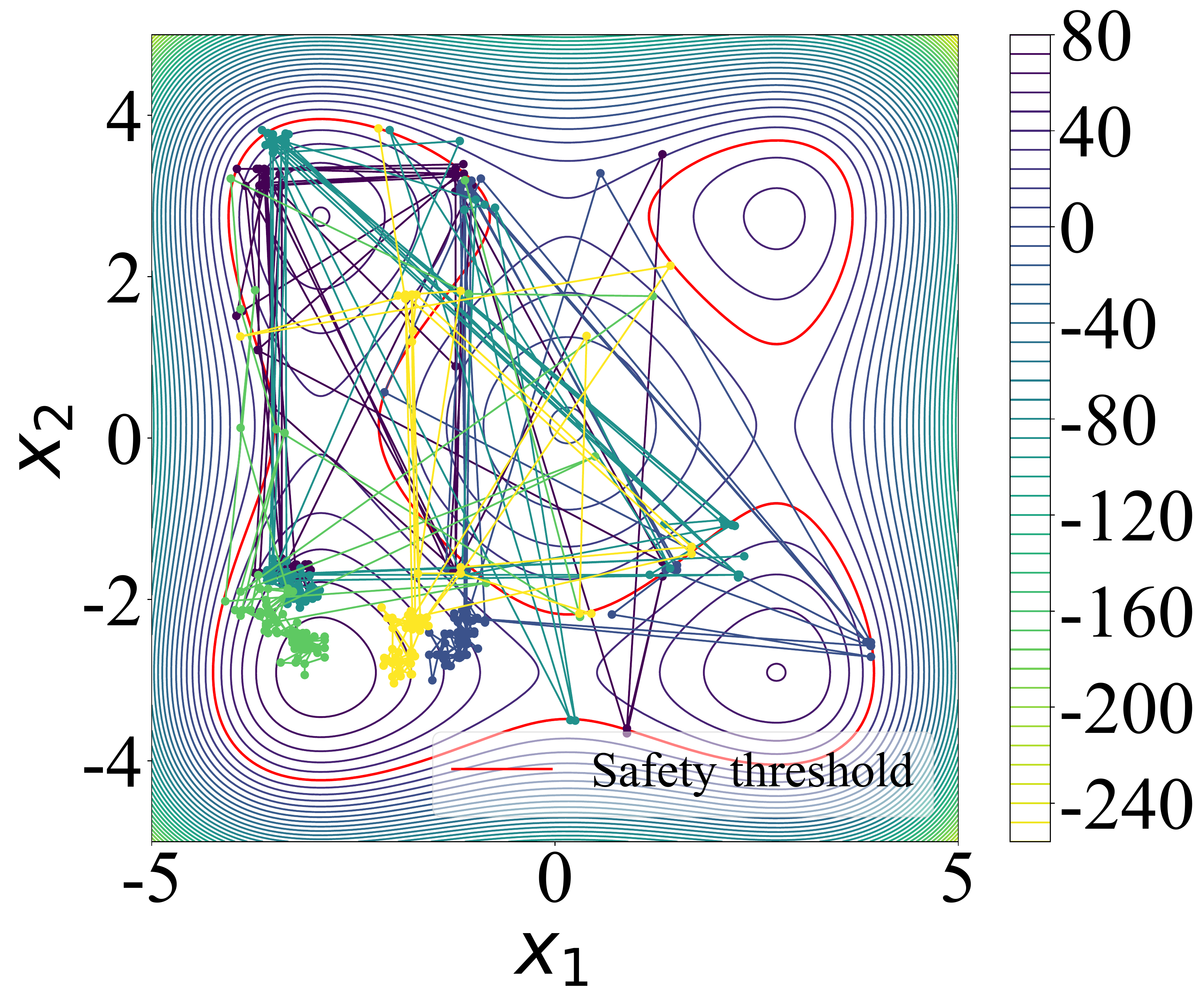}
         \caption{VA}
         \label{fig:trajVA}
     \end{subfigure}
     \begin{subfigure}{0.49\columnwidth}
         \centering
         \includegraphics[width=\textwidth]{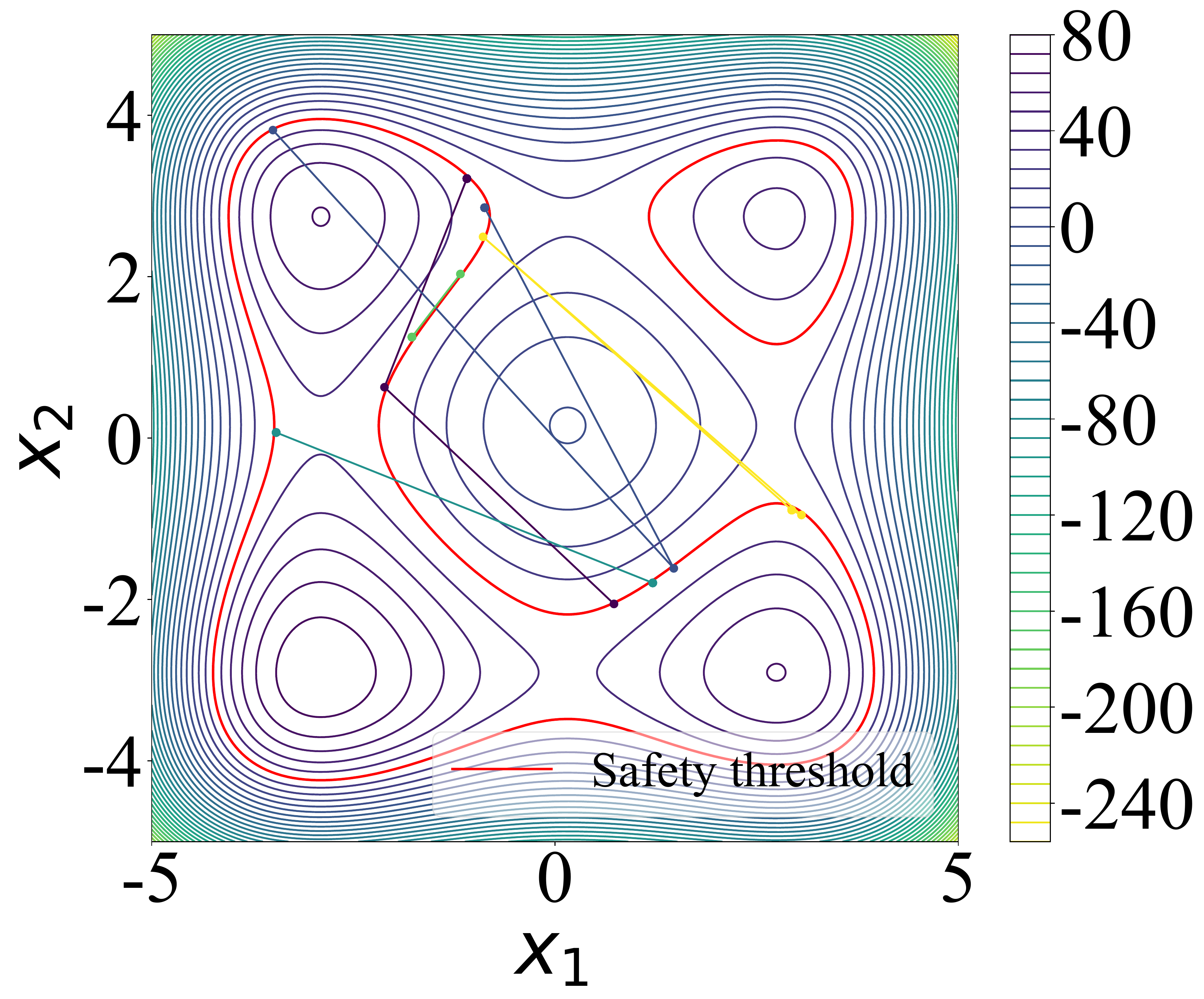}
         \caption{\SafeOpt}
         \label{fig:trajsafeopt}
     \end{subfigure}
     \hfill
     \begin{subfigure}{0.49\columnwidth}
         \centering
         \includegraphics[width=\textwidth]{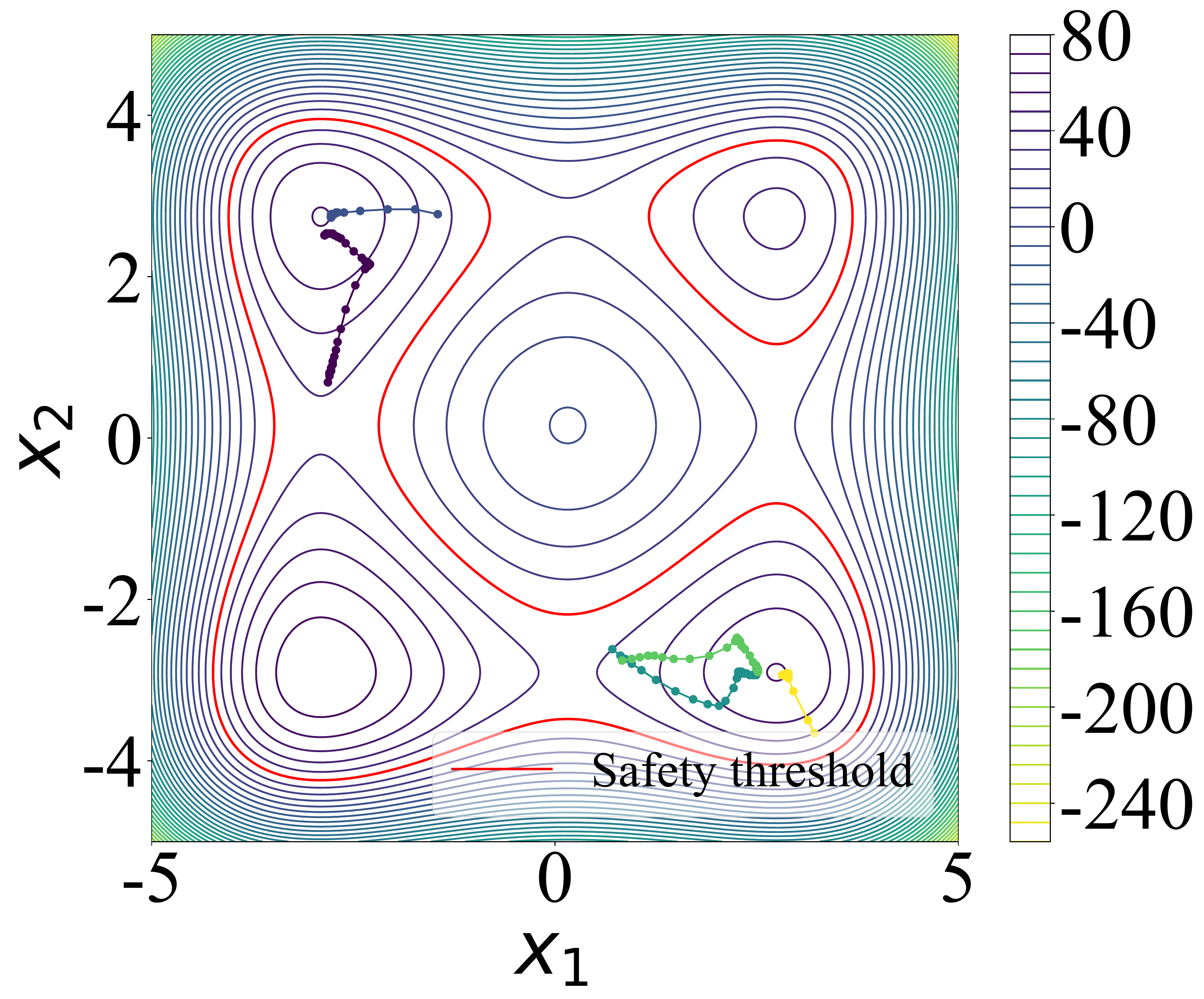}
         \caption{Modified \SafeOpt}
         \label{fig:trajmsafeopt}
     \end{subfigure}
     \begin{subfigure}{0.49\columnwidth}
         \centering
         \includegraphics[width=\textwidth]{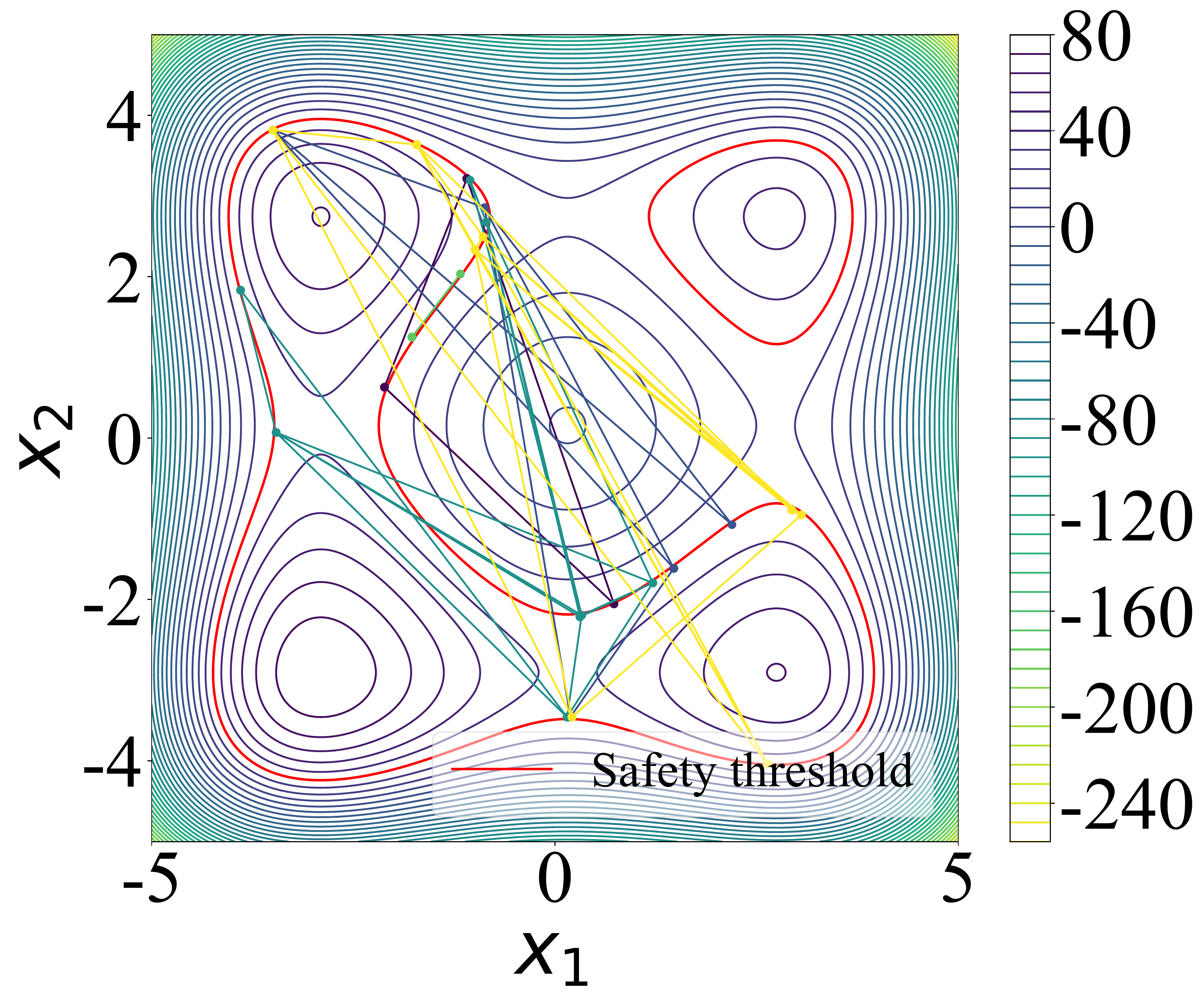}
         \caption{\SafeUCB}
         \label{fig:trajsafeucb}
     \end{subfigure}
     \hfill
     \begin{subfigure}{0.49\columnwidth}
         \centering
         \includegraphics[width=\textwidth]{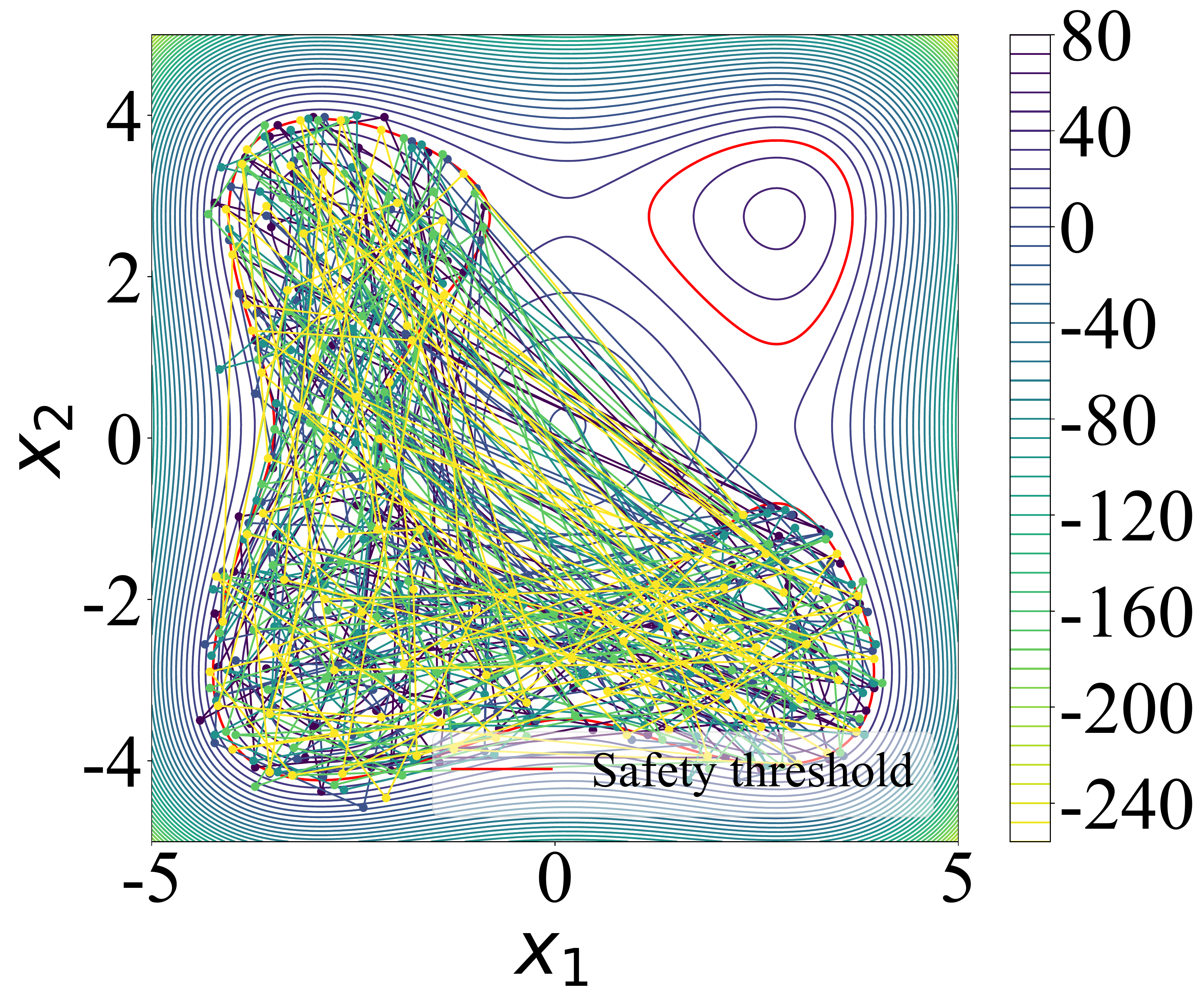}
         \caption{Modified \SafeUCB}
         \label{fig:trajmsafeucb}
     \end{subfigure}
        \caption{Trajectories of the algorithms when initialized with ten initial safe seeds sampled with scenario 3 (\mbox{$h=65\nth$}). Points connected with a line represent solutions generated in sequence within the same run. Each run evaluated 100 points and each color represents a different run out of 5 independent runs shown.}
        \label{fig:traj}
\end{figure}

Figure~\ref{fig:traj} provides the search trajectory of the first five runs of each algorithm initialized with ten initial safe seeds sampled with scenario 3 where $h=65\nth$. When we investigate the trajectories of EA-based optimizers, we can see that they use the information around the two local optima actively, e.g., the optimizers traverse the two optima actively and finally explore and exploit the global optimum, as shown in Figures~\ref{fig:trajEA} and~\ref{fig:trajVA}. Trajectories of modified \SafeUCB, which is the other algorithm that performed global optimization, show that the algorithm explored the safe region very actively. However, when we see the other algorithm that performed well (at least not stuck at a low BSF value), modified \SafeOpt quickly converges to one local optimum. Thus, the information given in initial safe seeds about safety in the two local optima is not fully used. What is worse is that \SafeOpt and \SafeUCB are stuck at particular input points whose quality of solution is significantly low.

As presented in Figure~\ref{fig:SafeThres}, when comparing the results of the EA and GP-based optimizers, we note that the EA-based optimizers (both VA and UnsafeEA) are doing significantly better than the GP-based optimizers in terms of BSF as the safety threshold increases (i.e. more of the search spaces becomes unsafe). We do not observe negative impact in terms of the number of unsafe evaluations along the increase of the level of safety threshold, and EA-based optimizers produce more unsafe evaluations than GP-based optimizers overall (except for modified \SafeUCB that showed extreme behavior). This is due to the combination of having few initial safe seeds combined with the drive of EA to explore more globally. 

\subsection{Budget of Unsafe Evaluations}

\begin{figure}
  \begin{minipage}{1.0\linewidth}
     \centering
     \begin{subfigure}[b]{0.49\columnwidth}
         \centering
         \includegraphics[width=\textwidth]{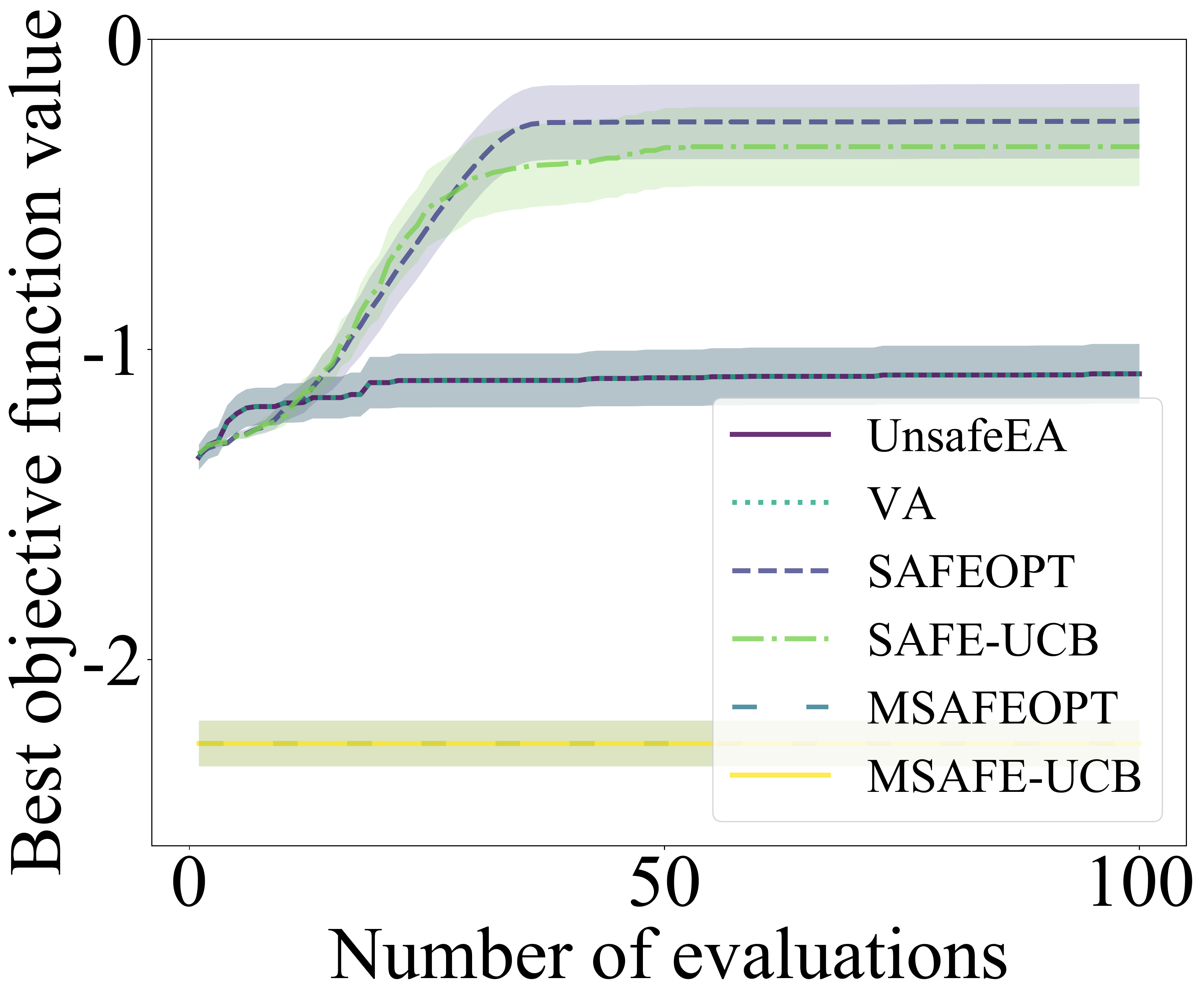}
         \caption{Sphere function}
         \label{fig:Spherezero}
     \end{subfigure}
     \hfill
     \begin{subfigure}[b]{0.49\columnwidth}
         \centering
         \includegraphics[width=\textwidth]{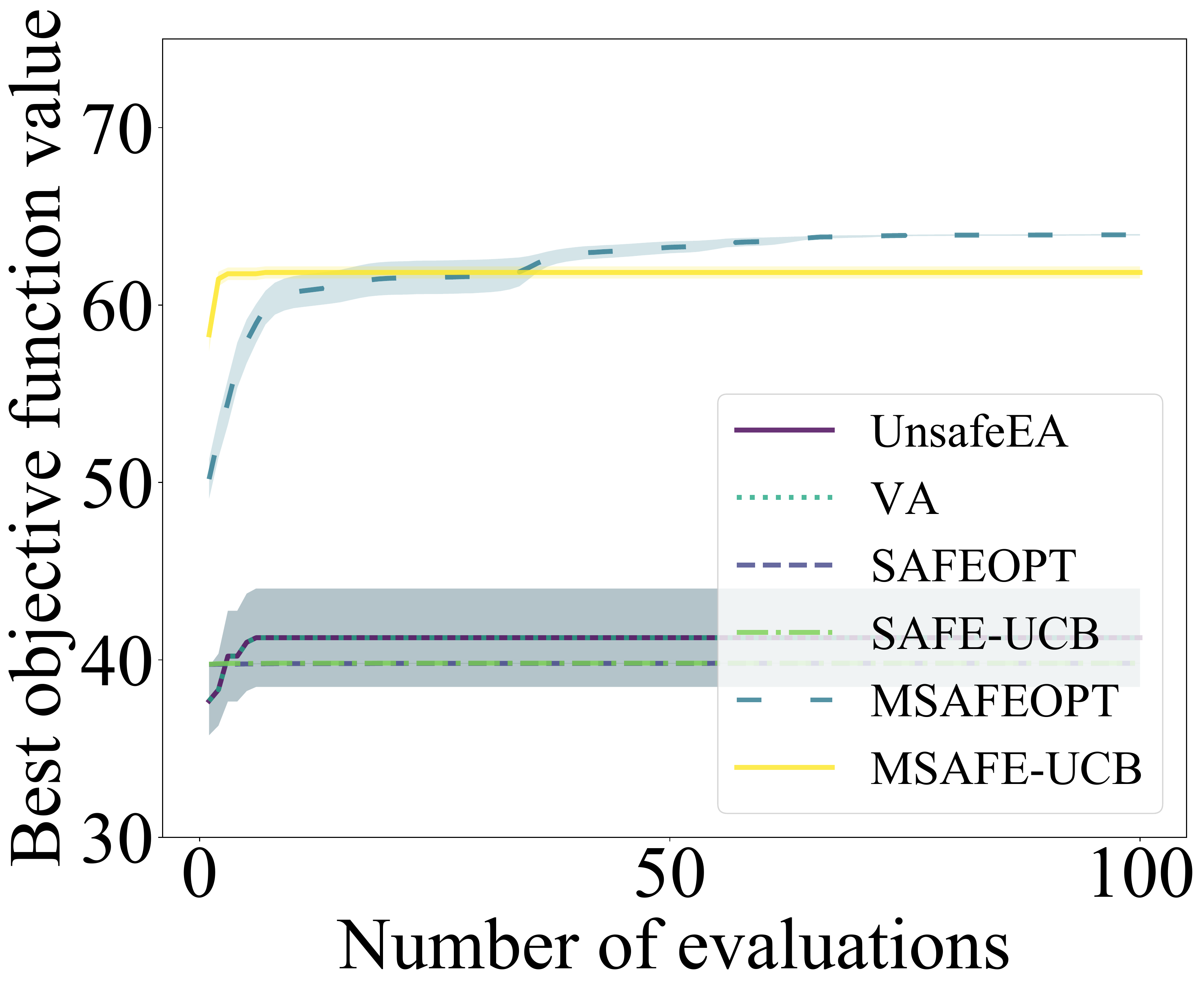}
         \caption{Styblinski-Tang function}
         \label{fig:Stybzero}
       \end{subfigure}
   \caption{Mean best objective function value (BSF) and standard error across 20 algorithmic runs over the number of function evaluations with  10 initial safe seeds, zero budget of unsafe evaluations, and \mbox{$h=95\nth$} for Sphere (\mbox{$h=75\nth$} for  Styblinski-Tang).}
        \label{fig:zerobudget}
  \end{minipage}
  \begin{minipage}{1.0\linewidth}
     \begin{subfigure}[c]{0.49\columnwidth}
         \centering
         \includegraphics[width=\textwidth]{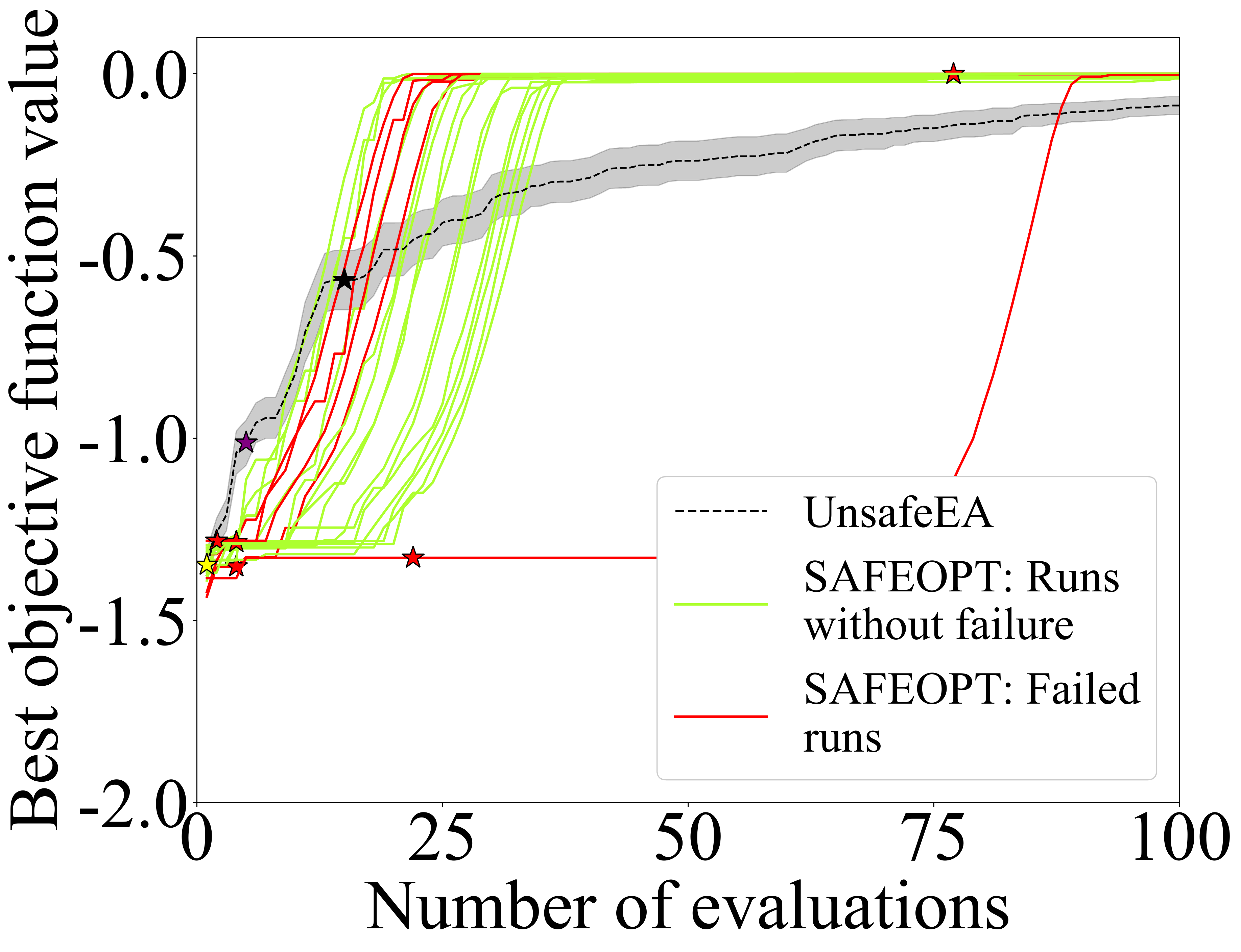}
         \caption{\SafeOpt (Sphere)}
         \label{fig:SpherezeroS}
     \end{subfigure}
     \hfill
     \begin{subfigure}[d]{0.49\columnwidth}
         \centering
         \includegraphics[width=\textwidth]{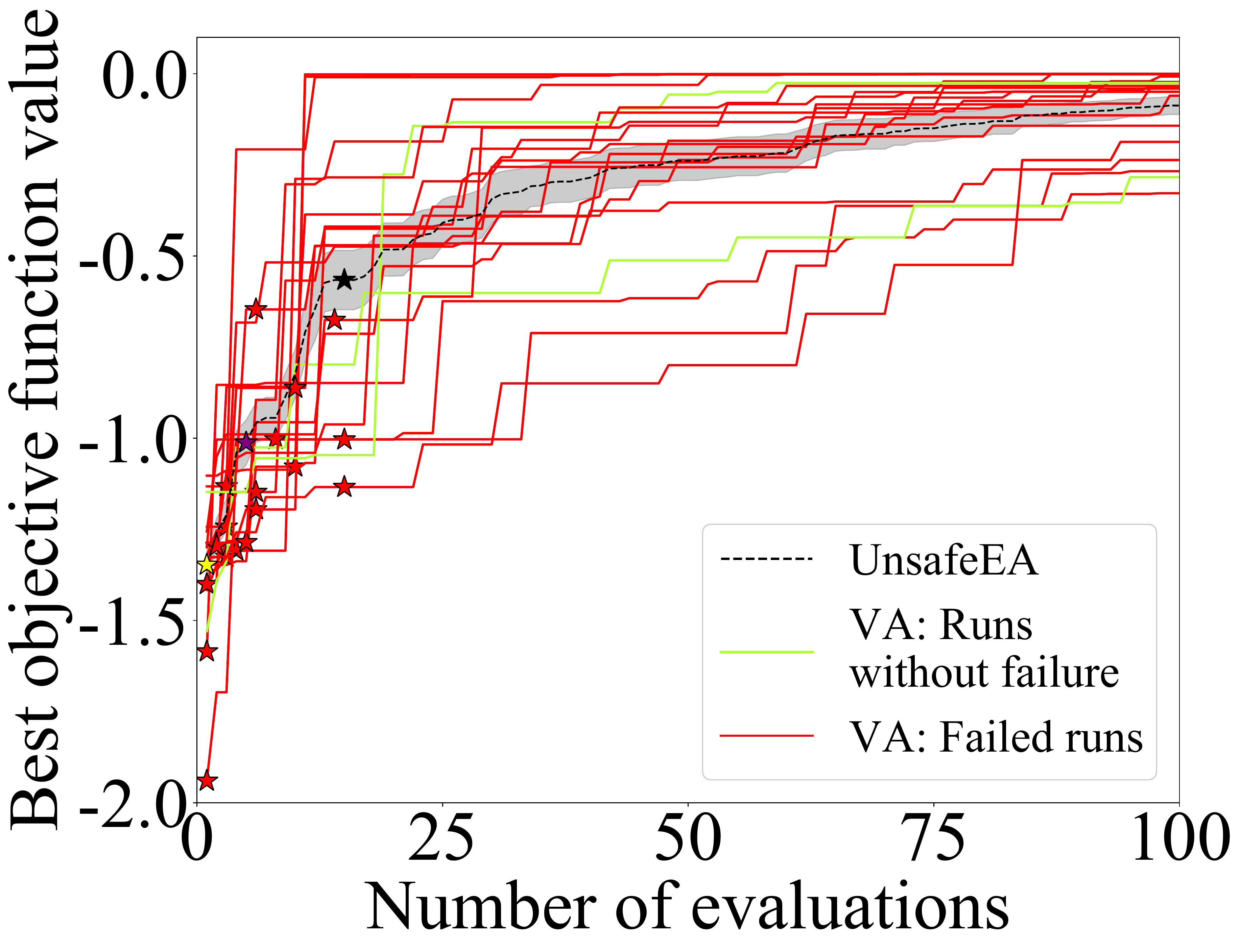}
         \caption{VA (Sphere)}
         \label{fig:SpherezeroVA}
     \end{subfigure}
     \begin{subfigure}[e]{0.49\columnwidth}
         \centering
         \includegraphics[width=\textwidth]{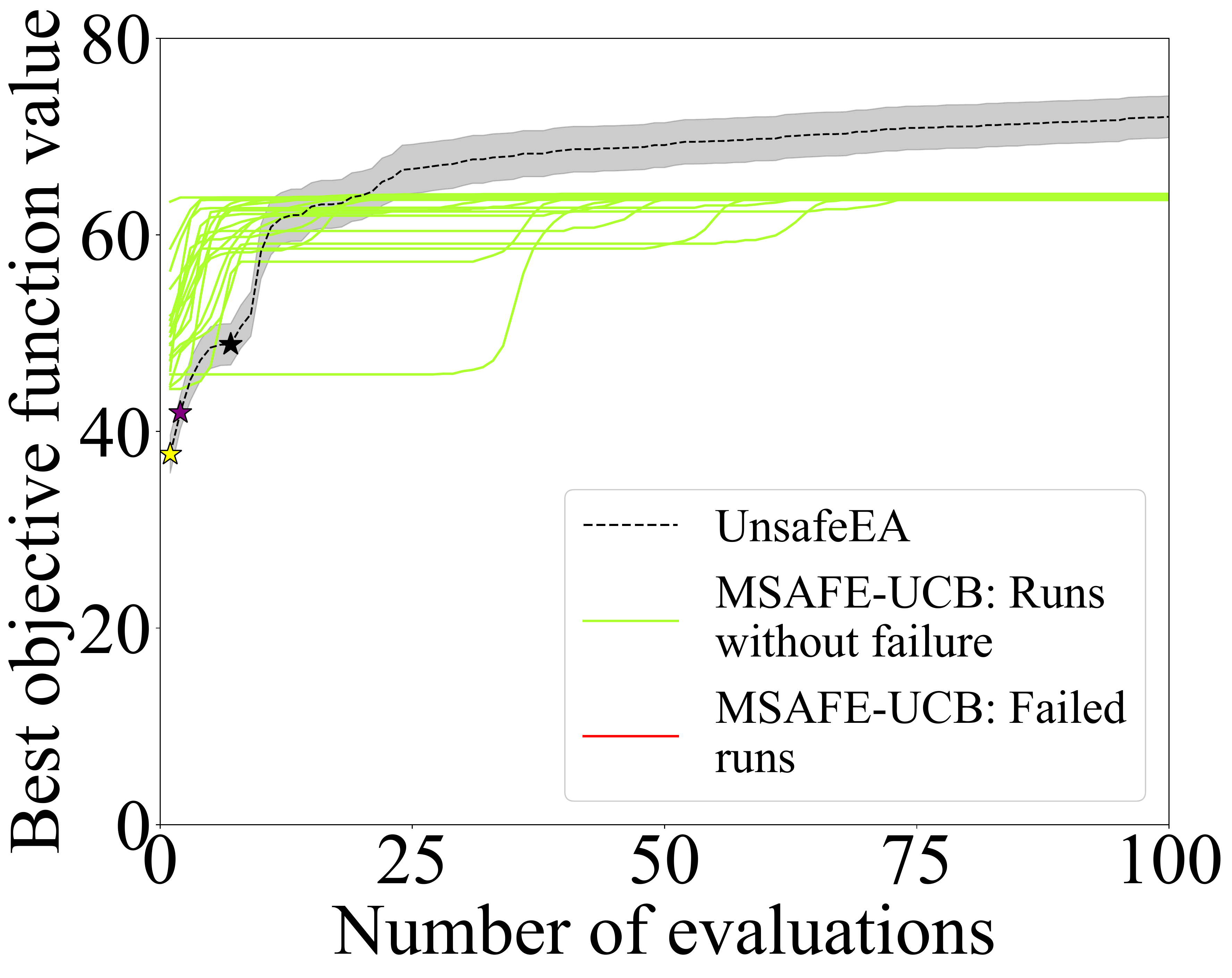}
         \caption{M\SafeOpt (Styblinski-Tang)}
         \label{fig:StybzeroMs}
     \end{subfigure}
     \hfill
     \begin{subfigure}[f]{0.49\columnwidth}
         \centering
         \includegraphics[width=\textwidth]{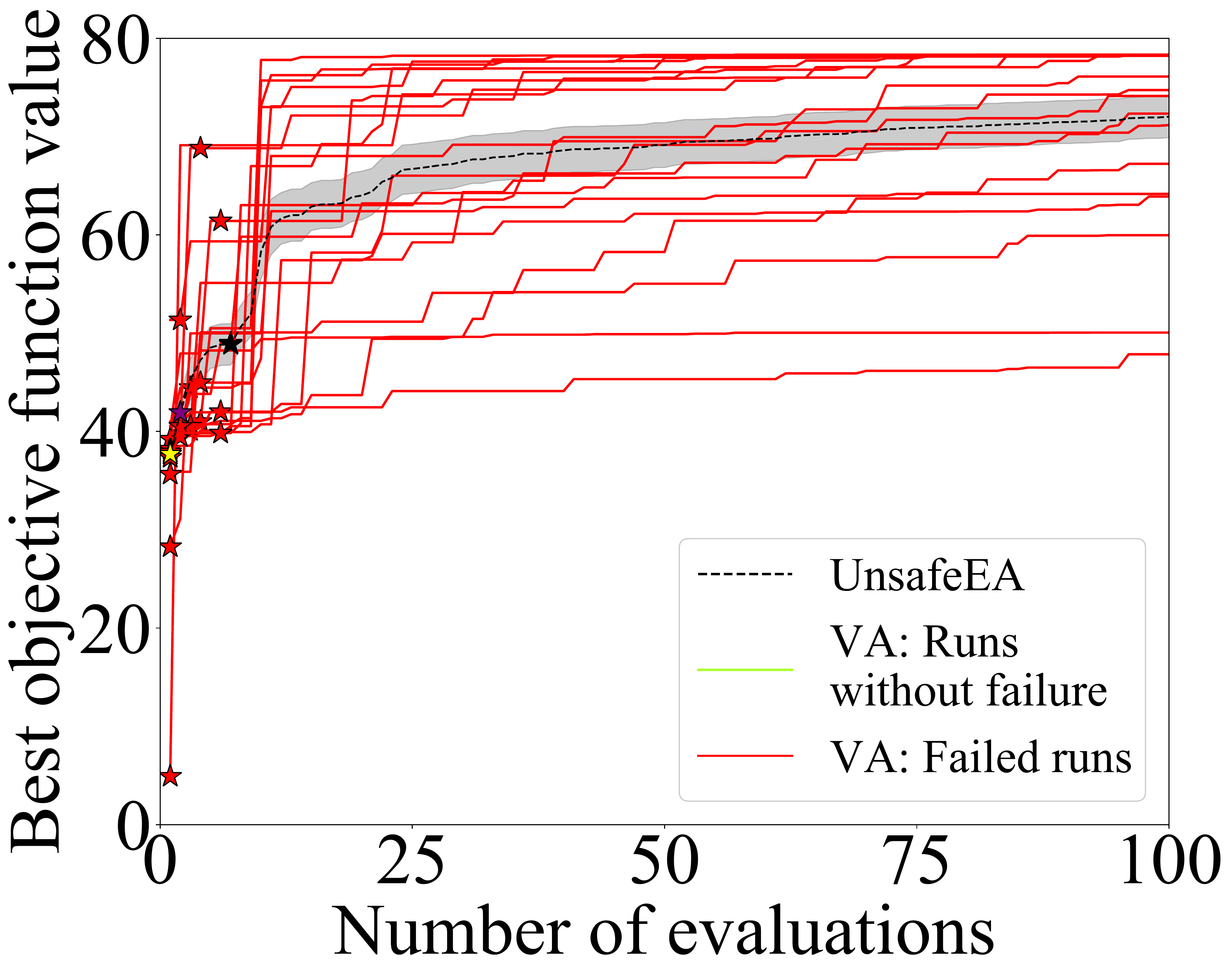}
         \caption{VA (Styblinski-Tang)}
         \label{fig:StybzeroVA}
     \end{subfigure}
        \caption{BSF over function evaluations of each algorithmic run of (middle row) \SafeOpt and VA on Sphere function and  of (bottom row) modified \SafeOpt and VA on Styblinski-Tang function. Same benchmark settings as in Fig.~\ref{fig:zerobudget}. In addition, mean $\pm$ standard error of UnsafeEA is shown as a shaded area around a black solid line;  red lines are runs with at least one unsafe evaluation and green lines with none; the red stars show when the first failure happened for failed runs; and yellow, purple, and black stars indicate the minimum, median, and maximum evaluation step of the first failure of all runs of UnsafeEA.}
        \label{fig:zerobudget_single}
      \end{minipage}
    \end{figure}

Here, we investigate the impact of the budget of unsafe evaluations. For this, let us compare the results obtained by the algorithms for a setting without a limit on the safety budget (shown in Figures~\ref{fig:Spherenuminit10BSF} and~\ref{fig:StybSafeThresBSF75}), with a zero safety budget (Figures~\ref{fig:zerobudget} and~\ref{fig:zerobudget_single}).
We consider the Sphere function first. For a zero safety budget (Figure~\ref{fig:Spherezero}), we observe that several algorithms (modified \SafeOpt, modified \SafeUCB, and VA) cannot match their performance obtained in an environment without a limit on the safety budget (shown in Figure~\ref{fig:Spherenuminit10BSF}).  This is because a zero safety budget causes these algorithms to terminate the optimization run as soon as the first unsafe solution has been evaluated.  \SafeOpt and \SafeUCB, which are provided with more information about the objective function (i.e., \Lipschitz constant), are the only two algorithms considered that efficiently converge to the optimum on the Sphere function.

When a zero budget of unsafe evaluations is applied to optimizing Styblinski-Tang function (Figure~\ref{fig:Stybzero}), VA shows extremely poorer results, and modified \SafeOpt and modified \SafeUCB in comparison become the best algorithms (see Figure~\ref{fig:StybSafeThresBSF75}). It is because the first failure happens too early for VA as shown in Figure~\ref{fig:StybzeroVA}. As mentioned in Section~\ref{sec:ResultsSafeThres}, VA shows a potential of global optimization, however, it costs to evaluate unsafe solutions, and the figure shows that the failure happens from the early stage of optimization. In contrast, as shown in Figure~\ref{fig:trajmsafeopt}, modified \SafeOpt has a tendency to converge to a local optimum, rather than performing global optimization.

The results of each of the 20 algorithmic runs of the best performing algorithms on Sphere and Styblinski-Tang function are presented in Figure~\ref{fig:SpherezeroS} (\SafeOpt) and Figure~\ref{fig:StybzeroMs} (modified \SafeOpt) respectively. In comparison, we also show the same data for  VA on the Sphere function (Figure~\ref{fig:SpherezeroVA}) and Styblinski-Tang function (Figure~\ref{fig:StybzeroVA}). We can observe that GP-based optimizers in Figure~\ref{fig:SpherezeroS} and~\ref{fig:StybzeroMs} are relatively more invariant to the location of initial safe seeds, however, VA shows a significant level of variation in the results across the 20 runs.

\section{Conclusion and Future Research}\label{sec:Conclusion}

The violation of safety constraints in an optimization problem causes an irrevocable loss to the system, e.g., breakage of an expensive hardware kit or loss of a human life. This paper has investigated such safe optimization problems (SafeOPs) by analyzing the performance of two different types of optimization algorithms: EA-based vs GP-based algorithms. This was the largest benchmark study in the safe optimization literature. We also considered a baseline EA, which was blind to safety constraints. Furthermore, we have developed a feature-rich open-source Python framework for (i)~converting any optimization problem into a SafeOP containing different user-specified properties, (ii)~benchmarking different optimization algorithms, and (iii)~visualizing the performance in different ways.

The focus of our experimental study was to determine whether EA-based methods can compete with modern GP-based methods and understand the search behaviour of all algorithms considered for problems of varying complexity. 
%
%
Overall, we observed that the performance of a safe optimization algorithm is affected by the modality of the fitness landscape, as well as key problem parameters, such as parameters governing the location of initial safe seeds, the number of initial safe seeds, the level of safety threshold, and the budget of unsafe evaluations. Addressing the big question of this paper on whether EAs are safe optimizers or not, we can now say that an EA blind to safety constraints did not perform that badly; while we cannot say that EAs are safe optimizers generally, we can certainly say that EAs have the potential to become safe optimizers, e.g. when augmented with  pre-screening methods as employed by VA. 


We suggest several topics for future research based on the above observations. 
An obvious topic is to focus on developing more efficient safe optimization algorithms, and then testing them for different problem settings, similar to how we have done here. This will help build up a fundamental understanding about this so important problem type. Also, one may investigate how to make the current optimizers more efficient, e.g.,~\cite{BlaSteSASBO} proposed a method that enhances \SafeOpt~\cite{SuiGotBur2015icml} with self-adapting hyperparameters.  There is also a need to expand the open-source Python framework that we have made available here to reduce the barrier for new scientists to enter this field and speed up development. Finally, it would be important to understand better the landscape properties of real safe optimization problems and how they vary from the artificial ones created here. This can provide indication about the suitability of our methods in practice.

\begin{acks}
M.\@ L\'opez-Ib\'a\~nez is a ``Beatriz Galindo'' Senior Distinguished Researcher (BEAGAL 18/00053) funded by the Spanish Ministry of Science and Innovation (MICINN).
\end{acks}

\bibliographystyle{ACM-Reference-Format}
\bibliography{bib/abbrev,bib/journals,bib/authors,bib/biblio,bib/crossref,refs}


\providecommand{\MaxMinAntSystem}{{$\cal MAX$--$\cal MIN$} {Ant} {System}}
  \providecommand{\rpackage}[1]{{#1}}
  \providecommand{\softwarepackage}[1]{{#1}}
  \providecommand{\proglang}[1]{{#1}}
\begin{thebibliography}{21}


\ifx \showCODEN    \undefined \def \showCODEN     #1{\unskip}     \fi
\ifx \showDOI      \undefined \def \showDOI       #1{#1}\fi
\ifx \showISBNx    \undefined \def \showISBNx     #1{\unskip}     \fi
\ifx \showISBNxiii \undefined \def \showISBNxiii  #1{\unskip}     \fi
\ifx \showISSN     \undefined \def \showISSN      #1{\unskip}     \fi
\ifx \showLCCN     \undefined \def \showLCCN      #1{\unskip}     \fi
\ifx \shownote     \undefined \def \shownote      #1{#1}          \fi
\ifx \showarticletitle \undefined \def \showarticletitle #1{#1}   \fi
\ifx \showURL      \undefined \def \showURL       {\relax}        \fi
\providecommand\bibfield[2]{#2}
\providecommand\bibinfo[2]{#2}
\providecommand\natexlab[1]{#1}
\providecommand\showeprint[2][]{arXiv:#2}

\bibitem[\protect\citeauthoryear{Allmendinger}{Allmendinger}{2012}]%
        {Allmendinger2012phd}
\bibfield{author}{\bibinfo{person}{Richard Allmendinger}.}
  \bibinfo{year}{2012}\natexlab{}.
\newblock \emph{\bibinfo{title}{Tuning Evolutionary Search for Closed-Loop
  Optimization}}.
\newblock \bibinfo{thesistype}{Ph.\,D. Dissertation}. \bibinfo{school}{The
  University of Manchester, UK}.
\newblock


\bibitem[\protect\citeauthoryear{Allmendinger and Knowles}{Allmendinger and
  Knowles}{2011}]%
        {AllKno2011ecta}
\bibfield{author}{\bibinfo{person}{Richard Allmendinger} {and}
  \bibinfo{person}{Joshua~D. Knowles}.} \bibinfo{year}{2011}\natexlab{}.
\newblock \showarticletitle{Evolutionary Search in Lethal Environments}. In
  \bibinfo{booktitle}{\emph{International Conference on Evolutionary
  Computation Theory and Applications}}. \bibinfo{publisher}{SciTePress},
  \bibinfo{pages}{63--72}.
\newblock
\urldef\tempurl%
\url{https://doi.org/10.5220/0003673000630072}
\showDOI{\tempurl}


\bibitem[\protect\citeauthoryear{Bachoc, Helbert, and Picheny}{Bachoc
  et~al\mbox{.}}{2020}]%
        {BacHelPic2020gaussian}
\bibfield{author}{\bibinfo{person}{Fran{\c c}ois Bachoc},
  \bibinfo{person}{C{\'e}line Helbert}, {and} \bibinfo{person}{Victor
  Picheny}.} \bibinfo{year}{2020}\natexlab{}.
\newblock \showarticletitle{Gaussian process optimization with failures:
  Classification and convergence proof}.
\newblock \bibinfo{journal}{\emph{Journal of Global Optimization}}
  (\bibinfo{year}{2020}).
\newblock
\urldef\tempurl%
\url{https://doi.org/10.1007/s10898-020-00920-0}
\showDOI{\tempurl}


\bibitem[\protect\citeauthoryear{Baeck, Fogel, and Michalewicz}{Baeck
  et~al\mbox{.}}{2000}]%
        {baeck2000evolutionary}
\bibfield{author}{\bibinfo{person}{Thomas Baeck}, \bibinfo{person}{DB Fogel},
  {and} \bibinfo{person}{Z Michalewicz}.} \bibinfo{year}{2000}\natexlab{}.
\newblock \bibinfo{booktitle}{\emph{Evolutionary Computation 1: Basic
  Algorithms and Operators}}. Vol.~\bibinfo{volume}{1}.
\newblock \bibinfo{publisher}{CRC Press}.
\newblock


\bibitem[\protect\citeauthoryear{Berkenkamp, Krause, and Schoellig}{Berkenkamp
  et~al\mbox{.}}{2016a}]%
        {BerKraSch2016bayesian}
\bibfield{author}{\bibinfo{person}{Felix Berkenkamp}, \bibinfo{person}{Andreas
  Krause}, {and} \bibinfo{person}{Angela~P. Schoellig}.}
  \bibinfo{year}{2016}\natexlab{a}.
\newblock \showarticletitle{Bayesian Optimization with Safety Constraints: Safe
  and Automatic Parameter Tuning in Robotics}.
\newblock \bibinfo{journal}{\emph{Arxiv preprint arXiv:1602.04450}}
  (\bibinfo{year}{2016}).
\newblock
\urldef\tempurl%
\url{http://arxiv.org/abs/1602.04450}
\showURL{%
\tempurl}


\bibitem[\protect\citeauthoryear{Berkenkamp, Krause, and Schoellig}{Berkenkamp
  et~al\mbox{.}}{2021}]%
        {BerKraSch2021bayesian}
\bibfield{author}{\bibinfo{person}{Felix Berkenkamp}, \bibinfo{person}{Andreas
  Krause}, {and} \bibinfo{person}{Angela~P. Schoellig}.}
  \bibinfo{year}{2021}\natexlab{}.
\newblock \showarticletitle{Bayesian optimization with safety constraints: safe
  and automatic parameter tuning in robotics}.
\newblock \bibinfo{journal}{\emph{Machine Learning}} (\bibinfo{date}{June}
  \bibinfo{year}{2021}).
\newblock
\urldef\tempurl%
\url{https://doi.org/10.1007/s10994-021-06019-1}
\showDOI{\tempurl}


\bibitem[\protect\citeauthoryear{Berkenkamp, Schoellig, and Krause}{Berkenkamp
  et~al\mbox{.}}{2016b}]%
        {BerSchKra2016safe}
\bibfield{author}{\bibinfo{person}{Felix Berkenkamp},
  \bibinfo{person}{Angela~P. Schoellig}, {and} \bibinfo{person}{Andreas
  Krause}.} \bibinfo{year}{2016}\natexlab{b}.
\newblock \showarticletitle{Safe controller optimization for quadrotors with
  {Gaussian} processes}. In \bibinfo{booktitle}{\emph{2016 IEEE International
  Conference on Robotics and Automation (ICRA)}}. IEEE,
  \bibinfo{pages}{491--496}.
\newblock
\urldef\tempurl%
\url{https://doi.org/10.1109/ICRA.2016.7487170}
\showDOI{\tempurl}


\bibitem[\protect\citeauthoryear{B{\i }y{\i }k, Margoliash, Alimo, and
  Sadigh}{B{\i }y{\i }k et~al\mbox{.}}{2019}]%
        {BiyMarAli2019acc}
\bibfield{author}{\bibinfo{person}{Erdem B{\i }y{\i }k},
  \bibinfo{person}{Jonathan Margoliash}, \bibinfo{person}{Shahrouz~Ryan Alimo},
  {and} \bibinfo{person}{Dorsa Sadigh}.} \bibinfo{year}{2019}\natexlab{}.
\newblock \showarticletitle{Efficient and Safe Exploration in Deterministic
  {Markov} Decision Processes with Unknown Transition Models}. In
  \bibinfo{booktitle}{\emph{2019 American Control Conference ({ACC})}}.
  \bibinfo{publisher}{{IEEE}}, \bibinfo{pages}{1792--1799}.
\newblock
\urldef\tempurl%
\url{https://doi.org/10.23919/ACC.2019.8815276}
\showDOI{\tempurl}


\bibitem[\protect\citeauthoryear{Blasi and Gepperth}{Blasi and
  Gepperth}{2020}]%
        {BlaSteSASBO}
\bibfield{author}{\bibinfo{person}{Stefano~De Blasi} {and}
  \bibinfo{person}{Alexander Gepperth}.} \bibinfo{year}{2020}\natexlab{}.
\newblock \showarticletitle{SASBO: Self-Adapting Safe Bayesian Optimization}.
  In \bibinfo{booktitle}{\emph{2020 19th IEEE International Conference on
  Machine Learning and Applications (ICMLA)}}. \bibinfo{pages}{220--225}.
\newblock
\urldef\tempurl%
\url{https://doi.org/10.1109/ICMLA51294.2020.00044}
\showDOI{\tempurl}


\bibitem[\protect\citeauthoryear{Coello}{Coello}{2002}]%
        {coecarconstraint}
\bibfield{author}{\bibinfo{person}{Carlos A~Coello Coello}.}
  \bibinfo{year}{2002}\natexlab{}.
\newblock \showarticletitle{Theoretical and numerical constraint-handling
  techniques used with evolutionary algorithms: a survey of the state of the
  art}.
\newblock \bibinfo{journal}{\emph{Computer methods in applied mechanics and
  engineering}} \bibinfo{volume}{191}, \bibinfo{number}{11-12}
  (\bibinfo{year}{2002}), \bibinfo{pages}{1245--1287}.
\newblock


\bibitem[\protect\citeauthoryear{Duivenvoorden, Berkenkamp, Carion, Krause, and
  Schoellig}{Duivenvoorden et~al\mbox{.}}{2017}]%
        {DuiBerCar2017constrained}
\bibfield{author}{\bibinfo{person}{Rikky R. P.~R. Duivenvoorden},
  \bibinfo{person}{Felix Berkenkamp}, \bibinfo{person}{Nicolas Carion},
  \bibinfo{person}{Andreas Krause}, {and} \bibinfo{person}{Angela~P.
  Schoellig}.} \bibinfo{year}{2017}\natexlab{}.
\newblock \showarticletitle{Constrained {Bayesian} Optimization with Particle
  Swarms for Safe Adaptive Controller Tuning}.
\newblock \bibinfo{journal}{\emph{{IFAC}-{PapersOnLine}}} \bibinfo{volume}{50},
  \bibinfo{number}{1} (\bibinfo{year}{2017}), \bibinfo{pages}{11800--11807}.
\newblock
\urldef\tempurl%
\url{https://doi.org/10.1016/j.ifacol.2017.08.1991}
\showDOI{\tempurl}


\bibitem[\protect\citeauthoryear{Hansen, Finck, Ros, and Auger}{Hansen
  et~al\mbox{.}}{2009}]%
        {hansen2009real}
\bibfield{author}{\bibinfo{person}{Nikolaus Hansen}, \bibinfo{person}{Steffen
  Finck}, \bibinfo{person}{Raymond Ros}, {and} \bibinfo{person}{Anne Auger}.}
  \bibinfo{year}{2009}\natexlab{}.
\newblock \bibinfo{booktitle}{\emph{{Real-Parameter Black-Box Optimization
  Benchmarking 2009: Noiseless Functions Definitions}}}.
\newblock \bibinfo{type}{Research Report} RR-6829.
  \bibinfo{institution}{{INRIA}}.
\newblock
\urldef\tempurl%
\url{https://hal.inria.fr/inria-00362633}
\showURL{%
\tempurl}


\bibitem[\protect\citeauthoryear{Kaji, Ikeda, and Kita}{Kaji
  et~al\mbox{.}}{2009}]%
        {KajIkeHaj2009cec}
\bibfield{author}{\bibinfo{person}{H. Kaji}, \bibinfo{person}{Kokolo Ikeda},
  {and} \bibinfo{person}{Hajime Kita}.} \bibinfo{year}{2009}\natexlab{}.
\newblock \showarticletitle{Avoidance of constraint violation for
  experiment-based evolutionary multi-objective optimization}. In
  \bibinfo{booktitle}{\emph{Proceedings of the 2009 Congress on Evolutionary
  Computation (CEC 2009)}}. \bibinfo{publisher}{IEEE Press},
  \bibinfo{address}{Piscataway, NJ}, \bibinfo{pages}{2756--2763}.
\newblock
\urldef\tempurl%
\url{https://doi.org/10.1109/CEC.2009.4983288}
\showDOI{\tempurl}


\bibitem[\protect\citeauthoryear{Kim, Allmendinger, and
  L{\'o}pez-Ib{\'a}{\~n}ez}{Kim et~al\mbox{.}}{2021}]%
        {KimAllLop2020safe}
\bibfield{author}{\bibinfo{person}{Youngmin Kim}, \bibinfo{person}{Richard
  Allmendinger}, {and} \bibinfo{person}{Manuel L{\'o}pez-Ib{\'a}{\~n}ez}.}
  \bibinfo{year}{2021}\natexlab{}.
\newblock \showarticletitle{Safe Learning and Optimization Techniques: Towards
  a Survey of the State of the Art}.
\newblock In \bibinfo{booktitle}{\emph{Trustworthy AI -- Integrating Learning,
  Optimization and Reasoning. TAILOR 2020}},
  \bibfield{editor}{\bibinfo{person}{Fredrik Heintz}, \bibinfo{person}{Michela
  Milano}, {and} \bibinfo{person}{Barry O'Sullivan}} (Eds.).
  \bibinfo{series}{Lecture Notes in Computer Science},
  Vol.~\bibinfo{volume}{12641}. \bibinfo{publisher}{Springer},
  \bibinfo{address}{Cham, Switzerland}, \bibinfo{pages}{123--139}.
\newblock
\urldef\tempurl%
\url{https://doi.org/10.1007/978-3-030-73959-1_12}
\showDOI{\tempurl}


\bibitem[\protect\citeauthoryear{Michalewicz}{Michalewicz}{1995}]%
        {Michalewicz1995ASO}
\bibfield{author}{\bibinfo{person}{Zbigniew Michalewicz}.}
  \bibinfo{year}{1995}\natexlab{}.
\newblock \showarticletitle{A Survey of Constraint Handling Techniques in
  Evolutionary Computation Methods}. In \bibinfo{booktitle}{\emph{Evolutionary
  Programming}}.
\newblock


\bibitem[\protect\citeauthoryear{Sacher, Duvigneau, Le~Maitre, Durand, Berrini,
  Hauville, and Astolfi}{Sacher et~al\mbox{.}}{2018}]%
        {SacDuvMai2018ego-ls-svm}
\bibfield{author}{\bibinfo{person}{Matthieu Sacher}, \bibinfo{person}{R{\'e}gis
  Duvigneau}, \bibinfo{person}{Olivier Le~Maitre}, \bibinfo{person}{Mathieu
  Durand}, \bibinfo{person}{Elisa Berrini}, \bibinfo{person}{Fr{\'e}d{\'e}ric
  Hauville}, {and} \bibinfo{person}{Jacques-Andr{\'e} Astolfi}.}
  \bibinfo{year}{2018}\natexlab{}.
\newblock \showarticletitle{A classification approach to efficient global
  optimization in presence of non-computable domains}.
\newblock \bibinfo{journal}{\emph{Structural and Multidisciplinary
  Optimization}} \bibinfo{volume}{58}, \bibinfo{number}{4}
  (\bibinfo{year}{2018}), \bibinfo{pages}{1537--1557}.
\newblock
\urldef\tempurl%
\url{https://doi.org/10.1007/s00158-018-1981-8}
\showDOI{\tempurl}


\bibitem[\protect\citeauthoryear{Schillinger, Hartmann, Skalecki, Meister,
  Nguyen-Tuong, and Nelles}{Schillinger et~al\mbox{.}}{2017}]%
        {SchHarSka2017safe}
\bibfield{author}{\bibinfo{person}{Mark Schillinger}, \bibinfo{person}{Benjamin
  Hartmann}, \bibinfo{person}{Patric Skalecki}, \bibinfo{person}{Mona Meister},
  \bibinfo{person}{Duy Nguyen-Tuong}, {and} \bibinfo{person}{Oliver Nelles}.}
  \bibinfo{year}{2017}\natexlab{}.
\newblock \showarticletitle{Safe active learning and safe {Bayesian}
  optimization for tuning a {PI}-controller}.
\newblock \bibinfo{journal}{\emph{{IFAC}-{PapersOnLine}}} \bibinfo{volume}{50},
  \bibinfo{number}{1} (\bibinfo{year}{2017}), \bibinfo{pages}{5967--5972}.
\newblock
\urldef\tempurl%
\url{https://doi.org/10.1016/j.ifacol.2017.08.1258}
\showDOI{\tempurl}


\bibitem[\protect\citeauthoryear{Srinivas, Krause, Kakade, and Seeger}{Srinivas
  et~al\mbox{.}}{2010}]%
        {srinivas10gaussian}
\bibfield{author}{\bibinfo{person}{Niranjan Srinivas}, \bibinfo{person}{Andreas
  Krause}, \bibinfo{person}{Sham Kakade}, {and} \bibinfo{person}{Matthias
  Seeger}.} \bibinfo{year}{2010}\natexlab{}.
\newblock \showarticletitle{Gaussian Process Optimization in the Bandit
  Setting: No Regret and Experimental Design}. In
  \bibinfo{booktitle}{\emph{Proceedings of the 27th International Conference on
  International Conference on Machine Learning}} (Haifa, Israel)
  \emph{(\bibinfo{series}{ICML'10})}. \bibinfo{publisher}{Omnipress},
  \bibinfo{address}{Madison, WI, USA}, \bibinfo{pages}{1015–1022}.
\newblock
\showISBNx{9781605589077}


\bibitem[\protect\citeauthoryear{Sui, Gotovos, Burdick, and Krause}{Sui
  et~al\mbox{.}}{2015}]%
        {SuiGotBur2015icml}
\bibfield{author}{\bibinfo{person}{Yanan Sui}, \bibinfo{person}{Alkis Gotovos},
  \bibinfo{person}{Joel~W. Burdick}, {and} \bibinfo{person}{Andreas Krause}.}
  \bibinfo{year}{2015}\natexlab{}.
\newblock \showarticletitle{Safe Exploration for Optimization with {Gaussian}
  Processes}. In \bibinfo{booktitle}{\emph{Proceedings of the 32nd
  International Conference on Machine Learning, {ICML} 2015}},
  \bibfield{editor}{\bibinfo{person}{Francis Bach} {and} \bibinfo{person}{David
  Blei}} (Eds.), Vol.~\bibinfo{volume}{37}. \bibinfo{pages}{997--1005}.
\newblock


\bibitem[\protect\citeauthoryear{Sui, Zhuang, Burdick, and Yue}{Sui
  et~al\mbox{.}}{2018}]%
        {SuiZhuBur2018stageopt}
\bibfield{author}{\bibinfo{person}{Yanan Sui}, \bibinfo{person}{Vincent
  Zhuang}, \bibinfo{person}{Joel~W. Burdick}, {and} \bibinfo{person}{Yisong
  Yue}.} \bibinfo{year}{2018}\natexlab{}.
\newblock \showarticletitle{Stagewise Safe {Bayesian} Optimization with
  {Gaussian} Processes}. In \bibinfo{booktitle}{\emph{Proceedings of the 35th
  International Conference on Machine Learning, {ICML} 2018}}
  \emph{(\bibinfo{series}{Proceedings of Machine Learning Research},
  Vol.~\bibinfo{volume}{80})}, \bibfield{editor}{\bibinfo{person}{Jennifer~G.
  Dy} {and} \bibinfo{person}{Andreas Krause}} (Eds.).
  \bibinfo{publisher}{{PMLR}}, \bibinfo{pages}{4788--4796}.
\newblock


\bibitem[\protect\citeauthoryear{Zubanovic, Hidic, Hajdarevic, Nosovic, and
  Konjicija}{Zubanovic et~al\mbox{.}}{2014}]%
        {ZubHidHaj2014}
\bibfield{author}{\bibinfo{person}{D. Zubanovic}, \bibinfo{person}{A. Hidic},
  \bibinfo{person}{A. Hajdarevic}, \bibinfo{person}{N. Nosovic}, {and}
  \bibinfo{person}{S. Konjicija}.} \bibinfo{year}{2014}\natexlab{}.
\newblock \showarticletitle{Performance analysis of parallel master-slave
  Evolutionary strategies ($\mu$,$\lambda$) model python implementation for CPU
  and GPGPU}. In \bibinfo{booktitle}{\emph{2014 37th International Convention
  on Information and Communication Technology, Electronics and Microelectronics
  (MIPRO)}}. \bibinfo{pages}{1609--1613}.
\newblock
\urldef\tempurl%
\url{https://doi.org/10.1109/MIPRO.2014.6859822}
\showDOI{\tempurl}


\end{thebibliography}



\providecommand{\MaxMinAntSystem}{{$\cal MAX$--$\cal MIN$} {Ant} {System}}
  \providecommand{\rpackage}[1]{{#1}}
  \providecommand{\softwarepackage}[1]{{#1}}
  \providecommand{\proglang}[1]{{#1}}
\begin{thebibliography}{9}


\ifx \showCODEN    \undefined \def \showCODEN     #1{\unskip}     \fi
\ifx \showDOI      \undefined \def \showDOI       #1{#1}\fi
\ifx \showISBNx    \undefined \def \showISBNx     #1{\unskip}     \fi
\ifx \showISBNxiii \undefined \def \showISBNxiii  #1{\unskip}     \fi
\ifx \showISSN     \undefined \def \showISSN      #1{\unskip}     \fi
\ifx \showLCCN     \undefined \def \showLCCN      #1{\unskip}     \fi
\ifx \shownote     \undefined \def \shownote      #1{#1}          \fi
\ifx \showarticletitle \undefined \def \showarticletitle #1{#1}   \fi
\ifx \showURL      \undefined \def \showURL       {\relax}        \fi
\providecommand\bibfield[2]{#2}
\providecommand\bibinfo[2]{#2}
\providecommand\natexlab[1]{#1}
\providecommand\showeprint[2][]{arXiv:#2}

\bibitem[\protect\citeauthoryear{Berkenkamp, Krause, and Schoellig}{Berkenkamp
  et~al\mbox{.}}{2016a}]%
        {BerKraSch2016bayesian}
\bibfield{author}{\bibinfo{person}{Felix Berkenkamp}, \bibinfo{person}{Andreas
  Krause}, {and} \bibinfo{person}{Angela~P. Schoellig}.}
  \bibinfo{year}{2016}\natexlab{a}.
\newblock \showarticletitle{Bayesian Optimization with Safety Constraints: Safe
  and Automatic Parameter Tuning in Robotics}.
\newblock \bibinfo{journal}{\emph{Arxiv preprint arXiv:1602.04450}}
  (\bibinfo{year}{2016}).
\newblock
\urldef\tempurl%
\url{http://arxiv.org/abs/1602.04450}
\showURL{%
\tempurl}


\bibitem[\protect\citeauthoryear{Berkenkamp, Schoellig, and Krause}{Berkenkamp
  et~al\mbox{.}}{2016b}]%
        {BerSchKra2016safe}
\bibfield{author}{\bibinfo{person}{Felix Berkenkamp},
  \bibinfo{person}{Angela~P. Schoellig}, {and} \bibinfo{person}{Andreas
  Krause}.} \bibinfo{year}{2016}\natexlab{b}.
\newblock \showarticletitle{Safe controller optimization for quadrotors with
  {Gaussian} processes}. In \bibinfo{booktitle}{\emph{2016 IEEE International
  Conference on Robotics and Automation (ICRA)}}. IEEE,
  \bibinfo{pages}{491--496}.
\newblock
\urldef\tempurl%
\url{https://doi.org/10.1109/ICRA.2016.7487170}
\showDOI{\tempurl}


\bibitem[\protect\citeauthoryear{Chowdhury and Gopalan}{Chowdhury and
  Gopalan}{2017}]%
        {ChoSayGopHanFinRosAug2009bbob}
\bibfield{author}{\bibinfo{person}{Sayak~Ray Chowdhury} {and}
  \bibinfo{person}{Aditya Gopalan}.} \bibinfo{year}{2017}\natexlab{}.
\newblock \showarticletitle{On Kernelized Multi-Armed Bandits}. In
  \bibinfo{booktitle}{\emph{Proceedings of the 34th International Conference on
  Machine Learning - Volume 70}} (Sydney, NSW, Australia)
  \emph{(\bibinfo{series}{ICML'17})}. \bibinfo{publisher}{JMLR.org},
  \bibinfo{pages}{844–853}.
\newblock


\bibitem[\protect\citeauthoryear{Duivenvoorden, Berkenkamp, Carion, Krause, and
  Schoellig}{Duivenvoorden et~al\mbox{.}}{2017}]%
        {DuiBerCar2017constrained}
\bibfield{author}{\bibinfo{person}{Rikky R. P.~R. Duivenvoorden},
  \bibinfo{person}{Felix Berkenkamp}, \bibinfo{person}{Nicolas Carion},
  \bibinfo{person}{Andreas Krause}, {and} \bibinfo{person}{Angela~P.
  Schoellig}.} \bibinfo{year}{2017}\natexlab{}.
\newblock \showarticletitle{Constrained {Bayesian} Optimization with Particle
  Swarms for Safe Adaptive Controller Tuning}.
\newblock \bibinfo{journal}{\emph{{IFAC}-{PapersOnLine}}} \bibinfo{volume}{50},
  \bibinfo{number}{1} (\bibinfo{year}{2017}), \bibinfo{pages}{11800--11807}.
\newblock
\urldef\tempurl%
\url{https://doi.org/10.1016/j.ifacol.2017.08.1991}
\showDOI{\tempurl}


\bibitem[\protect\citeauthoryear{Gretton}{Gretton}{2013}]%
        {gretton2013introduction}
\bibfield{author}{\bibinfo{person}{Arthur Gretton}.}
  \bibinfo{year}{2013}\natexlab{}.
\newblock \showarticletitle{Introduction to rkhs, and some simple kernel
  algorithms}.
\newblock \bibinfo{journal}{\emph{Adv. Top. Mach. Learn. Lecture Conducted from
  University College London}}  \bibinfo{volume}{16} (\bibinfo{year}{2013}).
\newblock


\bibitem[\protect\citeauthoryear{Kim, Allmendinger, and
  L{\'o}pez-Ib{\'a}{\~{n}}ez}{Kim et~al\mbox{.}}{2021}]%
        {KimAllMan2021}
\bibfield{author}{\bibinfo{person}{Youngmin Kim}, \bibinfo{person}{Richard
  Allmendinger}, {and} \bibinfo{person}{Manuel L{\'o}pez-Ib{\'a}{\~{n}}ez}.}
  \bibinfo{year}{2021}\natexlab{}.
\newblock \showarticletitle{Safe Learning and Optimization Techniques: Towards
  a Survey of the State of the Art}. In \bibinfo{booktitle}{\emph{Trustworthy
  AI - Integrating Learning, Optimization and Reasoning}},
  \bibfield{editor}{\bibinfo{person}{Fredrik Heintz}, \bibinfo{person}{Michela
  Milano}, {and} \bibinfo{person}{Barry O'Sullivan}} (Eds.).
  \bibinfo{publisher}{Springer International Publishing},
  \bibinfo{address}{Cham}, \bibinfo{pages}{123--139}.
\newblock
\showISBNx{978-3-030-73959-1}


\bibitem[\protect\citeauthoryear{Schulz, Speekenbrink, and Krause}{Schulz
  et~al\mbox{.}}{2018}]%
        {SchSpeKra2018gptut}
\bibfield{author}{\bibinfo{person}{Eric Schulz}, \bibinfo{person}{Maarten
  Speekenbrink}, {and} \bibinfo{person}{Andreas Krause}.}
  \bibinfo{year}{2018}\natexlab{}.
\newblock \showarticletitle{A tutorial on {Gaussian} process regression:
  {Modelling}, exploring, and exploiting functions}.
\newblock \bibinfo{journal}{\emph{Journal of Mathematical Psychology}}
  \bibinfo{volume}{85} (\bibinfo{date}{Aug.} \bibinfo{year}{2018}),
  \bibinfo{pages}{1--16}.
\newblock
\urldef\tempurl%
\url{https://doi.org/10.1016/j.jmp.2018.03.001}
\showDOI{\tempurl}


\bibitem[\protect\citeauthoryear{Srinivas, Krause, Kakade, and Seeger}{Srinivas
  et~al\mbox{.}}{2010}]%
        {srinivas10gaussian}
\bibfield{author}{\bibinfo{person}{Niranjan Srinivas}, \bibinfo{person}{Andreas
  Krause}, \bibinfo{person}{Sham Kakade}, {and} \bibinfo{person}{Matthias
  Seeger}.} \bibinfo{year}{2010}\natexlab{}.
\newblock \showarticletitle{Gaussian Process Optimization in the Bandit
  Setting: No Regret and Experimental Design}. In
  \bibinfo{booktitle}{\emph{Proceedings of the 27th International Conference on
  International Conference on Machine Learning}} (Haifa, Israel)
  \emph{(\bibinfo{series}{ICML'10})}. \bibinfo{publisher}{Omnipress},
  \bibinfo{address}{Madison, WI, USA}, \bibinfo{pages}{1015–1022}.
\newblock
\showISBNx{9781605589077}


\bibitem[\protect\citeauthoryear{Sui, Gotovos, Burdick, and Krause}{Sui
  et~al\mbox{.}}{2015}]%
        {SuiGotBur2015icml}
\bibfield{author}{\bibinfo{person}{Yanan Sui}, \bibinfo{person}{Alkis Gotovos},
  \bibinfo{person}{Joel~W. Burdick}, {and} \bibinfo{person}{Andreas Krause}.}
  \bibinfo{year}{2015}\natexlab{}.
\newblock \showarticletitle{Safe Exploration for Optimization with {Gaussian}
  Processes}. In \bibinfo{booktitle}{\emph{Proceedings of the 32nd
  International Conference on Machine Learning, {ICML} 2015}},
  \bibfield{editor}{\bibinfo{person}{Francis Bach} {and} \bibinfo{person}{David
  Blei}} (Eds.), Vol.~\bibinfo{volume}{37}. \bibinfo{pages}{997--1005}.
\newblock


\end{thebibliography}
\end{document}


\title{Are Evolutionary Algorithms Safe Optimizers?}
\thanks{Please cite as: Youngmin Kim, Richard Allmendinger, and Manuel López-Ibañez. 2022. Are Evolutionary Algorithms Safe Optimizers?. In
  \emph{Genetic and Evolutionary Computation Conference (GECCO ’22)}, July
  9–13, 2022, Boston, MA, USA. ACM, New York, NY, USA, 9
  pages. \url{https://doi.org/10.1145/3512290.3528818}}
\subtitle{Supplementary Material}
\author{Youngmin Kim}
\email{youngmin.kim@manchester.ac.uk}
\orcid{0000000276996532}
\affiliation{%
  \institution{Alliance Manchester Business School, University of Manchester}
  \streetaddress{Booth St W}
  \city{Manchester}
  \country{UK}
  \postcode{M15 6PB}
}

\author{Richard Allmendinger}
\email{richard.allmendinger@manchester.ac.uk}
\orcid{0000000312363143}
\affiliation{%
  \institution{Alliance Manchester Business School, University of Manchester}
  \streetaddress{Booth St W}
  \city{Manchester}
  \country{UK}
  \postcode{M15 6PB}
}

\author{Manuel López-Ibáñez}
\email{manuel.lopez-ibanez@manchester.ac.uk}
\orcid{0000000199741295}
\affiliation{%
  \institution{Alliance Manchester Business School, University of Manchester}
  \streetaddress{Booth St W}
  \city{Manchester}
  \country{UK}
  \postcode{M15 6PB}}
\additionalaffiliation{
  \institution{ITIS Software, School of Computer Science, University of Málaga}
  \city{Málaga}
  \country{Spain}
  \postcode{29071}
}

\maketitle
\tableofcontents
\listoffigures
\section{Parameter Setting for Safe GP Algorithms}

\SafeOpt~\cite{SuiGotBur2015icml} and modified \SafeOpt~\cite{BerSchKra2016safe} select an input point whose width of predicted confidence interval is the maximum for evaluation (\textit{Bayesian experimental design}~\cite{srinivas10gaussian}), whereas UCB criterion is applied in \SafeUCB~\cite{SuiGotBur2015icml} and modified \SafeUCB~\cite{BerSchKra2016safe} (\textit{multi-armed banit paradigm}~\cite{srinivas10gaussian}). Here, predicted confidence interval can be represented by lower ($l_{t}$) and upper ($u_{t}$) bound:
\begin{equation}\label{eq:cinaive}
    l_{t}=\mu_{t-1}(\vec{x})-\beta_t\sigma_{t-1}(\vec{x}),\
    u_{t}=\mu_{t-1}(\vec{x})+\beta_t\sigma_{t-1}(\vec{x})
\end{equation}
where $\beta_t$ is a parameter that decides the width of the predicted confidence interval and $\sigma_{t-1}$ is the predicted standard deviation at point
$\vec{x}$~\cite{SchSpeKra2018gptut}, at $t^{th}$ iteration step~\cite{KimAllMan2021}. There are two different approaches in selecting $\beta_t$ in Eq.~(\ref{eq:cinaive}). In \SafeOpt, $\beta_t$ is set as an iteration-varying scalar, while modified \SafeOpt used a fixed constant over iterations ($\beta_t=2$). As seen in~\cite{BerKraSch2016bayesian}, setting $\beta_t$ as a fixed constant over iterations approximately means that we apply a certain failure probability per iteration, not over all iterations. Thus, it would be ideal if the iteration-varying scalar can be estimated at each iteration step for $\beta_t$, however, there is a strict assumption for applying it to estimating predicted confidence interval. \SafeOpt~\cite{SuiGotBur2015icml} adopted a study~\cite{srinivas10gaussian}, which decides the tightness of predicted confidence interval by controlling the value of $\beta_t$ at each iteration step. In the study~\cite{srinivas10gaussian}, the parameter $\beta_t$ at $t^{th}$ iteration step is estimated as,
\begin{equation}\label{eq:itervaryingbeta}
    \beta_t = 2B + 300 \gamma_t \log^3(t/\delta)
\end{equation}
where $B$ is a bound on the \textit{reproducing kernel Hilbert space} (RKHS) norm, $\gamma_t$ is the maximal mutual information which can be obtained from t samples about the GP prior, and $\delta$ is a probability of failure allowed~\cite{SuiGotBur2015icml,srinivas10gaussian}. $B$ in Eq.~(\ref{eq:itervaryingbeta}) is available under the assumption that an objective function in an optimization problem lies in RKHS. Estimating $\beta_t$ based on Eq.~(\ref{eq:itervaryingbeta}) was appropriate in~\cite{SuiGotBur2015icml}, as the experiments on synthetic data in that paper were made on random objective functions sampled from a zero-mean GP with squared-exponential kernel. However, it is difficult to assume that an objective function in a test problem, created while assuming the real-world problem, lies in RKHS, and thus, we could not apply Eq.~(\ref{eq:itervaryingbeta}) for estimating $\beta_t$. Instead, we set $\beta_t$ equal to two for the experiments, which was used for modified \SafeOpt and is one of the typical choices used for estimating predicted confidence interval~\cite{DuiBerCar2017constrained}. For details about RKHS, please see~\cite{gretton2013introduction}, and to see details about Eq.~(\ref{eq:itervaryingbeta}) (e.g., how the equation was deducted, how to estimate maximal mutual information, etc.), please see~\cite{srinivas10gaussian,SuiGotBur2015icml,ChoSayGopHanFinRosAug2009bbob}.

In~\cite{SuiGotBur2015icml}, the gradient of several random functions, that were sampled from the same GP used for sampling objective functions for experiments on synthetic data in that study, was used for estimating Lipschitz constant. However, we cannot sample any function to estimate Lipschitz constant for our experiments; thus, we compute gradient directly from our objective functions on the discretized domain, estimate Lipschitz constant (by the definition of $L$-Lipschitz continuity, we set the maximum absolute value of gradient observed over the discretized domain as a Lipschitz constant), and provide it to \SafeOpt and \SafeUCB. Hence, the results of our benchmarking experiments are made with perfect information given for estimating the gradient of our objective function; thus, please bear in mind that in the condition where less information on gradient of an objective function is given, the result can be worse than ours, e.g., more safety constraint violations and slower optimization process (or even stuck at certain area), when smaller and greater value than our Lipschitz constant value is given as Lipschitz constant, respectively. 

\section{Supplementary Figures: Styblinski-Tang Function}
Here, plots for each scenario of sampling space of initial safe seeds for Styblinski-Tang function are provided.

\begin{figure}[!hp]
     \centering
     \begin{subfigure}{0.23\textwidth}
         \centering
         \includegraphics[width=\textwidth]{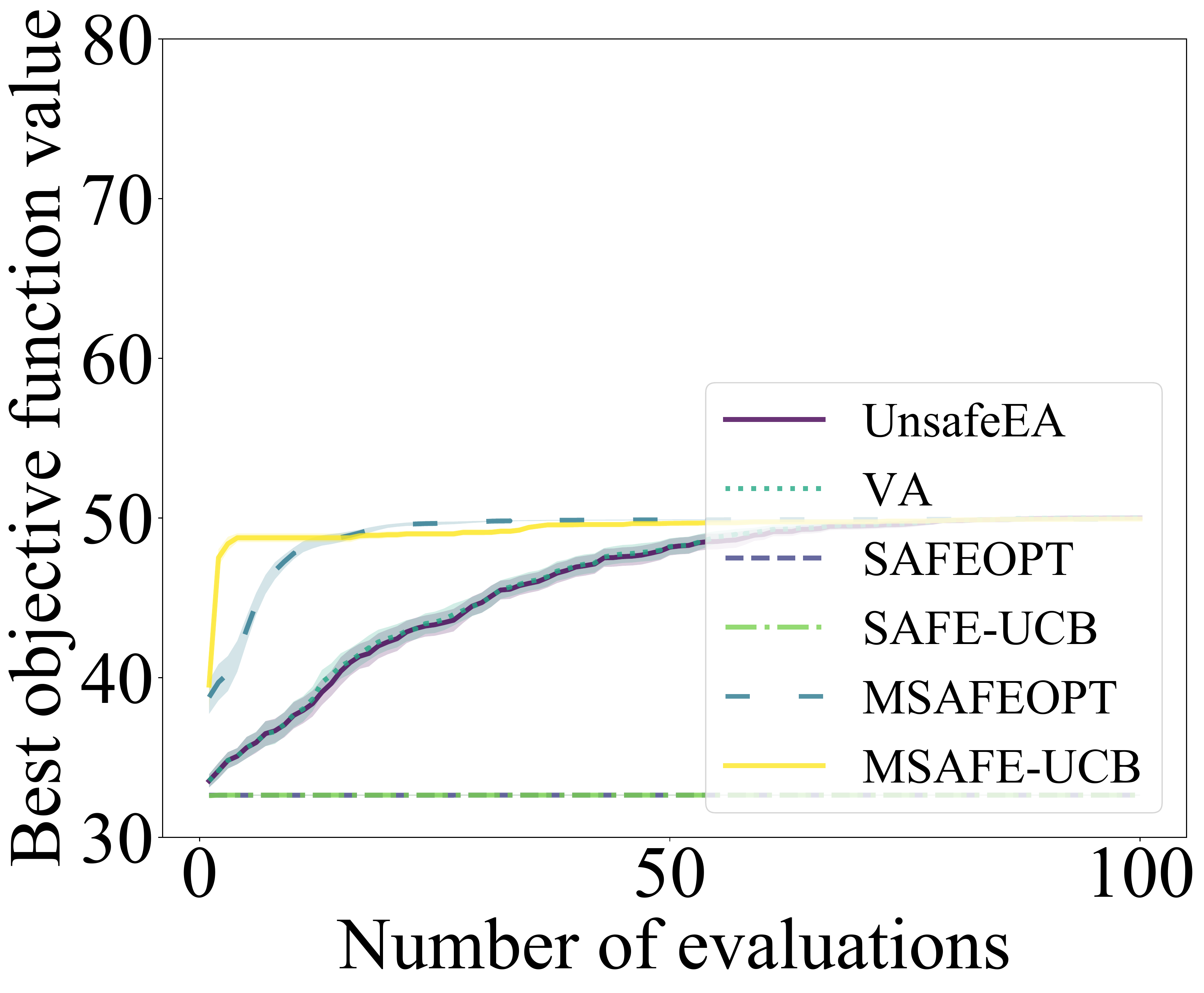}
         \caption{BSF: $h=65^\mathrm{th}$}
         \label{fig:Spherenuminit2BSF}
     \end{subfigure}
     \begin{subfigure}{0.23\textwidth}
         \centering
         \includegraphics[width=\textwidth]{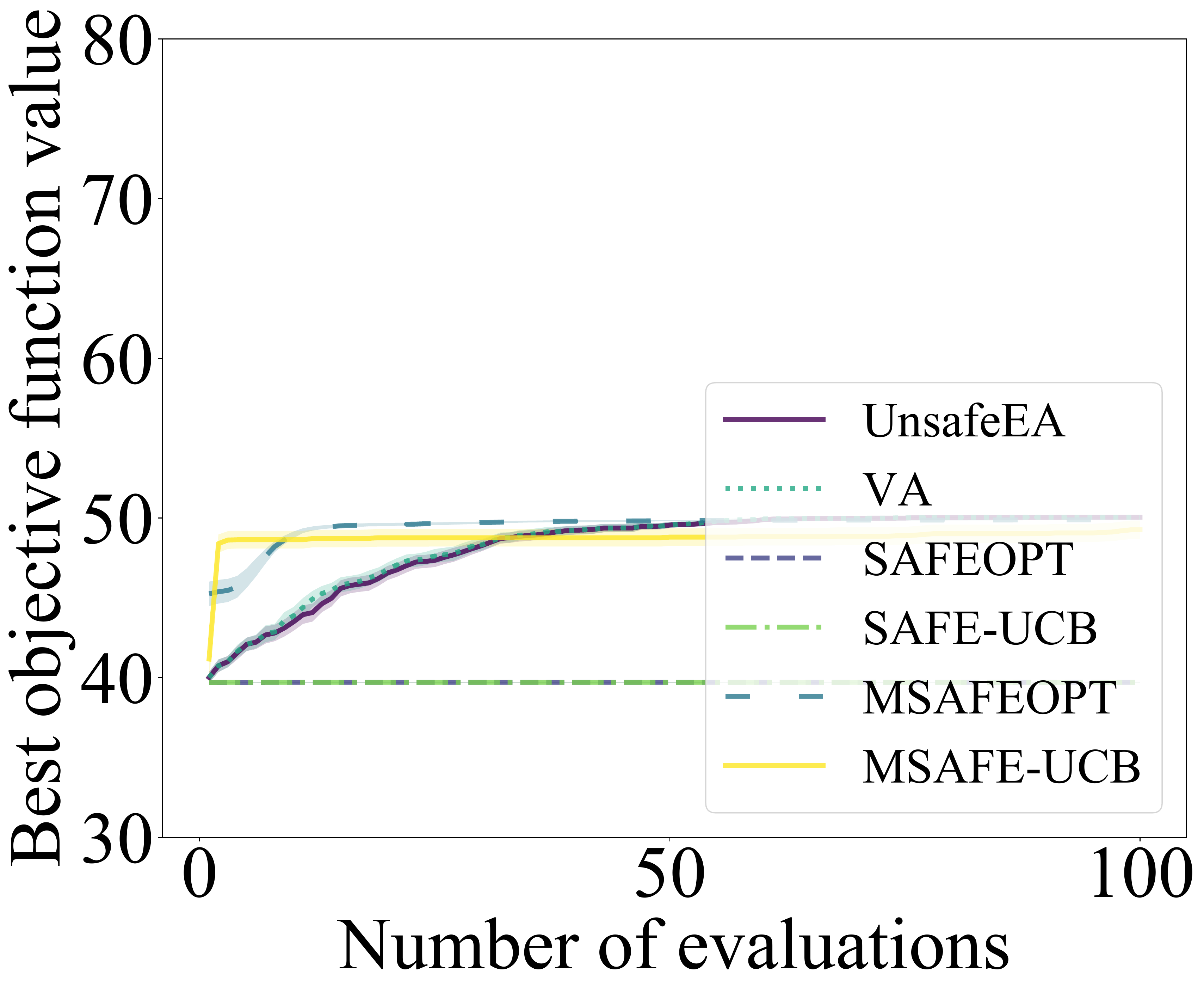}
         \caption{BSF: $h=75^\mathrm{th}$}
         \label{fig:Spherenuminit10BSF}
     \end{subfigure}
     \hfill
     \begin{subfigure}{0.23\textwidth}
         \centering
         \includegraphics[width=\textwidth]{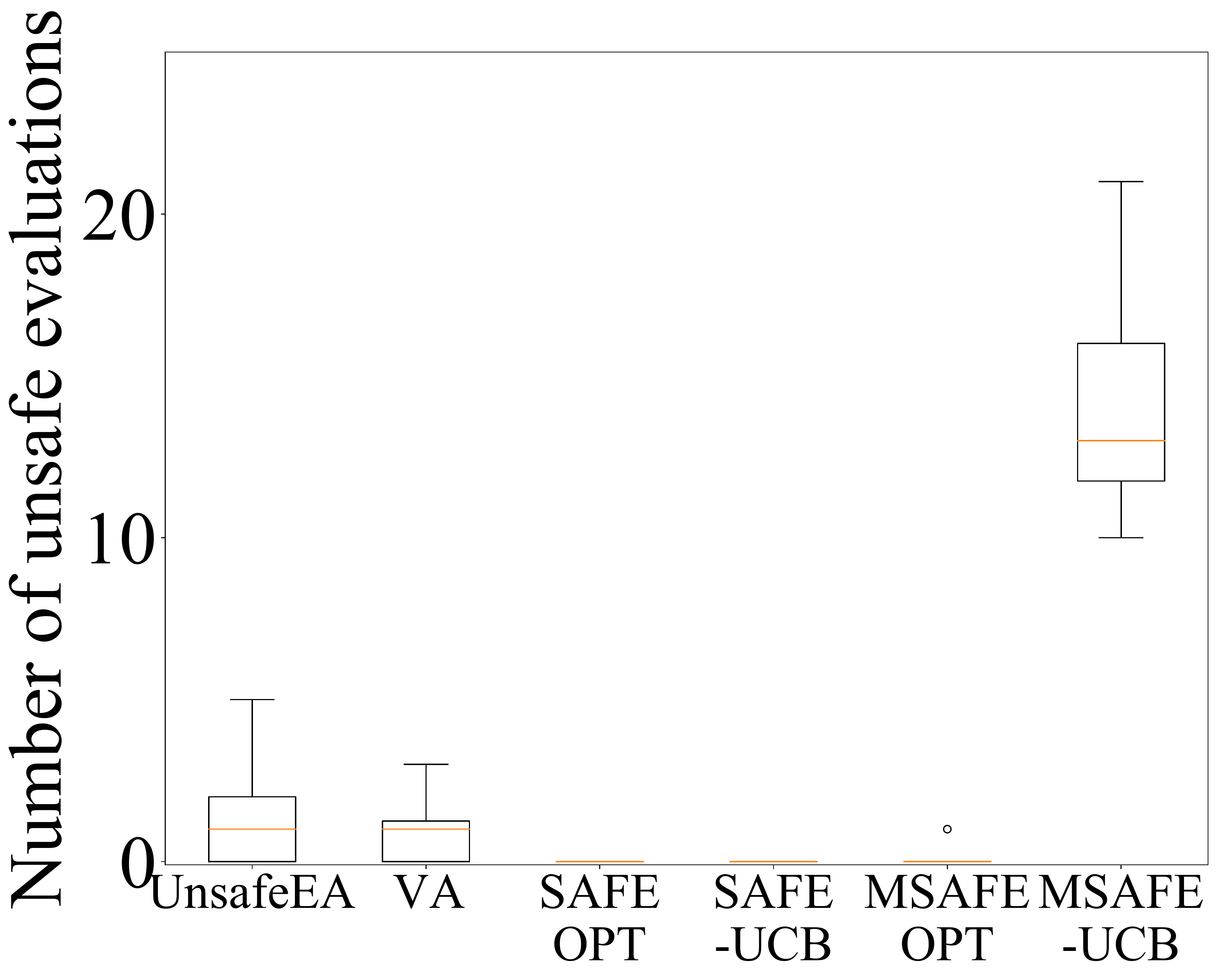}
         \caption{Unsafe: $h=65^\mathrm{th}$}
         \label{fig:Spherenuminit2Unsafe}
     \end{subfigure}
     \begin{subfigure}{0.23\textwidth}
         \centering
         \includegraphics[width=\textwidth]{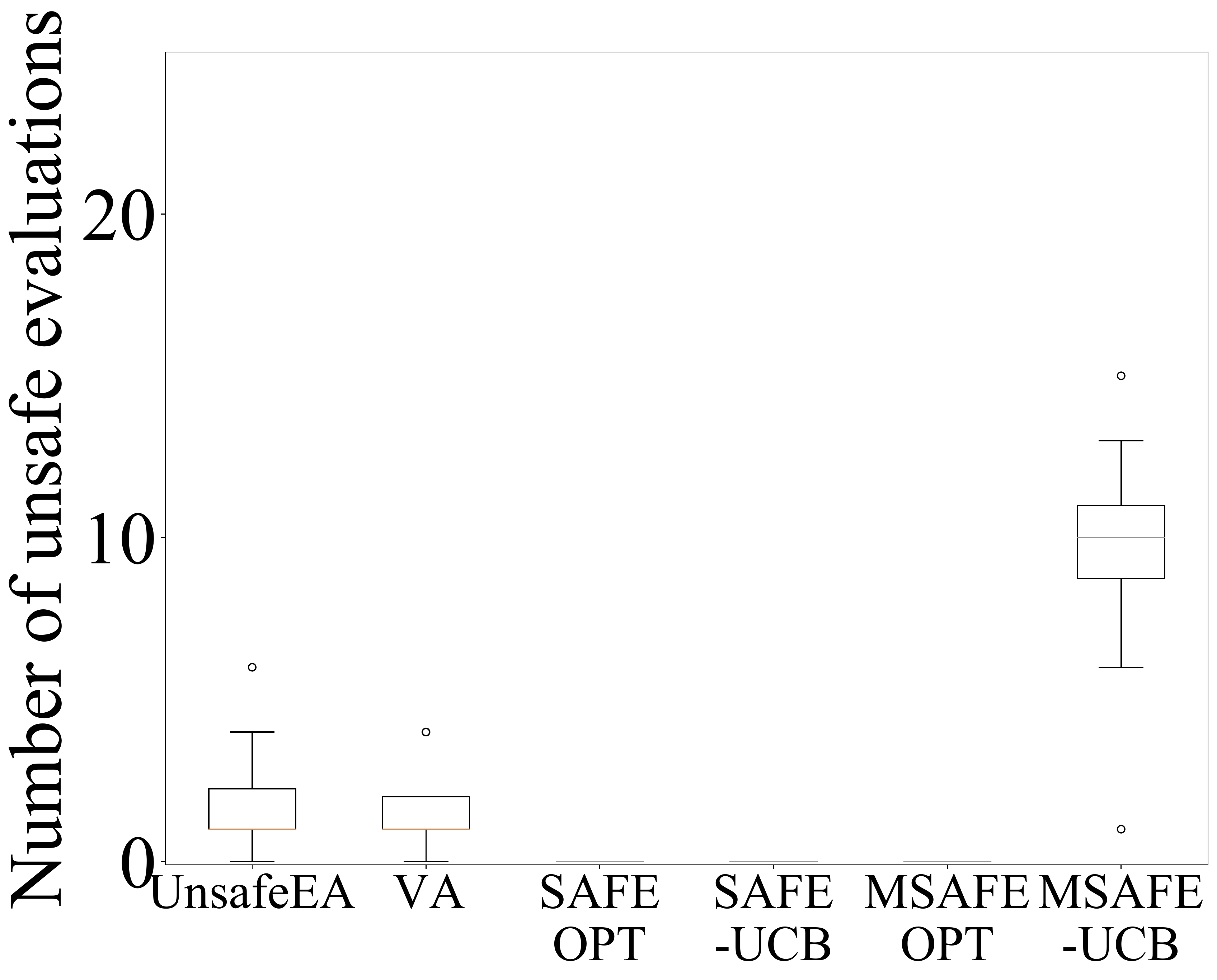}
         \caption{Unsafe: $h=75^\mathrm{th}$}
         \label{fig:Spherenuminit10Unsafe}
     \end{subfigure} 
        \caption{Plots show the average best objective function value (BSF) and standard error as function of the number of function evaluations and distribution of the number of unsafe solutions evaluated across 20 algorithmic runs on the Stylinski-Tang function (2 initial safe seeds). Initial safe seeds were sampled with scenario 1.}
        \label{fig:InitSafe}
\end{figure}

\begin{figure}[!hp]
     \centering
     \begin{subfigure}{0.23\textwidth}
         \centering
         \includegraphics[width=\textwidth]{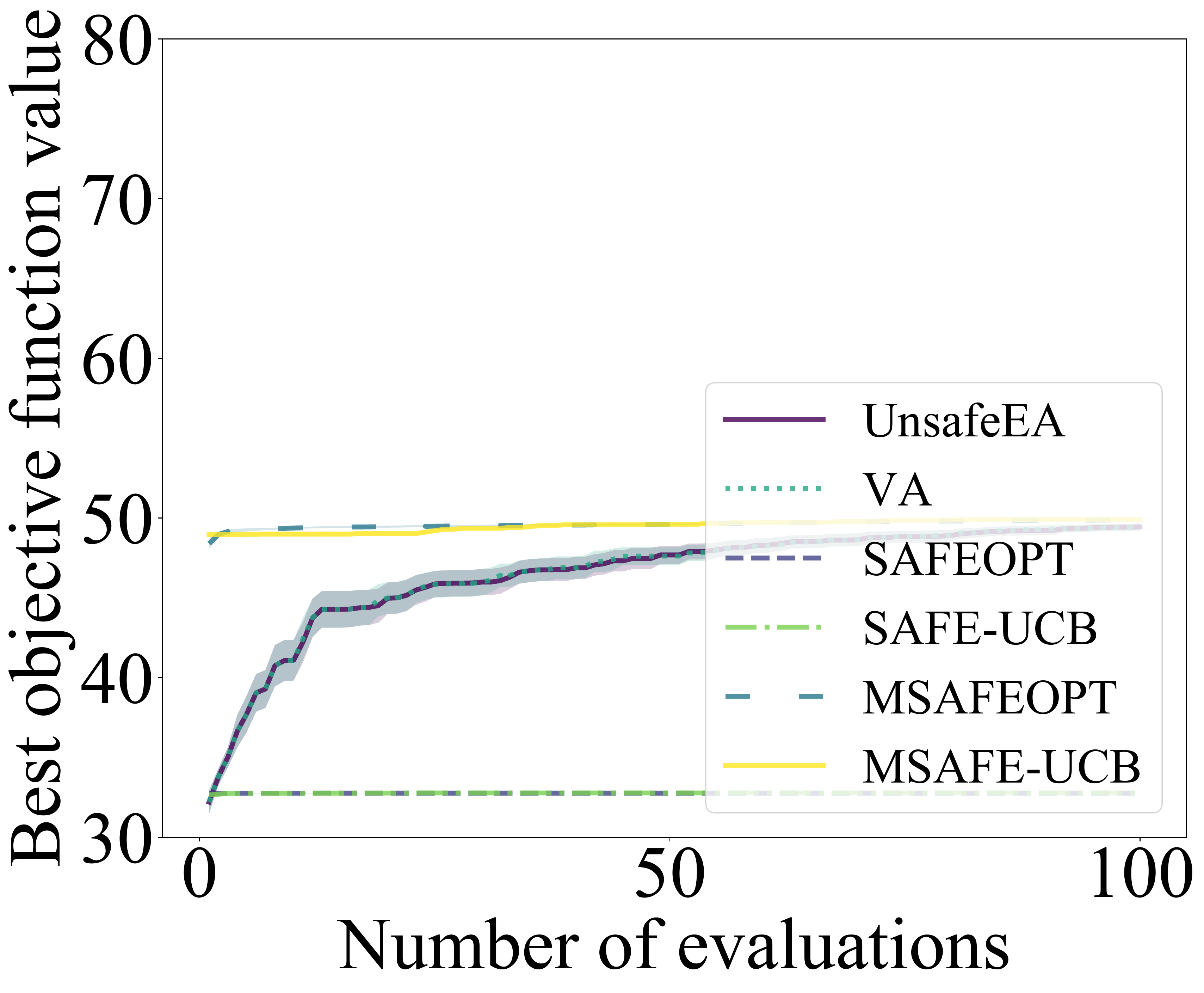}
         \caption{BSF: $h=65^\mathrm{th}$}
         \label{fig:Spherenuminit2BSF}
     \end{subfigure}
     \begin{subfigure}{0.23\textwidth}
         \centering
         \includegraphics[width=\textwidth]{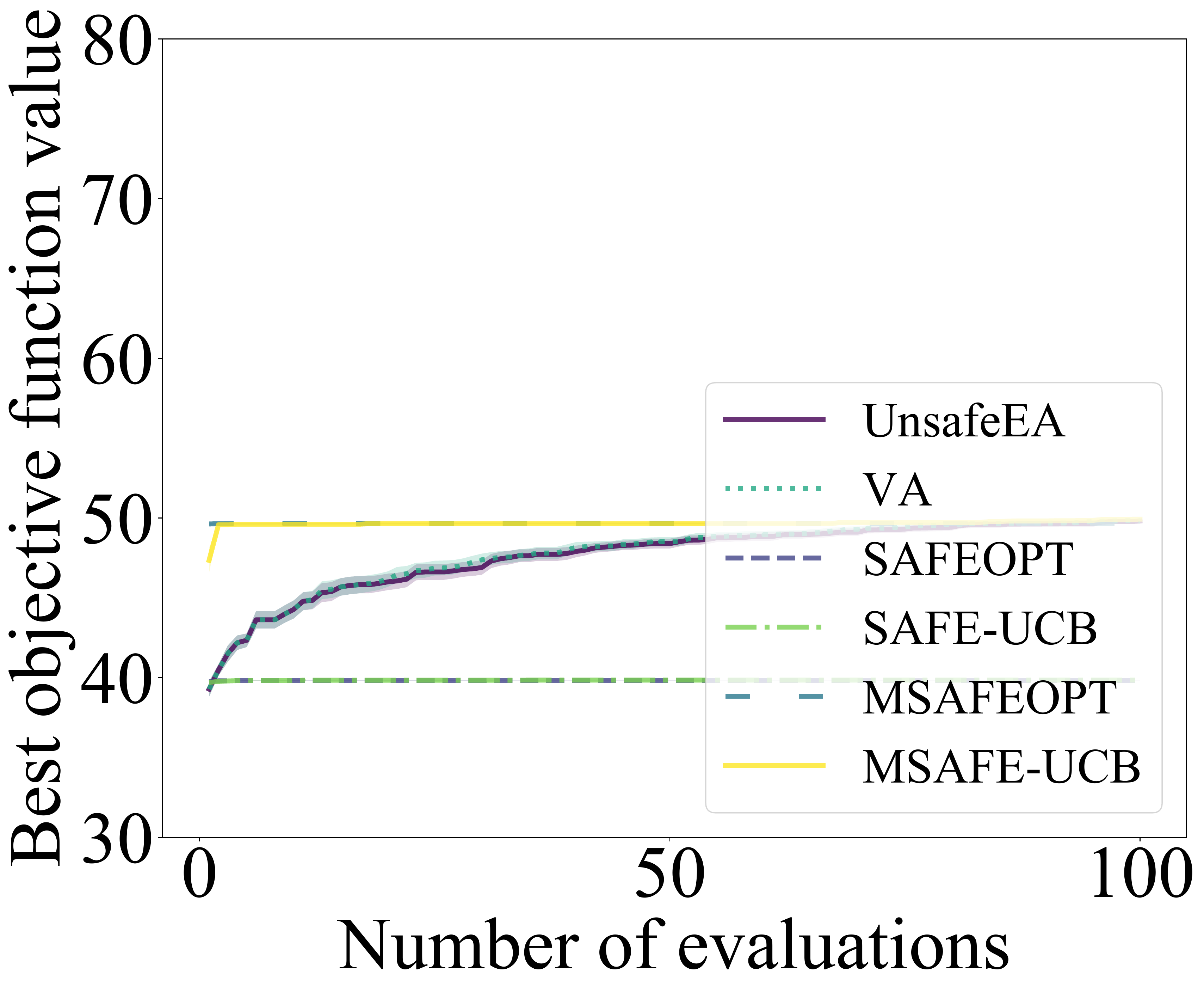}
         \caption{BSF: $h=75^\mathrm{th}$}
         \label{fig:Spherenuminit10BSF}
     \end{subfigure}
     \hfill
     \begin{subfigure}{0.23\textwidth}
         \centering
         \includegraphics[width=\textwidth]{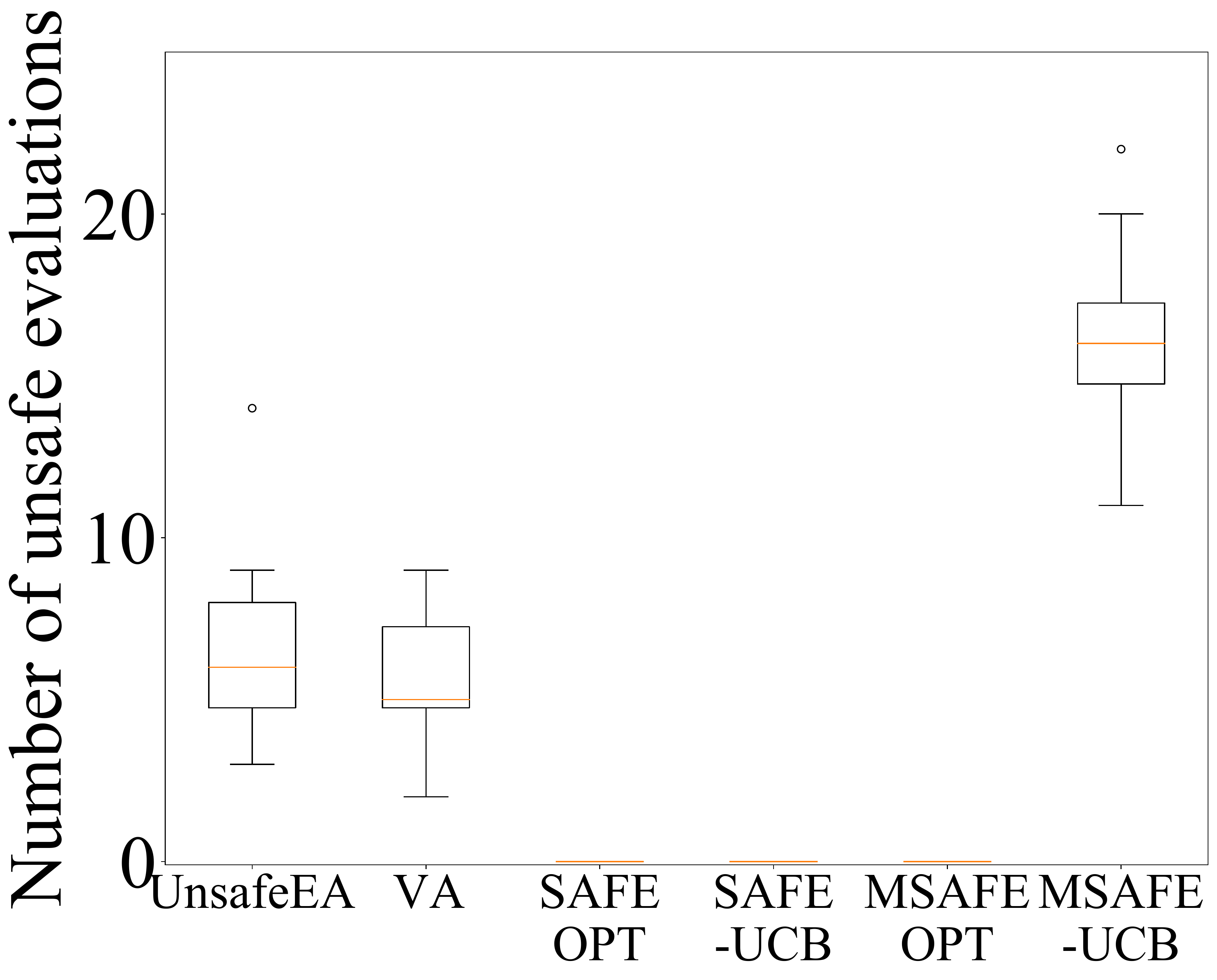}
         \caption{Unsafe: $h=65^\mathrm{th}$}
         \label{fig:Spherenuminit2Unsafe}
     \end{subfigure}
     \begin{subfigure}{0.23\textwidth}
         \centering
         \includegraphics[width=\textwidth]{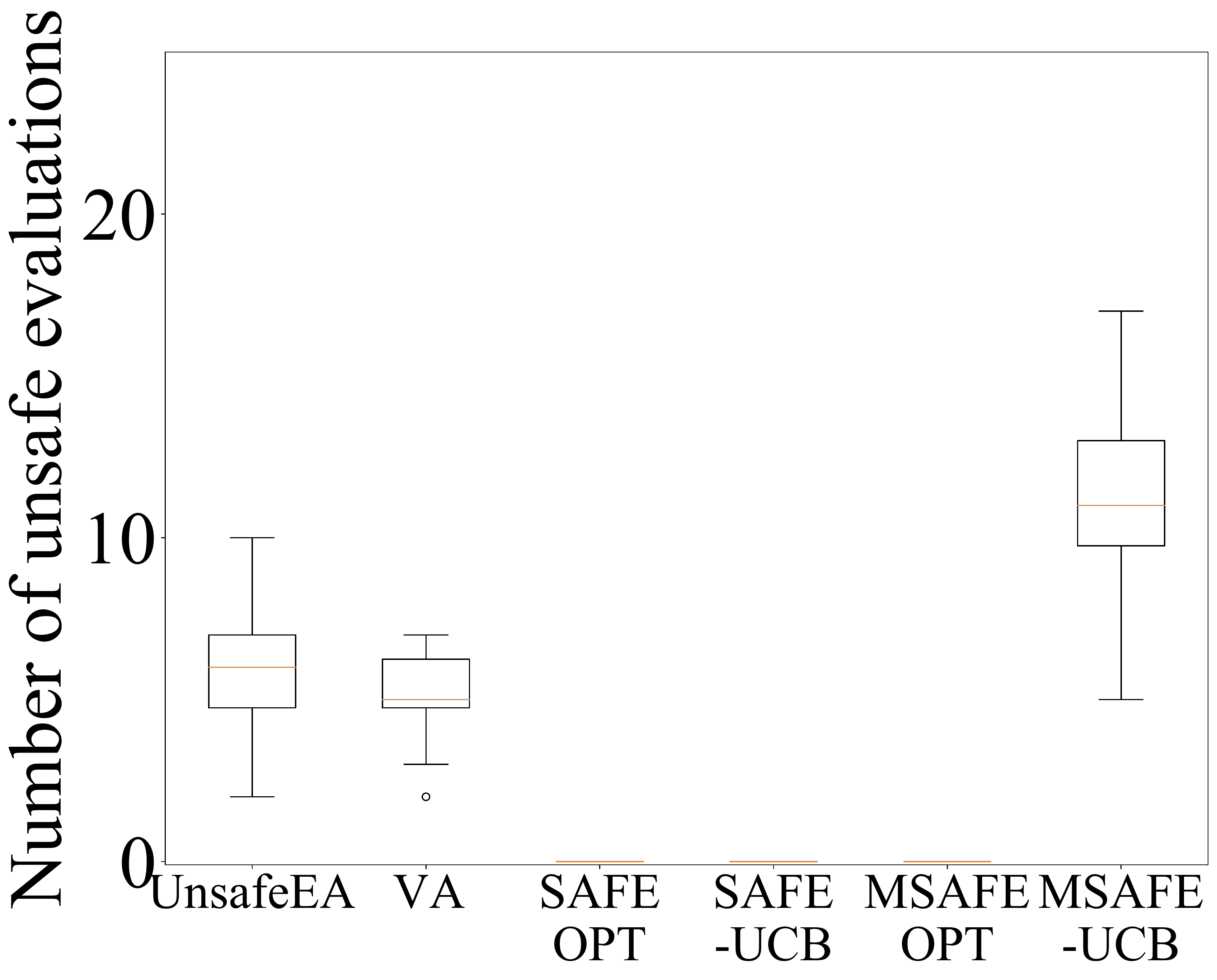}
         \caption{Unsafe: $h=75^\mathrm{th}$}
         \label{fig:Spherenuminit10Unsafe}
     \end{subfigure} 
        \caption{Plots show the average best objective function value (BSF) and standard error as function of the number of function evaluations and distribution of the number of unsafe solutions evaluated across 20 algorithmic runs on the Stylinski-Tang function (10 initial safe seeds). Initial safe seeds were sampled with scenario 1.}
        \label{fig:InitSafe}
      \end{figure}
      
\begin{figure}[!hp]
     \centering
     \begin{subfigure}{0.23\textwidth}
         \centering
         \includegraphics[width=\textwidth]{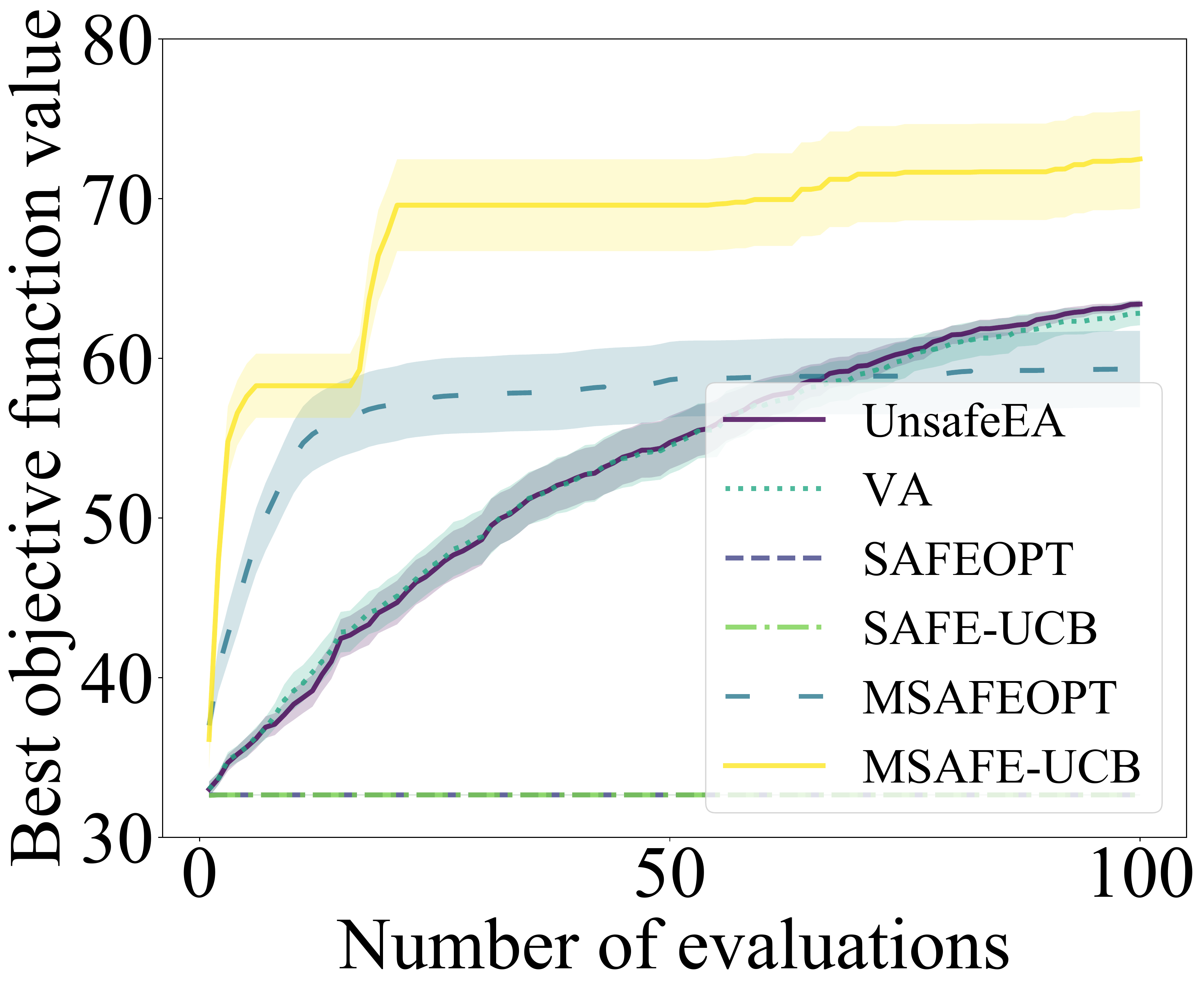}
         \caption{BSF: $h=65^\mathrm{th}$}
         \label{fig:Spherenuminit2BSF}
     \end{subfigure}
     \begin{subfigure}{0.23\textwidth}
         \centering
         \includegraphics[width=\textwidth]{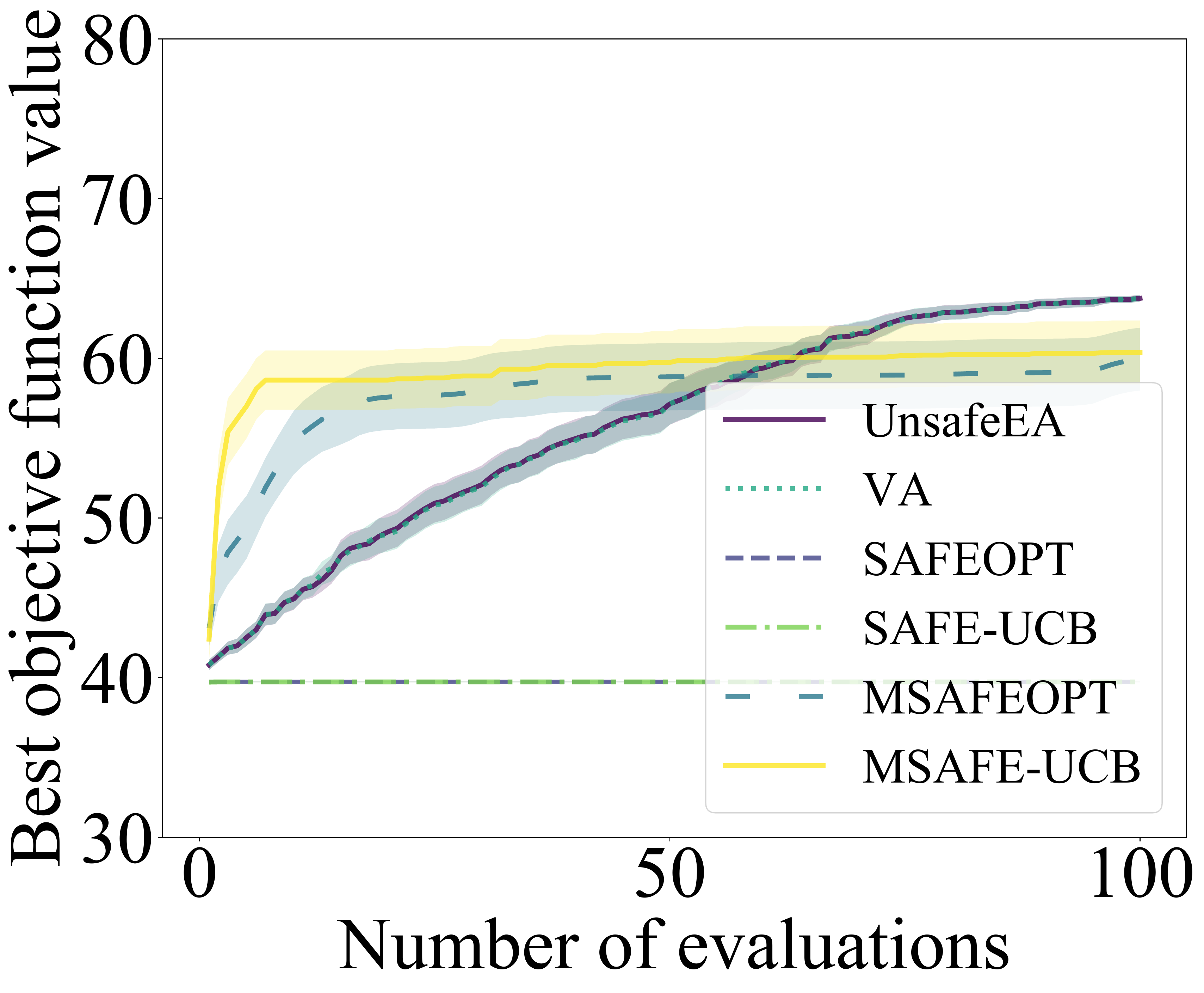}
         \caption{BSF: $h=75^\mathrm{th}$}
         \label{fig:Spherenuminit10BSF}
     \end{subfigure}
     \hfill
     \hfill
     \hfill
     \hfill
     \hfill
     \hfill
     \hfill
     \hfill
     \begin{subfigure}{0.23\textwidth}
         \centering
         \includegraphics[width=\textwidth]{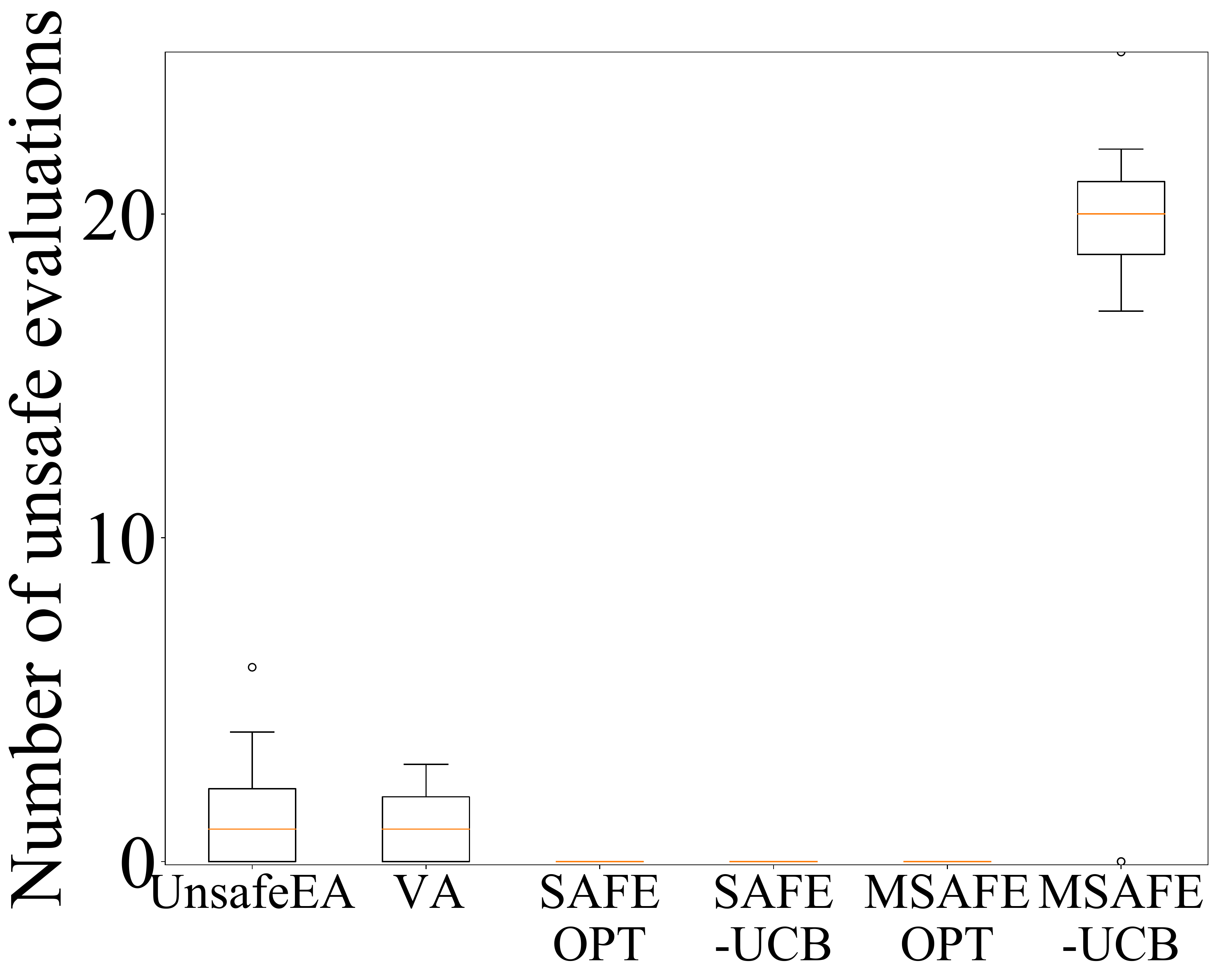}
         \caption{Unsafe: $h=65^\mathrm{th}$}
         \label{fig:Spherenuminit2Unsafe}
     \end{subfigure}
     \hfill
     \begin{subfigure}{0.23\textwidth}
         \centering
         \includegraphics[width=\textwidth]{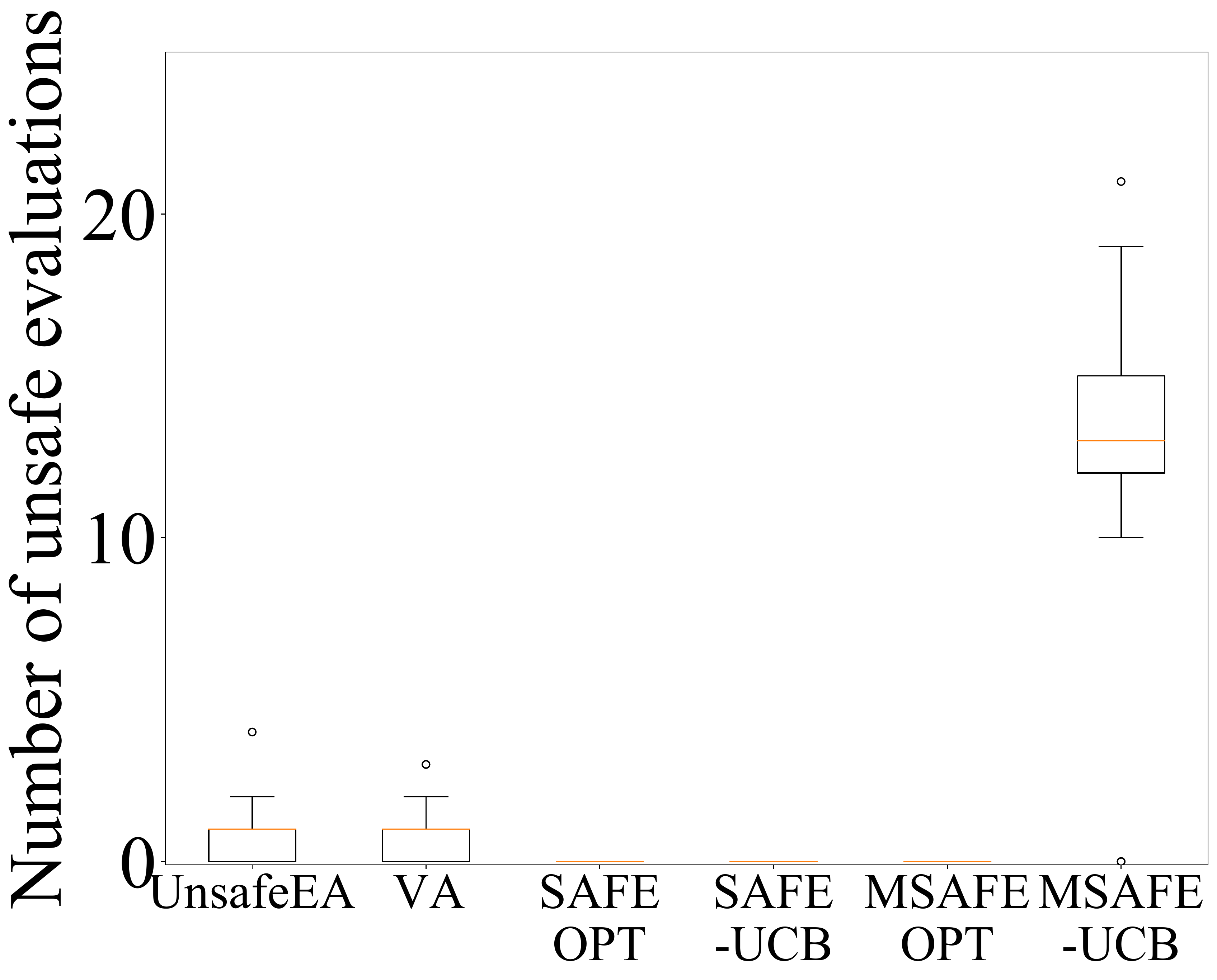}
         \caption{Unsafe: $h=75^\mathrm{th}$}
         \label{fig:Spherenuminit10Unsafe}
     \end{subfigure} 
        \caption{Plots show the average best objective function value (BSF) and standard error as function of the number of function evaluations and distribution of the number of unsafe solutions evaluated across 20 algorithmic runs on the Stylinski-Tang function (2 initial safe seeds). Initial safe seeds were sampled with scenario 2.}
        \label{fig:InitSafe}
\end{figure}

\begin{figure}[!p]
     \centering
     \begin{subfigure}{0.23\textwidth}
         \centering
         \includegraphics[width=\textwidth]{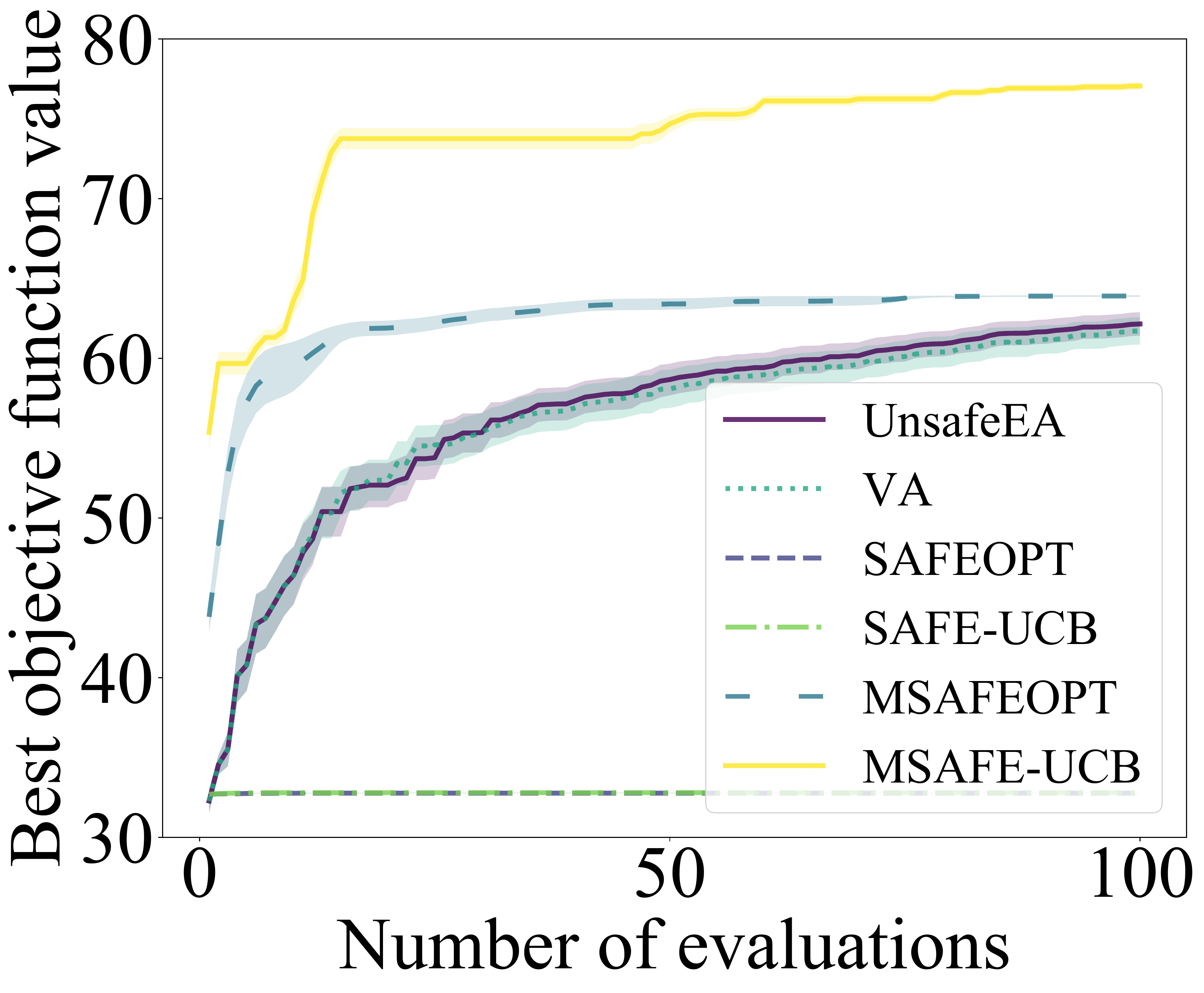}
         \caption{BSF: $h=65^\mathrm{th}$}
         \label{fig:Spherenuminit2BSF}
     \end{subfigure}
     \begin{subfigure}{0.23\textwidth}
         \centering
         \includegraphics[width=\textwidth]{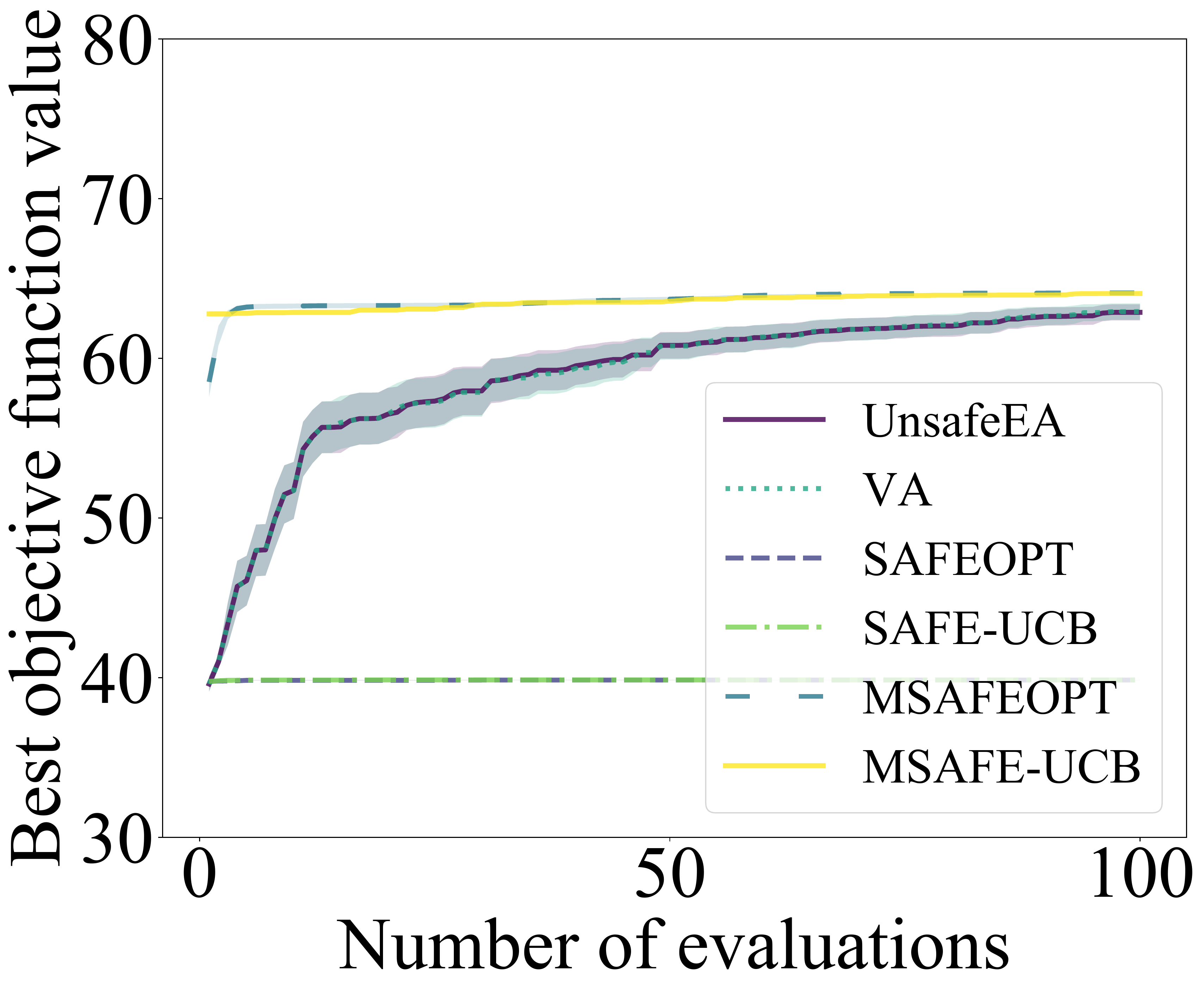}
         \caption{BSF: $h=75^\mathrm{th}$}
         \label{fig:Spherenuminit10BSF}
     \end{subfigure}
     \hfill
     \begin{subfigure}{0.23\textwidth}
         \centering
         \includegraphics[width=\textwidth]{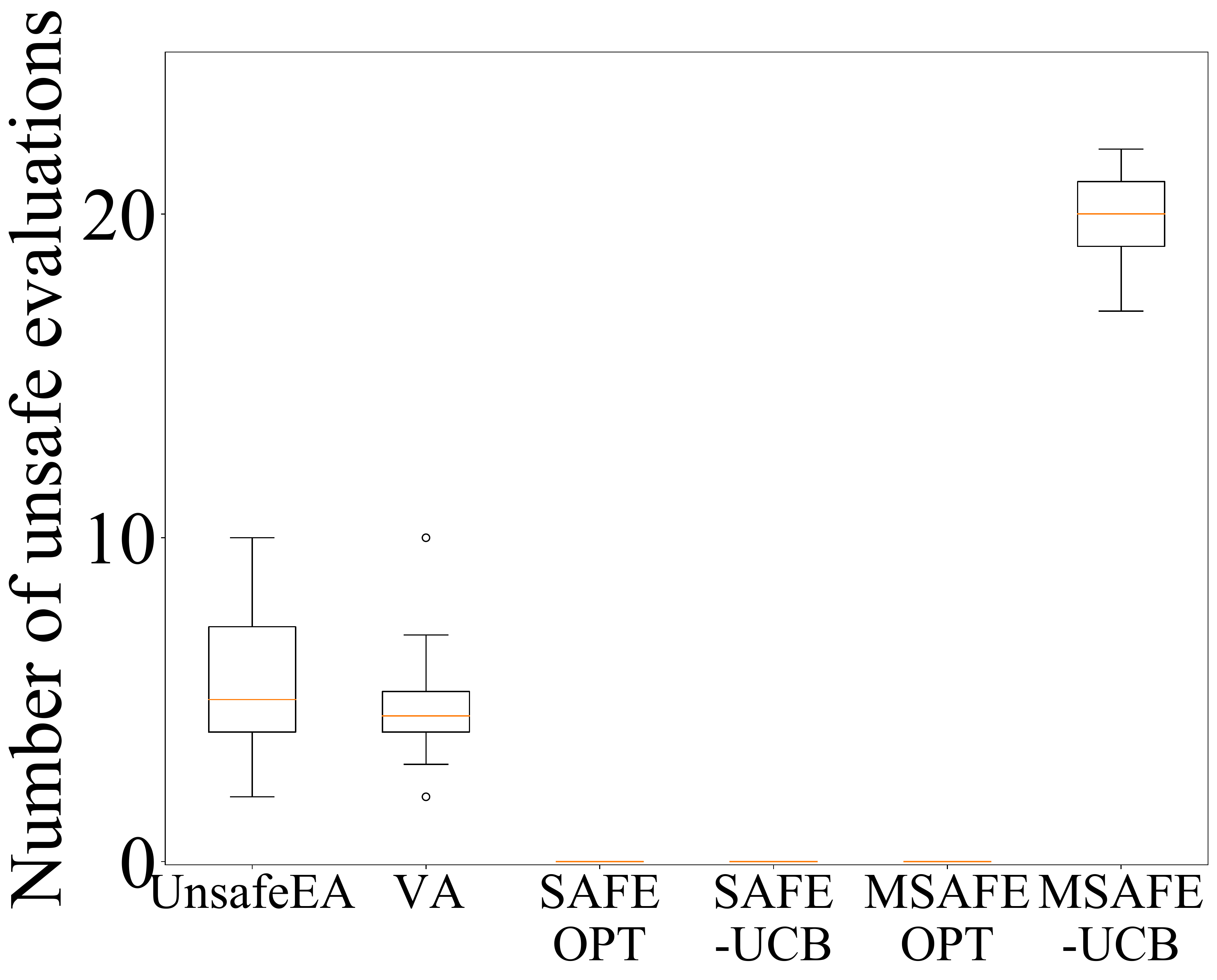}
         \caption{Unsafe: $h=65^\mathrm{th}$}
         \label{fig:Spherenuminit2Unsafe}
     \end{subfigure}
     \begin{subfigure}{0.23\textwidth}
         \centering
         \includegraphics[width=\textwidth]{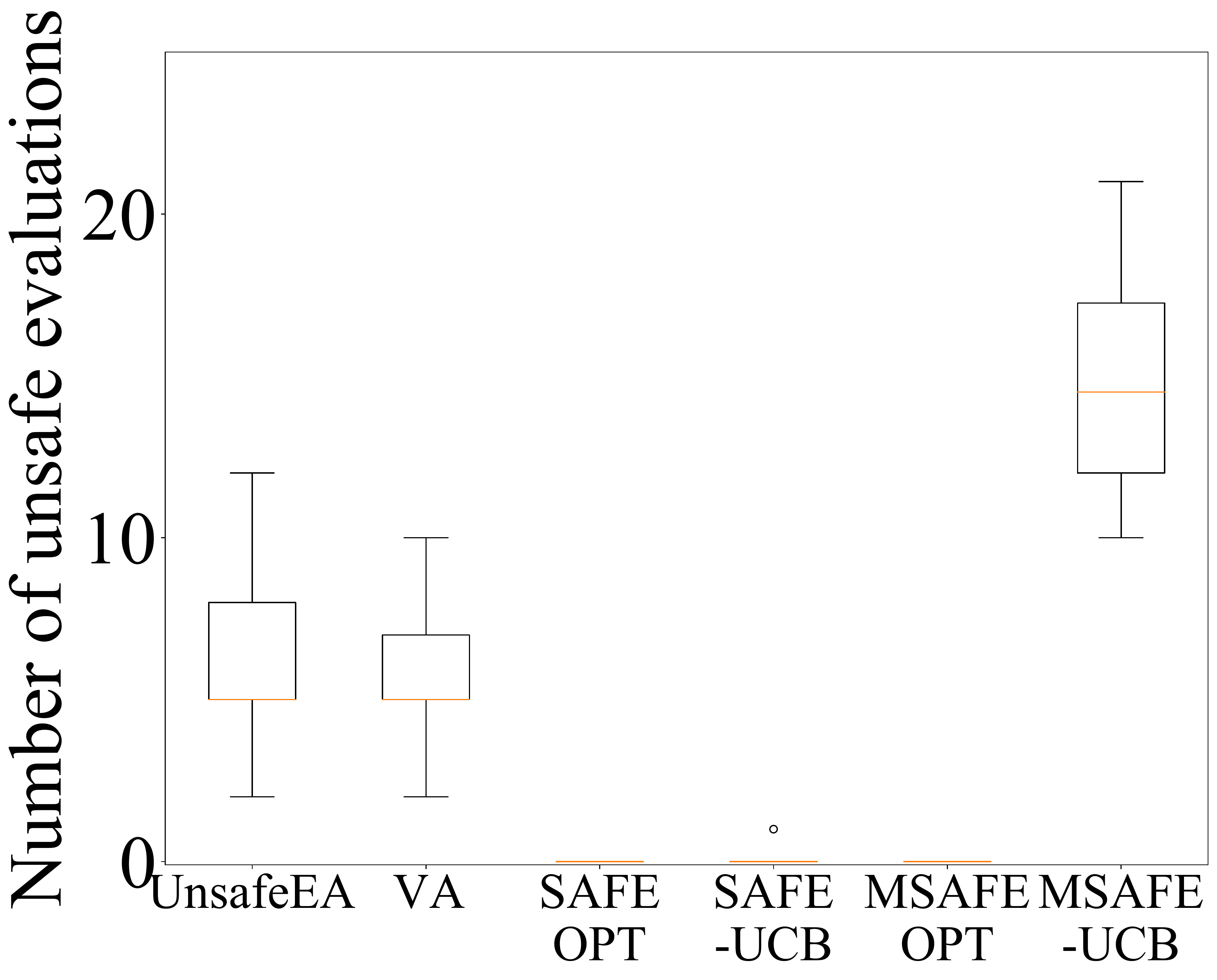}
         \caption{Unsafe: $h=75^\mathrm{th}$}
         \label{fig:Spherenuminit10Unsafe}
     \end{subfigure} 
        \caption{Plots show the average best objective function value (BSF) and standard error as function of the number of function evaluations and distribution of the number of unsafe solutions evaluated across 20 algorithmic runs on the Stylinski-Tang function (10 initial safe seeds). Initial safe seeds were sampled with scenario 2.}
        \label{fig:InitSafe}
\end{figure}
\newpage

\begin{figure}
     \centering
     \begin{subfigure}{0.23\textwidth}
         \centering
         \includegraphics[width=\textwidth]{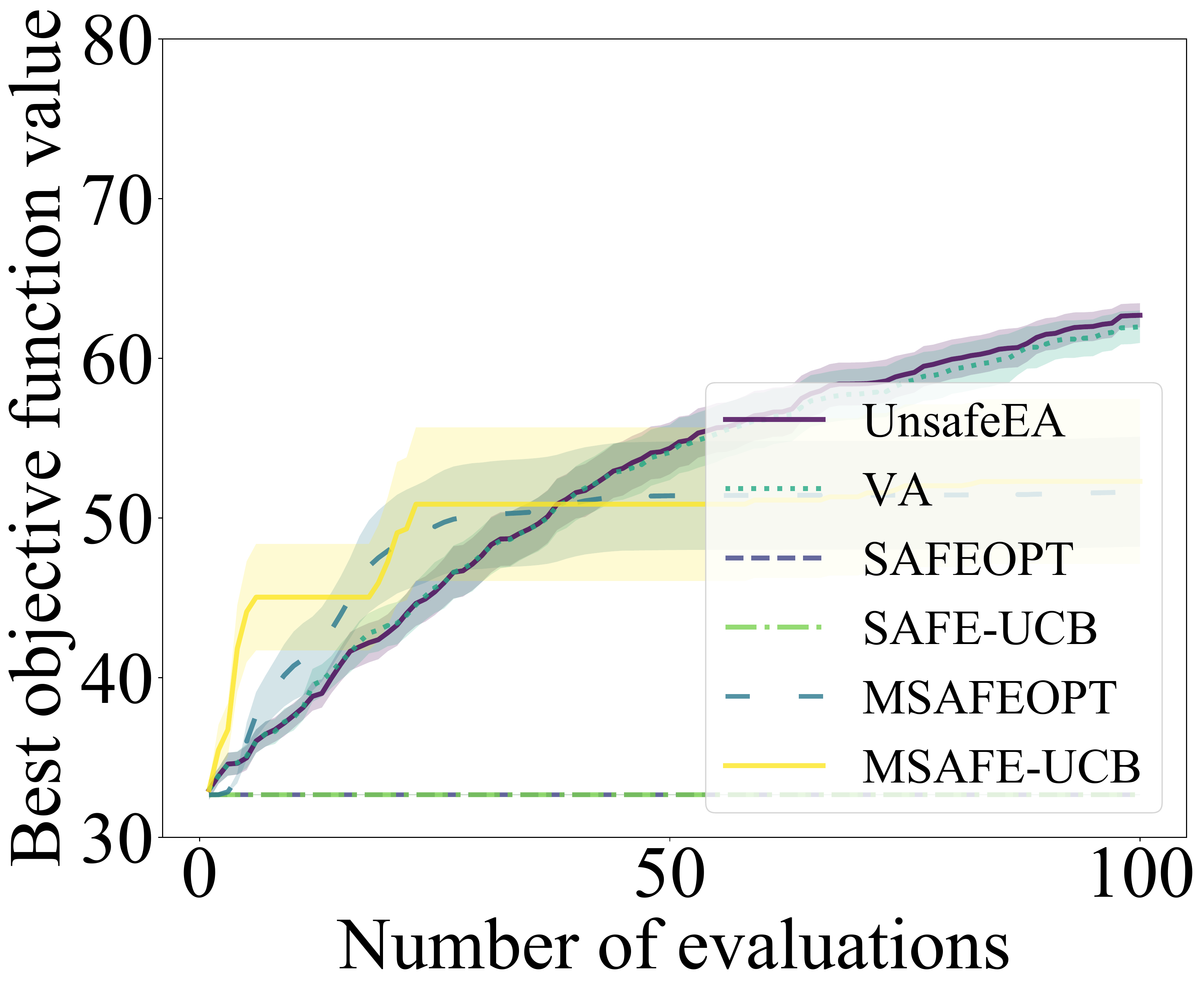}
         \caption{BSF: $h=65^\mathrm{th}$}
         \label{fig:Spherenuminit2BSF}
     \end{subfigure}
     \begin{subfigure}{0.23\textwidth}
         \centering
         \includegraphics[width=\textwidth]{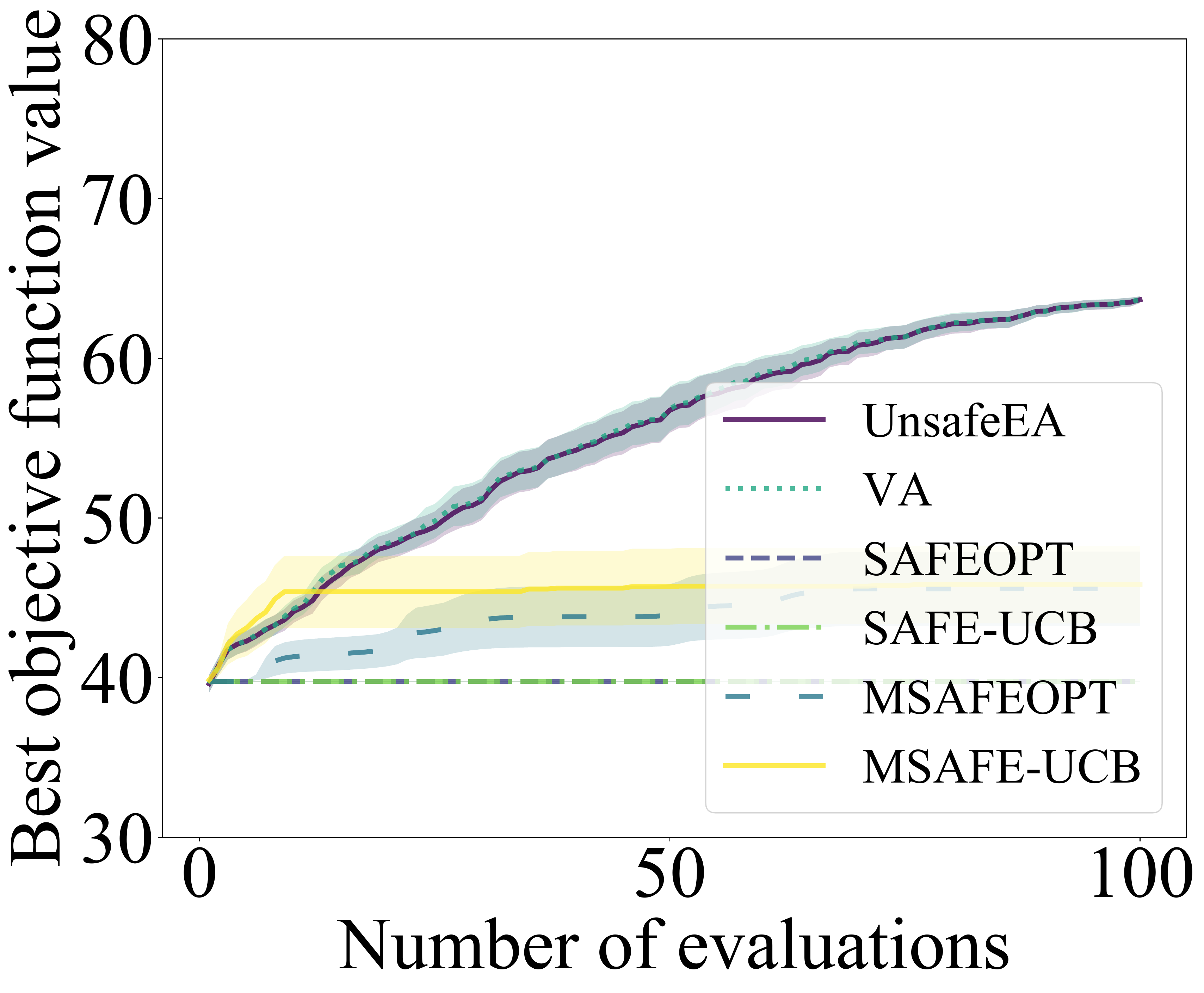}
         \caption{BSF: $h=75^\mathrm{th}$}
         \label{fig:Spherenuminit10BSF}
     \end{subfigure}
     \hfill
     \begin{subfigure}{0.23\textwidth}
         \centering
         \includegraphics[width=\textwidth]{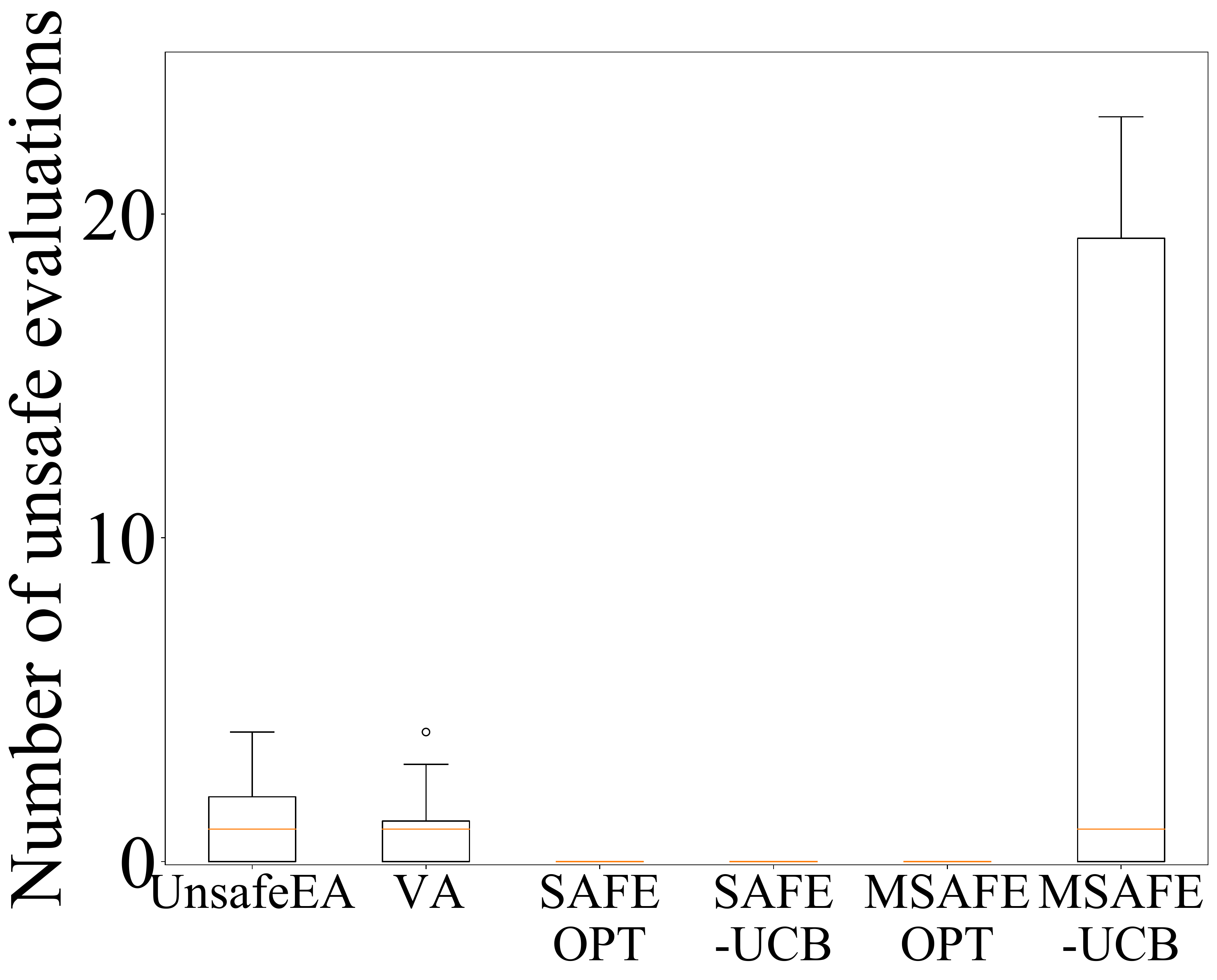}
         \caption{Unsafe: $h=65^\mathrm{th}$}
         \label{fig:Spherenuminit2Unsafe}
     \end{subfigure}
     \begin{subfigure}{0.23\textwidth}
         \centering
         \includegraphics[width=\textwidth]{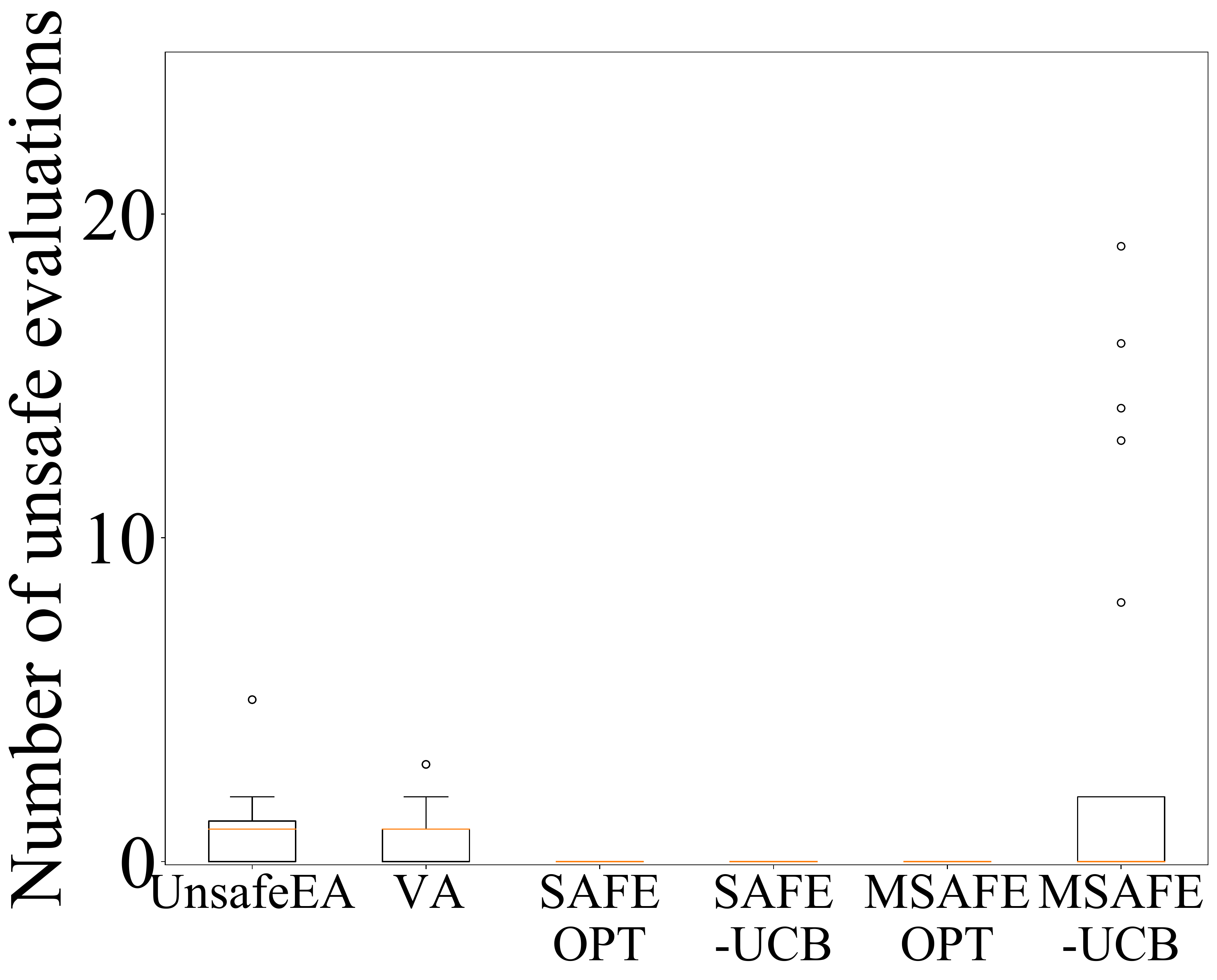}
         \caption{Unsafe: $h=75^\mathrm{th}$}
         \label{fig:Spherenuminit10Unsafe}
     \end{subfigure} 
        \caption{Plots show the average best objective function value (BSF) and standard error as function of the number of function evaluations and distribution of the number of unsafe solutions evaluated across 20 algorithmic runs on the Stylinski-Tang function (2 initial safe seeds). Initial safe seeds were sampled with scenario 3.}
        \label{fig:InitSafe}
\end{figure}

\begin{figure}
     \centering
     \begin{subfigure}{0.23\textwidth}
         \centering
         \includegraphics[width=\textwidth]{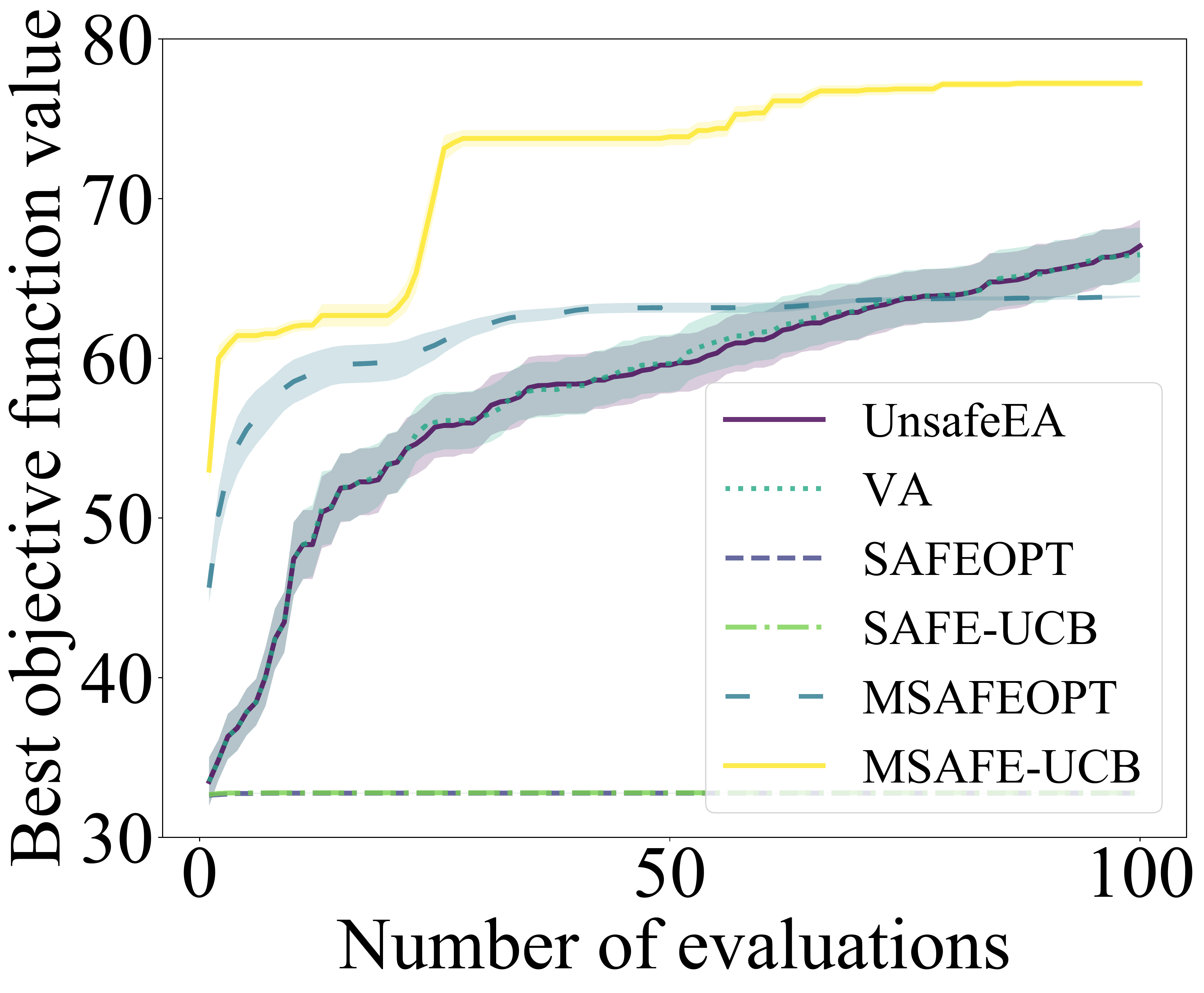}
         \caption{BSF: $h=65^\mathrm{th}$}
         \label{fig:Spherenuminit2BSF}
     \end{subfigure}
     \begin{subfigure}{0.23\textwidth}
         \centering
         \includegraphics[width=\textwidth]{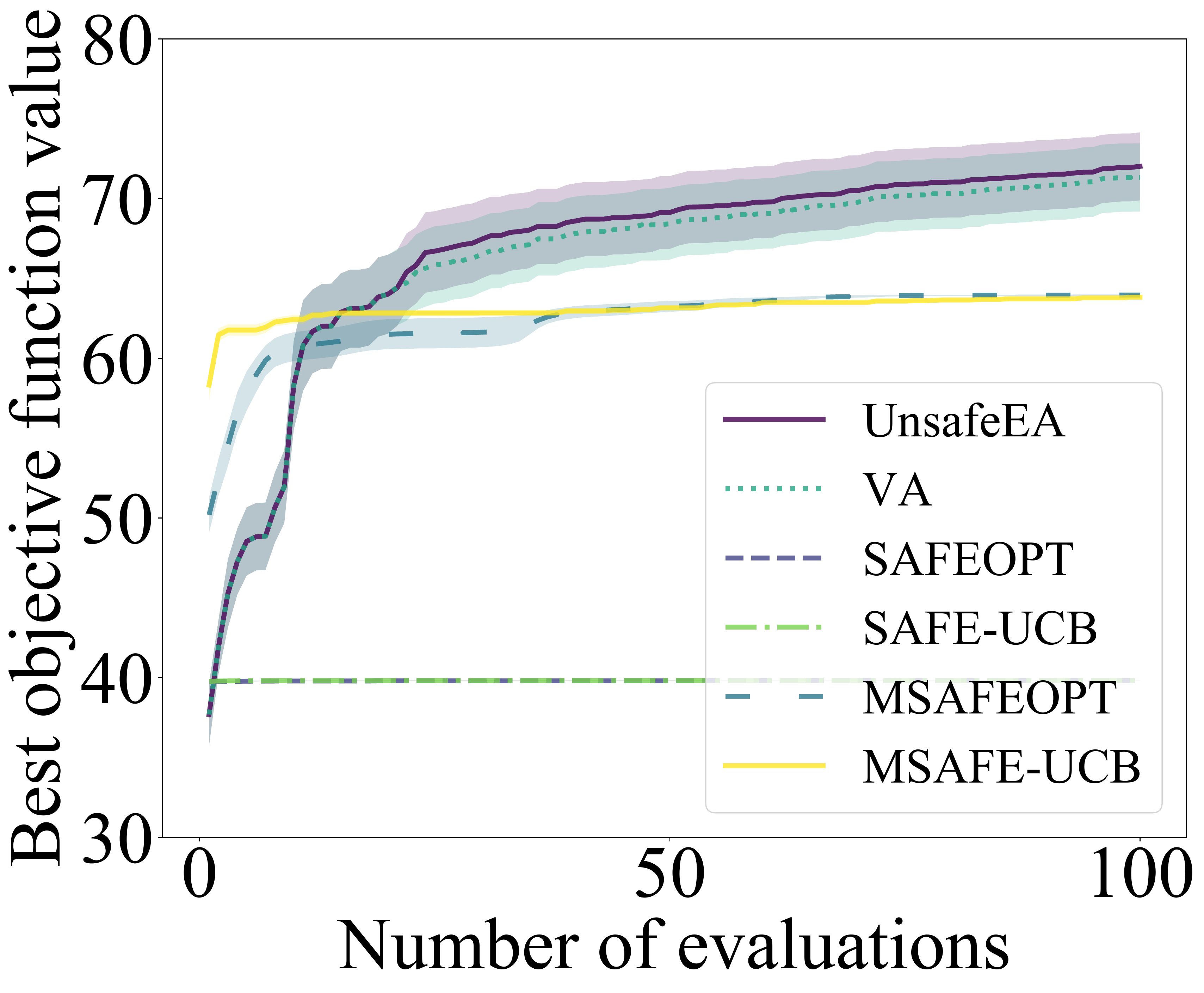}
         \caption{BSF: $h=75^\mathrm{th}$}
         \label{fig:Spherenuminit10BSF}
     \end{subfigure}
     \hfill
     \begin{subfigure}{0.23\textwidth}
         \centering
         \includegraphics[width=\textwidth]{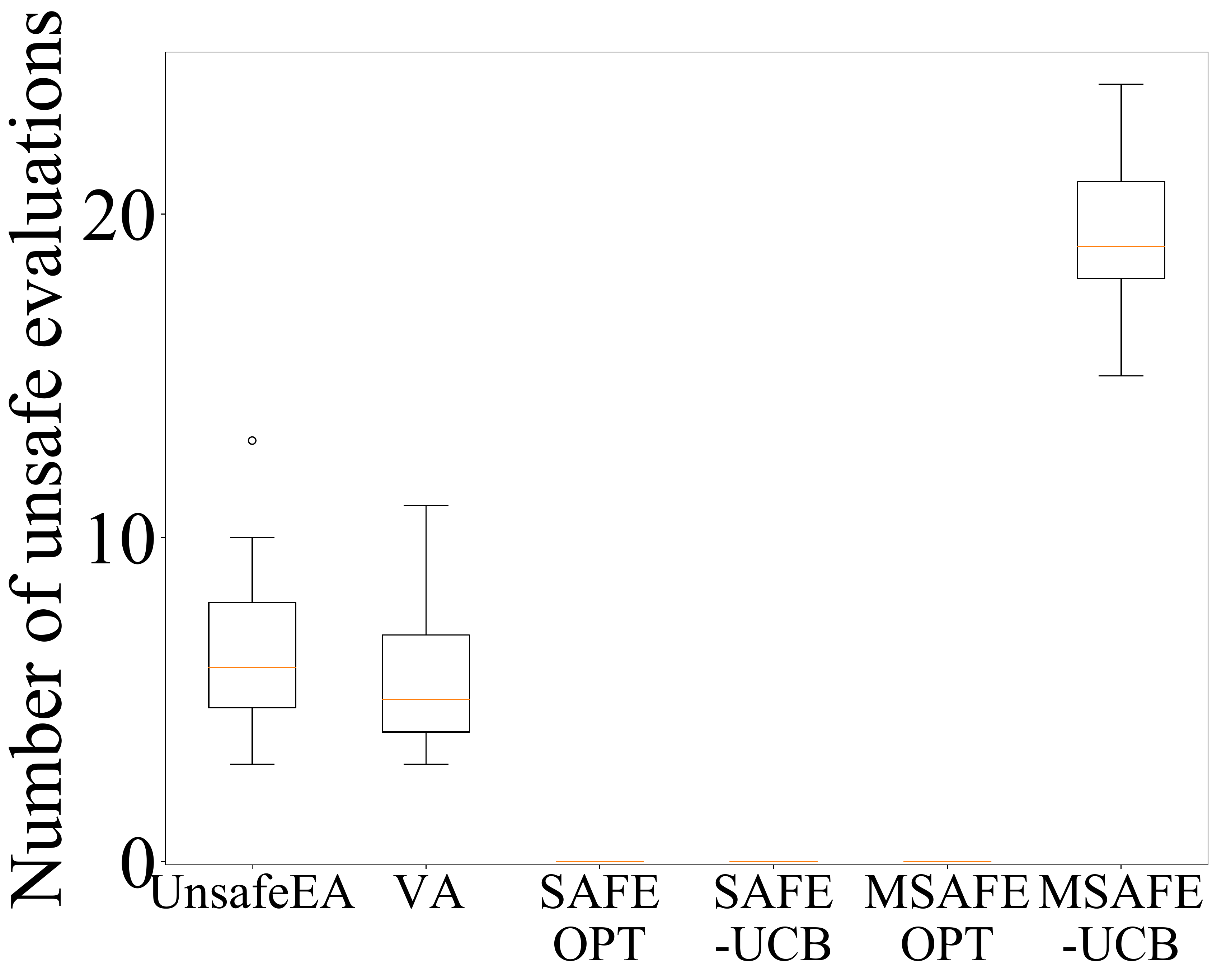}
         \caption{Unsafe: $h=65^\mathrm{th}$}
         \label{fig:Spherenuminit2Unsafe}
     \end{subfigure}
     \begin{subfigure}{0.23\textwidth}
         \centering
         \includegraphics[width=\textwidth]{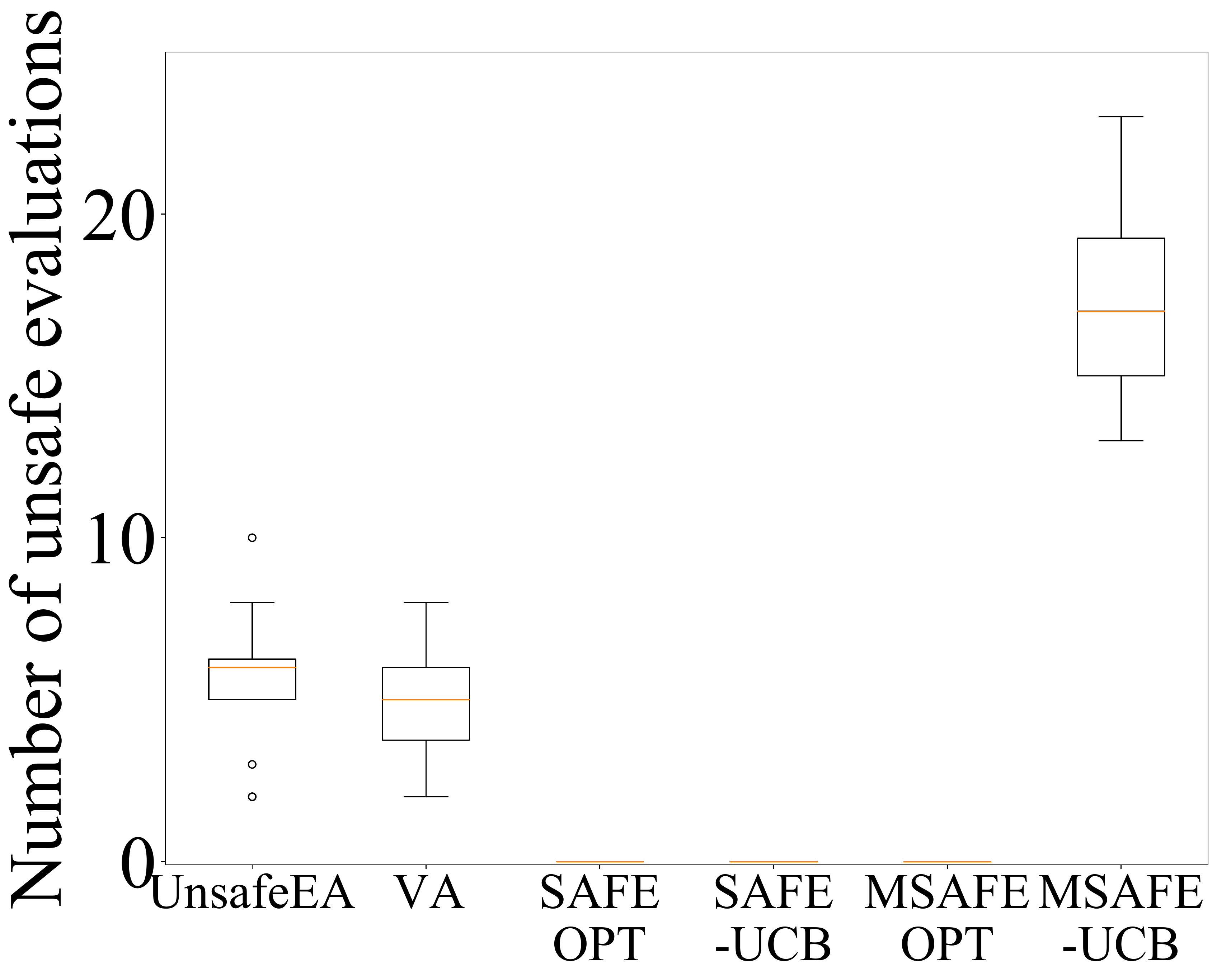}
         \caption{Unsafe: $h=75^\mathrm{th}$}
         \label{fig:Spherenuminit10Unsafe}
     \end{subfigure} 
        \caption{Plots show the average best objective function value (BSF) and standard error as function of the number of function evaluations and distribution of the number of unsafe solutions evaluated across 20 algorithmic runs on the Stylinski-Tang function (10 initial safe seeds). Initial safe seeds were sampled with scenario 3.}
        \label{fig:InitSafe}
\end{figure}

\clearpage

\bibliographystyle{ACM-Reference-Format}
\bibliography{bib/abbrev,bib/journals,bib/authors,bib/biblio,bib/crossref,refs}

